\documentclass{article}

     \PassOptionsToPackage{numbers, compress}{natbib}
\usepackage[preprint]{neurips_2024}

\usepackage[utf8]{inputenc} %
\usepackage[T1]{fontenc}    %
\usepackage{hyperref}       %
\usepackage{url}            %
\usepackage{booktabs}       %
\usepackage{amsfonts}       %
\usepackage{nicefrac}       %
\usepackage{microtype}      %
\usepackage{xcolor}         %

\usepackage{isabelle}
\usepackage{isabellesym}

\usepackage{amsmath}
\usepackage{amssymb}
\usepackage{url}
\usepackage{hyperref}
\usepackage{wrapfig}

\usepackage{amsmath, amssymb, amsthm}
\usepackage{mathtools}
\usepackage[ruled,noline,noalgohanging]{algorithm2e}
\usepackage{tabularx}
\usepackage{multirow}
\usepackage{todonotes}
\setuptodonotes{fancyline, color=blue!30}
\usepackage{subcaption}

\usepackage[most]{tcolorbox}
\tcbuselibrary{breakable,hooks,skins,theorems}

\newcommand{\varassign}{\coloneqq}

\newtheorem{theorem}{Theorem}

\theoremstyle{definition}
\newtheorem{myalgorithm}{Algorithm}

\newtheorem{specification}{Specification}

\usepackage[most]{tcolorbox}
\tcbuselibrary{breakable,hooks,skins,theorems}
\newtcbtheorem{IsabelleTheorem}{Listing}{
    enhanced jigsaw
    , breakable
    , float=h
    , top=0pt
    , right=0pt
    , bottom=0pt
    , left=0pt
    , boxrule=0pt
    , toprule=1pt
    , bottomrule=1pt
    , titlerule=1pt
    , colframe=black
    , sharp corners
    , colbacktitle=white
    , coltitle=black
    , fonttitle=\tiny
    , colback=white
    , fontupper=\ttfamily\tiny
    , before upper={\parindent0em}
}{isa}

\lstdefinelanguage{isabelle}{
    morekeywords={fixes, assumes, locale, record,type_synonym,definition,fun,function,primrec,where,lemma,theorem,unfolding,by,shows,assumes,and,datatype,using,abbreviation
,moreover,have,hence,thus,qed,proof,let,ultimately,show,next,in}
    , sensitive=true
    , showstringspaces=false
    , framerule=0pt
    , xleftmargin=2em
    , numbers=left
    , numberstyle=\ttfamily\tiny
    , firstnumber=1
    , stepnumber=2
    , basicstyle=\ttfamily\tiny
    , breaklines=true
    , showspaces=false
    , morecomment=[l]{--}
    , morecomment=[l]{―}
    , morecomment=[s]{(*}{*)}
    , commentstyle=\color{red}
    , morestring=[b]"
    , literate={\\<times>}{{$\times$}}{1} {\\<equiv>}{{$\equiv$}}{1} {\\<forall>}{{$\forall$}}{1} {\\<exists>}{{$\exists$}}{1} {\\<and>}{{$\land$}}{1}
        {\\<in>}{{$\in$}}{1} {\\<Rightarrow>}{{$\Rightarrow$}}{1} {\\<lambda>}{{$\lambda$}}{1} {::}{{$::$}}{1}
        {\\<subseteq>}{{$\subseteq$}}{1} 
        {\\<circ>}{{$\circ$}}{1} 
        {\\<^sub>m}{{$_m$}}{1} 
        {\\<longleftrightarrow>}{{$\longleftrightarrow$}}{3}
        {\\<pi>}{{$\pi$}}{1} {\\<delta>}{{$\delta$}}{1} 
        {\\<omega>}{{$\omega$}}{1}
        {\\<bind>}{$\bindop$}{1}
        {\\<And>}{$\bigwedge$}{1}
        {\\<dots>}{$\ldots$}{3}
        {\\<nu>}{$\nu$}{1}
        {\\<Sum>}{$\sum$}{2}
        {\\<integral>}{$\int$}{1}
        {\\<partial>}{$\partial$}{2}
        {\\<T>}{{$\mathcal T$}}{1}
        {\\<P>}{{$\mathcal P$}}{1}
        {\\<L>}{{$\mathcal L$}}{1}
        {\\<Squnion>}{{$\bigsqcup$}}{2}
        {\\<lbrakk>}{{$\llbracket$}}{1} {\\<rbrakk>}{{$\rrbracket$}}{1}
        {\\<Longrightarrow>}{{$\Longrightarrow$}}{3}
        {\\<longlonglongrightarrow>}{{$\xrightarrow{\hspace*{0.4cm}}$}}{4}
        {\\<not>}{{$\lnot$}}{1} {\\<le>}{{$\le$}}{1} {\\<rightharpoonup>}{{$\rightharpoonup$}}{2}
        {\\<^sub>\\<V>}{{$_{\mathcal V}$}}{1}
        {\\<^sub>H}{{$_{\texttt H}$}}{1}
        {\\<^sub>R}{{$_{\texttt R}$}}{1}
        {\\<^sub>D}{{$_{\texttt D}$}}{1}
        {\\<^sub>U}{{$_{\texttt U}$}}{1}
        {\\<^sub>b}{{$_{\texttt{b}}$}}{1}
        {\\<lparr>}{{$\llparenthesis$}}{1} {\\<rparr>}{{$\rrparenthesis$}}{1}
        {\\<leftarrow>}{{$\leftarrow$}}{1} {\\<^sub>\\<O>}{{$_{\mathcal O}$}}{1} {\\<^sub>I}{{$_{\texttt{I}}$}}{1}
        {\\<^sub>G}{{$_{\texttt{G}}$}}{2} {\\<phi>}{{$\varphi$}}{1} {\\<Phi>}{{$\Phi$}}{1} {\\<psi>}{{$\psi$}}{1} {\\<Psi>}{{$\Psi$}}{1}
        {\\<^sub>S}{{$_{\texttt S}$}}{1} {\\<inverse>}{{$^{-1}$}}{1} {\\<^sub>O}{{$_{\texttt O}$}}{1} {\\<^bold>\\<And>}{{$\bm\bigwedge$}}{1}
        {\\<^bold>\\<or>}{{$\bm\lor$}}{1} {\\<^sub>G}{{$_{\texttt G}$}}{1} {\\<Pi>}{{$\Pi$}}{1} {\\<^sub>I}{{$_{\texttt I}$}}{1} {\\<noteq>}{{$\neq$}}{1}
        {\\<bottom>}{{$\bot$}}{1}
        {\\<^sub>+}{{$_\texttt +$}}{1}
        {\\<^bold>\\<and>}{{$\bm\land$}}{1} {\\<^bold>\\<not>}{{$\bm\lnot$}}{1}
        {\\<^sub>1}{{$_\texttt 1$}}{1} {\\<^sub>2}{{$_2$}}{1} {\\<A>}{{$\mathcal A$}}{1} {\\<Turnstile>}{{$\models$}}{2} {\\<^sub>\\<forall>}{{$_\forall$}}{1}
        {\\<^sub>0}{{$_0$}}{1} {\\<tau>}{{$\tau$}}{1}  {\\<^sub>\\<Omega>}{{$_\Omega$}}{1} {\\<^sub>V}{{$_V$}}{1} {\\<^bold>\\<Or>}{{$\bm\bigvee$}}{1}
        {\\<^sub>P}{{$_\texttt P$}}{1}
        {\\<^sub>X}{{$_\texttt X$}}{1}
        {\\<^sub>M}{{$_\texttt M$}}{1}
        {\\<^sub>L}{{$_\texttt L$}}{1}
        {\\<longrightarrow>}{{$\longrightarrow$}}{2}
        {\\<or>}{{$\lor$}}{1}
        {\\<^sub>\\<pi>}{{$_\pi$}}{1}
        {\\<^sub>s}{{$_s$}}{1}
        {\\<^sub>t}{{$_t$}}{1}
        {\\<^sub>a}{{$_a$}}{1}
        {\\<^sub>r}{{$_r$}}{1}
        {\\<^sub>t}{{$_t$}}{1}
        {\\<^sub>e}{{$_e$}}{1}
        {\\<^sub>n}{{$_n$}}{1}
        {\\<^sub>d}{{$_d$}}{1}
        {\\<^sub>i}{{$_i$}}{1}
        {\\<^sub>v}{{$_v$}}{1}
        {\\<^sub>j}{{$_j$}}{1}
        {\\<^sub>b}{{$_\texttt b$}}{1}
        {\\<inter>}{{$\cap$}}{1}
        {\\<union>}{{$\cup$}}{1}
        {\\<Union>}{{$\bigcup$}}{1}
        {\\<^sup>c\\<TTurnstile>\\<^sub>=}{{${}^c\models_=$}}{1}
        {\\<open>}{{<}}{1}
        {\\<close>}{{>}}{1}
        {\\<langle>}{{$\langle$}}{1}
        {\\<rangle>}{{$\rangle$}}{1}
        {\\<ge>}{{$\ge$}}{1}
        {―}{{---}}{1}
        {‹}{{\guilsinglleft}}{1}
        {›}{{\guilsinglright}}{1}
        {⇒}{{$\Rightarrow$}}{2}
        {∀}{{$\forall$}}{1}
        {∈}{{$\in$}}{2}
        {∧}{{$\land$}}{2}
        {∩}{{$\cap$}}{2}
        {⊆}{{$\subseteq$}}{2}
        {∑}{{$\sum$}}{1}
        {≥}{{$\ge$}}{2}
        {⋀}{{$\bigwedge$}}{2}
        {⟹}{{$\Longrightarrow$}}{2}
        {≠}{{$\neq$}}{2}
}

\lstdefinestyle{mystyle}{
    basicstyle=\tiny\ttfamily,
    keywordstyle=\bfseries\color{blue},
    commentstyle=\itshape\color{red},
    stringstyle=\color{green!70!black},
    backgroundcolor=\color{gray!10},
    frame=single,
    numbers=left,
    numberstyle=\ttfamily\tiny\color{gray},
    tabsize=4,
    language=isabelle
}

\newtcblisting[auto counter]{IsabelleSnippet}[2][]{
    , listing only
    , breakable
    , title=Listing \thetcbcounter: #2
    , #1
}

\lstdefinestyle{isainline}{
  language=isabelle,
  basicstyle=%
    \ttfamily\small
}

\usepackage{etoolbox}

\newcounter{listingnum}

\newcommand{\isasnip}[2][]{%
  \refstepcounter{listingnum}%
  \noindent\hrulefill\\%
  \textbf{Listing \thelistingnum: #2}%
  \ifx\relax#1\relax%
  \else
    \label{#1}%
  \fi
  \begin{isabelle}%
}

\newenvironment{isasnipenv}[2][]{%
  \isasnip[#1]{#2}%
}{%
  \end{isabelle}%
  \vspace{-0.8em}
  \noindent\hrulefill%
}
 
\title{Formally Verified Approximate Policy Iteration}

\author{%
    Maximilian Schäffeler\\
    Technische Universität München\\
    Germany
    \And 
    Mohammad Abdulaziz\\
    King’s College London\\
    United Kingdom
}

\begin{document}
\maketitle
\begin{abstract}
We formally verify an algorithm for approximate policy iteration on Factored Markov Decision Processes using the interactive theorem prover Isabelle/HOL.
Next, we show how the formalized algorithm can be refined to an executable, verified implementation. 
The implementation is evaluated on benchmark problems to show its practicability.
As part of the refinement, we develop verified software to certify Linear Programming solutions.
The algorithm builds on a diverse library of formalized mathematics and pushes existing methodologies for interactive theorem provers to the limits.  
We discuss the process of the verification project and the modifications to the algorithm needed for formal verification.
\end{abstract}

\newcommand{\const}[2]{\newcommand{#1}{\textrm{#2}}}
\newcommand{\type}[2]{\newcommand{#1}{\textrm{#2}}}
\newcommand{\key}[2]{\newcommand{#1}{\textbf{#2}}}

\newcommand{\typef}[1]{\textrm{#1}}
\newcommand{\constf}[1]{\textrm{#1}}

\newcommand{\Kzero}{\textsf{P}_\textsf{X}}
\newcommand{\Kstep}{\constf{P}_\textsf{step}}
\newcommand{\Lact}{\constf{L}_\textsf{act}}
\newcommand{\Sstep}{\constf{S}_\textsf{step}}
\newcommand{\measurable}{\mathbin{\rightarrow_M}}
\newcommand{\bindop}{\mathbin{>\!\!\!>\mkern-6.7mu=}}
\newcommand{\tendsto}{\xrightarrow{\hphantom{AAA}}}

\const{\bind}{bind}

\const{\MDPreward}{MDP-reward}
\const{\MDP}{MDP}
\const{\C}{C}

\const{\Ane}{A-ne}
\const{\subprobalgebra}{subprob-algebra}
\const{\policystep}{policy-step}
\const{\policyimprovement}{policy-improvement}
\const{\policyiteration}{policy-iteration}
\const{\streamspace}{stream-space}
\const{\completespace}{complete-space}
\const{\argmaxA}{arg-max}
\const{\hasargmax}{has-arg-max}
\const{\maxLex}{max-L-ex}
\const{\findpolicy}{find-policy}
\const{\vipolicy}{vi-policy}
\const{\conserving}{conserving}
\const{\clog}{log}
\const{\vi}{value-iteration}
\const{\improving}{improving}
\const{\pmf}{pmf}
\const{\isdec}{is-dec}
\const{\ispolicy}{is-policy}
\const{\isdecdet}{is-dec-det}
\const{\mkdecdet}{mk-dec-det}
\const{\mkstationary}{mk-stationary}
\const{\falseA}{False}
\const{\trueA}{True}
\const{\Suc}{Suc}
\const{\snd}{snd}
\const{\probspace}{$\mathcal{P}$}
\const{\tracespace}{$\mathcal{T}$}

\const{\return}{return}
\const{\cempty}{empty}

\const{\prob}{$\mathbb{P}$}
\const{\setpmf}{set-pmf}
\const{\mappmf}{map-pmf}
\const{\mapA}{map}
\const{\returnpmf}{return-pmf}
\const{\cspace}{space}
\const{\sets}{sets}
\const{\indicator}{indicator}
\const{\countspace}{count-space}
\const{\clim}{lim}
\const{\csup}{$\bigsqcup$}
\const{\policies}{$\Pi$}
\const{\vecB}{$V_B$}

\const{\bfun}{bfun}
\const{\bounded}{bounded}
\const{\range}{range}
\const{\undefined}{undefined}
\const{\dist}{$d_\infty$}
\const{\reverse}{reverse}
\const{\IT}{IT}
\const{\norm}{norm}
\const{\borel}{borel}
\const{\distr}{distr}
\const{\univ}{UNIV}
\const{\real}{$\mathbb{R}$}
\const{\ereal}{ereal}
\const{\Up}{Up}
\const{\Right}{Right}
\const{\Down}{Down}
\const{\action}{action}
\const{\Left}{Left}
\const{\cAE}{AE}
\const{\cin}{in}
\const{\X}{X}
\const{\Y}{Y}
\const{\Pt}{$\mathcal{X}$}
\const{\id}{id}
\const{\Trap}{Trap}
\const{\stateA}{state}
\const{\Pos}{Pos}
\newcommand{\LL}{\Q^*}
\newcommand{\LLb}{\mathcal{L}_b}
\newcommand{\Kst}{\textsf{K}_\mathsf{st}}
\newcommand{\EK}{\mathcal{K}_\mathsf{st}}
\newcommand{\pushexp}{\textsf{pushexp}}
\newcommand{\rdec}{\textsf{r}_\mathsf{dec}}
\newcommand{\rb}{\textsf{r}_\mathsf{b}}
\newcommand{\etrfin}{\textsf{etr}_\mathsf{fin}}
\newcommand{\etropt}{\nu^*}
\newcommand{\rM}{\mathsf{r}_\mathsf{M}}
\newcommand{\TT}{\mathcal{T}}
\newcommand{\XX}{\textsf{X}}
\newcommand{\YY}{\textsf{Y}}
\newcommand{\KKzero}{\textsf{P}_\textsf{X}}
\const{\condpmf}{cond-pmf}
\const{\asmarkovian}{as-markovian}
\const{\actstar}{act*}
\const{\finite}{finite}
\const{\isargmax}{is-arg-max}

\newcommand{\pisuc}{\pi\textsf{-Suc}}
\newcommand{\YX}{\textsf{Y}^\textsf{X}}

\key{\klocale}{locale}
\key{\typedef}{typedef}
\key{\datatype}{datatype}
\key{\kfix}{fixes}
\const{\cfix}{fix}
\key{\kand}{and}
\key{\klet}{let}
\key{\kin}{in}
\key{\kif}{if}
\key{\then}{then}
\key{\kelse}{else}
\key{\kdo}{do}
\key{\kcase}{case}
\key{\kof}{of}
\key{\assume}{assumes}
\key{\kshow}{shows}

\type{\boolA}{bool}
\type{\nat}{$\mathbb{N}$}
\type{\set}{set}
\type{\tlist}{list}
\type{\stream}{stream}
\type{\measurepmf}{measure-pmf}
\type{\probalgebra}{prob-algebra}
\type{\vsigma}{vsigma}
\type{\streams}{streams}
\type{\cbind}{bind}
\type{\measure}{measure}
\type{\emeasure}{emeasure}
\type{\metricspace}{metric-space}
\type{\realnormedvector}{real-normed-vector}
\type{\realvector}{real-vector}
\type{\countable}{countable}
\type{\dec}{dec}

\renewcommand{\iff}{\longleftrightarrow}

\newcommand{\ic}[1]{\mathit{#1}}
\newcommand{\iconst}[2]{\newcommand{#1}{\ic{#2}}}

\newcommand{\ilet}{\textbf{let}}
\newcommand{\where}{\textbf{where}}
\newcommand{\iin}{\textbf{in}}
\newcommand{\iif}{\textbf{if}}
\newcommand{\ithen}{\textbf{then}}
\newcommand{\ior}{\textbf{or}}
\newcommand{\ielse}{\textbf{else}}
\iconst{\updpol}{greedy\_\pi}
\iconst{\sortpol}{sort\_\pi}
\iconst{\Ta}{\ic{T}}
\iconst{\bonus}{\delta}
\iconst{\vw}{\ic{v_w}}
\iconst{\states}{X}
\newcommand{\consistent}{\sqsubseteq}
\iconst{\bellmanerr}{factored\_err}
\iconst{\brancherr}{branch\_err}
\iconst{\maxsum}{max_\Sigma}
\iconst{\fold}{\ic{fold}}
\iconst{\dims}{\ic{n}}
\iconst{\elimstep}{max\_step}
\iconst{\iO}{\ic{\mathcal{O}}}
\iconst{\dimset}{\{0\ldots<\dims\}}
\iconst{\map}{map}
\iconst{\fst}{fst}
\iconst{\fn}{fn}
\iconst{\inst}{inst}
\iconst{\cons}{::}
\iconst{\emptymap}{\bot_M}
\iconst{\apistep}{api\_step}
\iconst{\updweights}{upd\_w}
\iconst{\api}{api}
\iconst{\length}{len}
\iconst{\dimh}{m}
\iconst{\branchweights}{branch\_lp}
\iconst{\weightslp}{weight\_lp}
\iconst{\cstates}{X_{(t, ts)}}
\iconst{\R}{\mathbb{R}}
\iconst{\N}{\mathbb{N}}
\iconst{\minlp}{min\_lp}
\newcommand\mdoubleplus{\mathbin{+\mkern-10mu+}}
\iconst{\append}{\cdot}
\iconst{\iset}{set}
\iconst{\doms}{X}
\iconst{\domty}{\alpha}
\iconst{\statety}{X}
\iconst{\reward}{R}
\iconst{\dimr}{r}
\iconst{\dom}{Dom}
\iconst{\rscope}{\mathbf{U}}
\iconst{\scope}{scope}
\iconst{\disc}{\gamma}
\iconst{\transition}{\mathcal{P}}
\newcommand{\tscope}{\Gamma}
\newcommand{\I}{\mathbf{I}}
\iconst{\g}{g}
\iconst{\h}{h}
\iconst{\hscope}{\mathbf{H}}
\iconst{\T}{\mathbf{T}}
\iconst{\Q}{Q}
\iconst{\etr}{\nu}
\iconst{\pol}{\pi}
\iconst{\effects}{effects}
\iconst{\E}{\mathbb{E}}
\iconst{\True}{True}
\newcommand{\actions}{A}
\newcommand{\qsopt}{\mbox{QSopt\_ex}}
\newcommand{\soplex}{\mbox{SoPlex}} %
\section{Introduction}
\label{sec:intro}

Markov Decision Processes (MDPs) are models of probabilistic systems, with applications in AI, model checking, and operations research.
In AI, for instance, given a description of the world in terms of states and actions, where actions can change those states in a randomised fashion, a solution is a \emph{policy} that instructs a decision-maker as to which actions to choose in every state, with the aim of accruing maximum possible \emph{reward}.
There is a large number of methods to solve MDPs, most notably, value and policy iteration, which are some of the oldest and most studied algorithms, and which can compute policies with guarantees on the optimality of those policies.

In many safety-critical applications, e.g.\ in AI or autonomous systems, an important factor of whether a system for solving MDPs can be used is the trustworthiness of that system.
One important aspect here is the assurance that the output of the MDP solving system is correct.
Such assurance can be attained to some degree by testing and other software engineering methods.
However, the best guarantee one can hope is to mathematically prove the correctness of the MDP solver.
A successful way of mathematically proving correctness properties of (i.e.\ formally verify) pieces of software is using Interactive Theorem Provers (ITPs), which are formal mathematical systems that one can use to devise machine-checked proofs.
Indeed, ITPs have been used to prove correctness properties of compilers~\cite{leroy2009formal}, operating systems kernels~\cite{klein2009sel4}, model checkers~\cite{esparza2013fully}, planning systems~\cite{ictai2018,TemporalSemantics,verifiedSATPlan}, and, most related to the topic of this work, algorithms to solve MDPs~\cite{ValIterIsabelle}.
A challenge with using ITPs to prove algorithms correct, nonetheless, is that they require intense human intervention.
Thus for an ITP to be successfully employed in a serious verification effort, novel ideas in the design of the software to be verified as well as the underlying mathematical argument have to be made.

In this paper, we consider formally verifying algorithms for solving \emph{factored} MDPs.
A challenge to using MDPs to model realistic systems is their, in many cases, enormous size.
For such systems, MDPs are succinctly represented as factored MDPs.
The system's state is characterised as an assignment to a set of state variables and actions are represented in a compact way by exploiting the structure present in the system, e.g.\ the fact that each action only changes a small set of state variables.
Such representations are common in AI~\cite{guestrin2003efficient,RDDLref,younes2004ppddl1} and in model checking~\cite{PRISM2006,stormModelChecker}.
Although ITPs have been used prove the correctness of multiple types of software and algorithms, including algorithms on MDPs, algorithms on factored MDPs are particularly challenging. 
The root of this difficulty is that the succinctness of the representation comes at a cost.
Naively finding a solution for a given factored MDP could entail the construction of structures that are exponentially bigger than the factored MDP.
This necessitates the usage of advanced data structures and heuristics, in addition to combining a large number of computational techniques, to avoid that full exponential blow up.

Our main contribution is that we formally verify the algorithm by Guesterin et al.~\cite{guestrin2003efficient} using the Isabelle/HOL~\cite{IsabelleHOLRef} theorem prover.
This algorithm computes approximate policies, i.e.\ sub-optimal policies with guarantees on their optimality, for one type of factored MDPs.
For instance, the algorithm we consider here combines scoped functions and decision lists, which are data structures that exploit factoring of the representation, linear programming, in addition to probabilistic reasoning and dynamic programming. 
The combination of this wide range of mathematical/algorithmic concepts and techniques is what makes this algorithm particularly hard from a verification perspective.
To get an idea of the scale, the reader is advised to look at Fig.~\ref{fig:locales}, which shows the hierarchy of concepts and definitions which we had to develop (aka \emph{formalize}) within Isabelle/HOL just to be able to state the algorithm and its correctness statement.
This is, of course, in addition to the predefined notions of analysis, probabilities, and MDPs, which already exist in Isabelle/HOL.
Furthermore, to be able to prove the algorithm correct, we had to design an architecture of the implementation that allows feasible verification.
This architecture mixes verification and certification: we verified the entire algorithm, except for linear program solving, for which we build a verified a certificate checker.
In addition to proving the algorithm correct, we obtain a formally verified implementation of the algorithm, the we experimentally show to be practical.
Our work, as far as we are aware, is the first work on verifying algorithms for solving factored MDPs.
We believe that our work provides infrastructure as well as methodological insights enabling the verification of other algorithms for factored MDPs, e.g.\ algorithms for planning under uncertainty and probabilistic model checking.
\section{Background}
\label{sec:bg}
We introduce the interactive theorem prover Isabelle/HOL, our formal development of factored MDPs and show how it relates to existing formalizations of MDPs in Isabelle/HOL.

\subsection{Isabelle/HOL}

An interactive theorem prover (ITP) is a program that implements a formal mathematical system,
in which definitions and theorem statements can be expressed, and proofs are constructed from a set of axioms.
To prove a fact in an ITP, the user only provides the high-level steps while the ITP fills in the details at the level of axioms.
Specifically, our developments use the interactive theorem prover Isabelle/HOL~\cite{IsabelleHOLRef} based on Higher-Order Logic, a combination of functional programming with logic.
Isabelle is highly trustworthy, as the basic inference rules are implemented in a small, isolated kernel. 
Outside the kernel, several tools implement proof tactics, data types, recursive functions, etc..

For improved readability, our presentation of definitions and theorems deviates slightly from the formalization in Isabelle/HOL.
Specifically, we use subscript notation for list indexing and some function applications.
We use parentheses for function application opposed to juxtaposition in Isabelle/HOL.
For a list $\ic{xs}$, $\length(\ic{xs})$ returns the length, while $\ic{xs}_{:i}$ returns the first $i$ elements of $\ic{xs}$.
List concatenation is written as $\ic{xs} \append \ic{ys}$, $x~\cons~\ic{xs}$ inserts $x$ at the front of the list $\ic{xs}$, $\map(f, \ic{xs})$ applies $f$ to every element of $\ic{xs}$.
Finally, $\N_{<i}$ is short for the first $i$ natural numbers (including $0$).

\subsection{Factored MDPs}
Factored MDPs are compactly represented MDPs that exploit commonly found structure in large MDPs.
This can often lead to an exponential reduction in the size of the model.
Common formats to store factored MDPs are JANI~\cite{jani}, the PRISM language~\cite{PRISM2006}, and RDDL~\cite{RDDLref}.
We implement the factoring described in~\cite{guestrin2003efficient}.
In Isabelle/HOL, we define factored MDPs using \emph{locales}~\cite{LocalesBallarin} (see Listings~\ref{snip:fmdp}, \ref{snip:fmdpd}).
A locale introduces a mathematical context with constants and assumptions, in which we develop our formalization.
Locales can be instantiated with concrete constants and a proof that discharges the assumptions of the locale, yielding all the theorems proved within the locale.
For example, we instantiate the existing MDP formalization with the factored systems introduced in this section, and are therefore able to reuse important definitions and theorems. 
Moreover, reducing factored MDPs to a tested and reviewed library improves the trust in our definitions.

\paragraph*{State Space}
A state of the MDP is an assignment of values to the $\dims$ state variables.
Each state variable $i \in \N_{<n}$ has a finite, nonempty domain $\doms_i$.
In Isabelle/HOL, we implement a state $x$ of the MDP with a map, i.e. a function with an explicit domain $\dom(x)$ (see Listing \ref{snip:states}).
A \emph{partial state} is a map where only a subset of the state variables are assigned ($\dom(x) \subseteq \N_{<\dims}$), but all entries are valid, i.e. $x_i \in \doms_i$.
The set $\states$ of \emph{states} of the MDP  consists of all partial states $x$ where $\dom(x) = \N_{<\dims}$.
The domain of a state can be restricted to a set $Y$ with the notation $x |_Y$.
One partial state $x$ is called \emph{consistent} with another partial state $t$ (written $x \consistent t$), if and only if $x |_{\dom(t)} = t$. 

\newcommand{\wkn}{\text{W}}
\newcommand{\bkn}{\text{B}}

\paragraph*{Example}
As a running example, we will use a model of a computer network with ring topology from~\citeauthor{guestrin2003efficient}'s original paper (see \autoref{fig:ring_network_connected}).
In the ring, each machine could be either working or broken, and their states change stochastically.
Each machine $C_i$'s state of operation is characterised by a state variable $i$, s.t.\ all domains $\doms_i = \{ \text{\wkn, \bkn} \}$. In a ring with four machines, one valid partial state is $s \coloneqq [ 1 \mapsto \text{\wkn}, 3 \mapsto \text{\bkn} ]$, $\dom(s) = \{1, 3\}$.
It holds that $s \consistent [ 1 \mapsto \text{\wkn}]$, but $s \not\consistent [2 \mapsto \text{\bkn}]$.

\paragraph*{Scoped Functions}

For many MDPs, the transition behavior and the rewards can be computed from a combination of functions that only depend on a few state variables.
Such \emph{scoped functions} take a partial state as input and determine the output inspecting only dimensions that are within their scope.
The algorithm we formalize shows how to express policy evaluation and policy improvement with scoped functions, which avoids enumerating the state space.
Scoped functions pose a challenge for formalization, as scopes could be represented implicitly or explicitly: we can either prove that a given function has a restricted scope, or the function can store its own scope explicitly as data.
In Isabelle/HOL we do both: 
proving that a function has a certain scope only after the function definition is more flexible, as one may derive multiple scopes for a single function (see Listing~\ref{snip:scoped}).
However, since it is infeasible to calculate the scope of a function in general, we use explicit scopes in the executable version of the algorithm.
An important operation on scoped functions is instantiation with a partial state $t$. The operation $\inst_t$ applied to a scoped function returns a new scoped function with reduced scope, where all input dimensions that $t$ provides have been fixed to the values of $t$.

\paragraph*{Transitions}
The MDP provides a finite set of actions $\actions$ with a default action $d \in \actions$.
For each action, transition probabilities $\transition^a_i: \statety \to \prob(\doms_i)$ with $\scope(\transition^a_i) \subseteq\N_{<\dims}$ determine the system's evolution.
The sets $\effects_a$ define the state variables where the behavior of $a$ differs from the default action (i.e.\ variables $i$ s.t.\ $\transition^a_i \neq \transition^d_i$).
The combined transition probabilities, induced by action $a$, between two states $x$ and $x'$ are defined as
$\transition^a(x, x') \coloneqq \prod_{i < \dims} ~ \transition^a_i(x, x'_i)$ (see Listing~\ref{snip:trans}).

\newcommand{\defaultact}{\ensuremath{d}}

\paragraph*{Example}
In the ring topology domain, there is an action to restart each machine.
Restarting a machine guarantees that it will work in the next step. 
The default action \defaultact~ is to do nothing.
The probability that a machine works in the next step depends on its own state and the state of the predecessor. 
This means that e.g. $\scope(\transition^\defaultact_2) = {1, 2}$.
The exact conditional probability distribution for the MDP's evolution under this action is shown in \autoref{fig:ring_network_connected}.
Using an explicit representation to model this factored action, we would need, in the worst-case, $16$ transitions, each of which with a distribution over $16$ possible successor states.
This is in contrast to four tables in the factored case.
A common way to model these scopes is using dependency graphs as shown in \autoref{fig:ring_network_connected}.

\paragraph*{Rewards}
Each action comes with scoped reward functions $\reward^a_i : \statety \to \R$, for $i < r_a$. 
The reward for taking an action is a sum of its reward functions
$\reward^a(x) \coloneqq \textstyle\sum_{i < \dimr_a} \reward^a_i(x).$
We assume that the first $r_d$ reward functions are the same for all actions.
Now, given a policy $\pol : X \to A$ we are interested in the discounted expected total reward
$ \etr_\pol(x) \coloneqq \E_{\omega \sim \mathcal{T}(\pi, x)} \left[ \textstyle\sum_i \disc^i \reward^{\pol(\omega_i)}(\omega_i) \right]$ with discount factor $\disc < 1$ and trace space $\mathcal{T}$.
Our goal is to achieve the optimal reward
$ \etr^*(x) \coloneqq \textstyle\sup_{\pol \in \Pi} \etr_\pi(x)$.
For a value estimate $v : \states \to \R$ and an action $a$, the $Q$ function is defined as
$$\Q^a_v(x) \coloneqq \reward^a(x) + \disc \textstyle\sum_{x' \in \states} \transition^a(x, x') \cdot v(x').$$
For each state, the maximum lookahead w.r.t. all actions is~$\Q^*_v(x)$.
A \emph{greedy policy} $\pi$ is a policy where $\Q^{\pol(x)}_v(x) = \LL_v(x)$ for all states $x$. 
The Bellman error, denoted by $\| v - Q^\pi_v \|$, is the maximum distance between $Q^{\pol(x)}_v(x)$ and $v(x)$ for any state $x \in X$, i.e.\ the $L_\infty$ distance. 

\paragraph*{Example} 
In our example, a factored representation of the rewards of $\defaultact$, $\reward^\defaultact_i$, is $\reward^\defaultact_i(\wkn)=1$ and $\reward^\defaultact_i(\bkn)=0$, for all $1\leq i\leq 4$.
This gives rise to an exponentially smaller reward function in comparison to an explicitly represented MDP, where the reward function for $\defaultact$ would have $16$ entries.

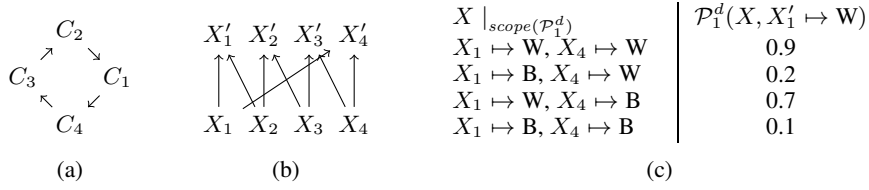
\begin{figure}
  \centering
  \small
  \begin{subfigure}[b]{0.2\textwidth}
    \centering
  \begin{tikzpicture}[scale=0.45]
  \def \n {4}
  \def \radius {1.4cm}
  \foreach \s in {1,...,\n}
  {
    \node[] (\s) at ({360/\n * (\s - 1)}:\radius) {$C_{\s}$};
  }
  \foreach \s in {1,...,\n}
  {
    \pgfmathtruncatemacro{\next}{mod(\s, \n) + 1}
    \draw[->] (\next) -- (\s);
  }
  \end{tikzpicture}
 \caption{}
 \end{subfigure}
  \begin{subfigure}[b]{0.2\textwidth}
    \centering
  \begin{tikzpicture}[scale=0.4]
  \def \n {4}
  \foreach \s in {1,...,\n}
  {
    \node (\s) at (1.5*\s, 0) {$X_{\s}$};
  }
  \foreach \s in {1,...,\n}
  {
    \node (prime\s) at (1.5*\s, 3) {$X_{\s}'$};
  }
  \foreach \s in {1,...,\n}
  {
    \pgfmathtruncatemacro{\next}{mod(\s, \n) + 1}
    \draw[->] (\s) -- (prime\s);
    \draw[->] (\next) -- (prime\s);
  }
  \end{tikzpicture}
  \caption{}
  \end{subfigure}
   \begin{subfigure}[b]{0.5\textwidth}
  \small
  \centering
  \begin{tabularx}{0.85\textwidth}{X|c}
   $X\mid_{\scope(\transition^\defaultact_1)}$ & $\transition^\defaultact_1(X,X_1'\mapsto\wkn)$\\
   $X_1\mapsto\wkn$, $X_4\mapsto\wkn$ & 0.9\\
   $X_1\mapsto\bkn$, $X_4\mapsto\wkn$ & 0.2\\
   $X_1\mapsto\wkn$, $X_4\mapsto\bkn$ & 0.7\\
   $X_1\mapsto\bkn$, $X_4\mapsto\bkn$ & 0.1
  \end{tabularx}

  \caption{}
  \end{subfigure}
\caption{(a) Ring network of 4 connected computers. (b) Variable dependencies of the default action. (c) Probabilities of machine $C_1$ working in the next step, for every state of the machines $C_1$ and $C_4$.}
  \label{fig:ring_network_connected}
\end{figure}

\subsection{Linear Value Functions}
\newcommand{\basis}{\ensuremath{h}}
\renewcommand{\equiv}{\ensuremath{\coloneqq}}
Even if all reward and transition functions are scoped functions, the value function $\etr_\pol$ may still be unstructured, i.e.\ computing $\etr_\pol$ might require the construction of an exponentially big mapping, in the worst case~\cite{guestrin2003efficient}.
However, the value of a policy can be approximated as the weighted sum of $\dimh$ basis functions $\basis_i : \statety \to \R$.
Given a weight $w_i$ for each $\basis_i$, the value of a state can be computed as a weighted sum $\etr_w(x) \coloneqq \textstyle\sum_{i < \dimh} w_i \basis_i(x)$.
Note that the efficiency of the algorithm we verify here crucially depends on the fact that $\basis_i$ is a scoped function, where $\scope(\basis_i)\subset\doms$.

For each action choice $a$ and a basis function $\basis_i$, we can compute its expected evaluation $\g^a_i$ in the successor state.
Since $\g$ does not depend on the weights, it only needs to be computed once for a set of basis functions.
In Isabelle/HOL (see Listing \ref{snip:linval}), we prove that $\g^a_i$ has a structured representation with scope $\Gamma^a_i \equiv \bigcup_{j \in \scope(\basis_i)} \scope(\transition^a_j)$ as follows
  \begin{alignat*}{4}
  \g^a_i(x) \coloneqq& \textstyle\sum {x' \in \states} . ~ \transition^a(x, x') \cdot \h_i(x')
    = \textstyle\sum {x' \in \states |_{\Gamma^a_i}}.~\transition^a(x |_{\Gamma^a_i}, x') \cdot \h_i(x').
  \end{alignat*}
The second equality holds since $\sum{x' \in \states |_{Y}}. ~\transition^a(x, x') = 1$ for any scope $Y$.
This also leads to an efficient computation of the $Q$ functions:
$\Q_w^a(x) \equiv Q^a_{\etr_w}(x) = \reward^a(x) + \disc \textstyle\sum_{i < \dimh} w_i \g^a_i(x)$.

\paragraph*{Example} The value functions in the ring domain can be approximately represented using the basis function $h_0 = 1$ and one function per machine: $\basis_i = 1$ if $X_i = \wkn$ and $0$ otherwise, for $1 \leq i \leq 4$.
Note that because these basis functions have very limited scopes, the accuracy of the best possible approximation of the value function using these basis will be limited.
This may be improved by including more dimensions in the scopes to reflect dependencies between variables.
\section{Approximate Policy Iteration}
Approximate Policy Iteration (API) is a variant of policy iteration that uses basis functions to scale to large systems~\cite{guestrin2003efficient}.
Each iteration of API consists of three parts: policy evaluation, policy improvement and Bellman error computation.
The algorithm terminates when either a timeout~$t_{\ic{max}}$ is reached, the error dips below a threshold~$\epsilon$, or the weights given to the basis functions converge.
In Isabelle/HOL, the algorithm is implemented as a function $\api(t, \pol, w)$ (\autoref{algo:api}, Listing~\ref{snip:api}).
It takes a time $t$, weights for the basis functions $w$ and a policy $\pol$.
The initial call to the algorithm is $\api(0, \pol^0, w^0)$ where $w_0 = 0$ and $\pol^0$ is some greedy policy w.r.t.~$w^0$.
A single iteration of API first uses the current policy $\pol$ to compute updated weights $w'$,
then computes a new greedy policy $\pol'$, and finally determines the Bellman error $\ic{err}$ of the updated policy.
If one of the termination conditions is met, the algorithm returns the current iteration number, weights and policy, as well as the error and whether the weights converged.
Otherwise, $\api$ is called again recursively with updated inputs.
\begin{myalgorithm}[Approximate Policy Iteration]\label{algo:api}
    \makeatletter\@fleqntrue\makeatother
    \begin{alignat*}{2}
        & \api(t, \pol, w) \coloneqq \iif ~ t \ge t_{\ic{max}} ~ \ior ~ \ic{err} \le \epsilon ~ \ior ~ w_=
        ~ \ithen~(t, \pol', w', \ic{err}, w_=)                                   
        ~\ielse~\ic{api}(t+1, \pol', w')                                             \\
              & \quad \begin{alignedat}{4}\where ~
                          & w'       &&  \varassign \updweights(\pol)             \\
                          & \pol'    &&  \varassign \updpol(w')                   \\
                          & \ic{err} &&  \varassign \bellmanerr(\pol', w')        \\
                          & w_=      &&  \varassign w' = w
                      \end{alignedat}%
    \end{alignat*}%
    \makeatletter\@fleqnfalse\makeatother
\end{myalgorithm}

We structure and decouple the algorithm using locales.
Conceptually, this usage of locales is similar to using modules in programming languages.
Here, we use locales to postulate the existence of three functions ($\updweights$, $\updpol$, $\bellmanerr$) along with their specifications:
\begin{specification}[$\updweights$]
    \label{spec:updweights}
    A decision list policy a list representation of policies where each entry (called branch) is a pair $(t, a)$ of a partial state and an action.
To select an action in a state $x$, we search the list for the first branch where $x$ is consistent with $t$.
    Now fix a decision list policy $\pol$, let $w' = \updweights(\pol)$. Then $\etr_{w'}$ is the best possible estimate of $\etr_\pol$:
    $ \| \etr_{w'} - \etr_\pol \| = \textstyle\inf_w \| \etr_w - \etr_\pol \|. $
\end{specification}
\begin{specification}[$\updpol$]
    \label{spec:updpol}
    For all weights $w$, $\updpol(w)$ is a greedy decision list policy for $\etr_{w}$.
\end{specification}
\begin{specification}[$\bellmanerr$]
    \label{spec:bellmanerr}
    Given weights $w$ and a greedy decision list policy $\pol$ for $\etr_{w}$,
    $\bellmanerr$ determines the Bellman error: $\bellmanerr(\pol, w) = \| \LL_w - \etr_w \|$.
\end{specification}
For now, we merely state the specifications for the algorithms approximate policy iteration builds upon,
only later will we show how to implement the specifications concretely.
Viewed differently, the specifications serve as inefficient or even non-executable implementations of algorithms.
This approach keeps the assumptions on individual parts of the algorithm explicit and
permits an easier exchange of implementations,
e.g. in our developments one may swap the LP certification algorithm for a verified LP solver implementation.
It also facilitates gradual verification of software: the correct behavior of the software system can be proved top-down starting from assumptions on each component.
Finally, the small number of axioms improves performance of automated proof methods in ITPs. 
In the following sections we show that these specifications have efficient implementations.
See~\autoref{fig:locales} for an overview of all components.

Within the context of the locale, assuming all specifications, we can derive the same error bounds as presented in \cite{guestrin2003efficient} (see Listing~\ref{snip:api}).
One exemplary important result is that if the weights converge during API, then in the last step the bellman error is equal to the approximation error.
This observation leads to the following \emph{a posteriori} optimality bound:
\begin{theorem}
    \label{thm:errbound}
    Let $api(w_0, \pol_0) = (t', \pol, w, \ic{err}, \True)$. Then
    $(1 - \disc) \| \etr^* - \etr_{w} \| \le 2\disc\cdot\ic{err}$.
\end{theorem}
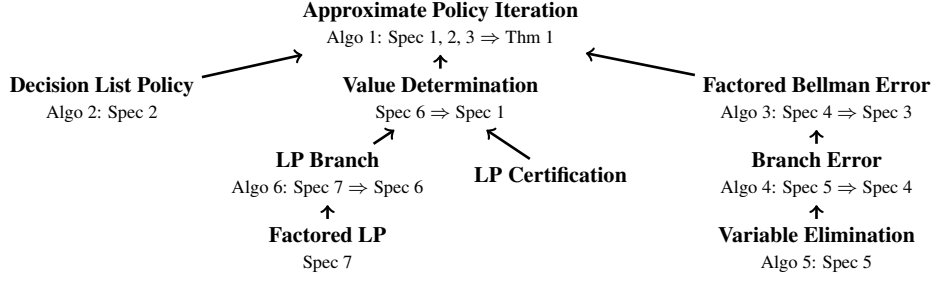
\begin{figure*}
    \smaller
    \centering
    \begin{tikzpicture}[
            every node/.style={align=center},
            level/.style={level distance=1.0cm},
            scale=1,
            edge from parent path={[<-,line width=1pt](\tikzparentnode) -- (\tikzchildnode)}
        ]

            \node (API) {
        \textbf{Approximate Policy Iteration}\\
        \smaller Algo \ref{algo:api}:
        Spec \ref{spec:updweights}, \ref{spec:updpol}, \ref{spec:bellmanerr} 
        $\Rightarrow$
        Thm \ref{thm:errbound}
        }
                child[sibling distance = 4.5cm] {node (DLP) {
                    \textbf{Decision List Policy}\\
        \smaller Algo \ref{algo:dec_pol}:
        Spec \ref{spec:updpol}
                    }}
                child[sibling distance = 8cm] {node [sibling distance=9cm] (VD) {
                    \textbf{Value Determination}\\
        \smaller 
        Spec \ref{spec:branchweights} 
        $\Rightarrow$
        Spec \ref{spec:updweights}
                    }
                        child[sibling distance = 3cm] {node (LPB) {
                            \textbf{LP Branch}\\
        \smaller Algo \ref{algo:branchlp}:
        Spec \ref{spec:minlp}
        $\Rightarrow$
        Spec \ref{spec:branchweights}
                            }
                                child {node (FLP) {
                                    \textbf{Factored LP}\\
        \smaller
        Spec \ref{spec:minlp} 
                                    }}}
                        child[sibling distance = 3cm] {node (LPC) {
                            \textbf{LP Certification}
                            }}
                    }
                child[sibling distance = 5cm] {node (FBE) {
                    \textbf{Factored Bellman Error}\\
        \smaller Algo \ref{algo:ferr}:
        Spec \ref{spec:brancherr} 
        $\Rightarrow$
        Spec \ref{spec:bellmanerr}
                    }
                        child {node (LPB) {
                            \textbf{Branch Error}\\
        \smaller Algo \ref{algo:brancherr}:
        Spec \ref{spec:maxsum} 
        $\Rightarrow$
        Spec \ref{spec:brancherr}
                            }
                                child {node (FLP) {
                                    \textbf{Variable Elimination}\\
        \smaller Algo \ref{algo:varelim}: Spec \ref{spec:maxsum}
                                    }}}}
        ;

    \end{tikzpicture}
    \caption{\label{fig:locales} Simplified Locale Hierarchy}
\end{figure*} \section{Policy Improvement}
Given weights $w$, the policy improvement phase determines a greedy policy w.r.t. $\nu_w$.
The policy takes the form of a decision list, so each element of the list stores a partial state and an action.
The main idea for an efficient computation is that only actions better than the default action need to be considered.
This notion is made precise by the bonus function~$\delta_a$ (\autoref{algo:dec_pol}, Listing~\ref{snip:decpoldef}) with scope~$\T_a$.
Originally, $\tscope_d$ was not part of $\T_a$\cite{guestrin2003efficient}, which
we assume to be an oversight in the definition, since the behavior of the default action does influence the bonus.
Unless components cancel out, the scope of a function difference is the union of the scopes of both arguments.  
Finally, we concatenate the branches $\pol_a$ for every action but $d$, add the default action as a fallback and sort the decision list policy by decreasing bonus.
Here, the empty map with no entries is $\bot_M$, and $\dom(\bot_M) = \emptyset$.
We can show that $\updpol$ satisfies \autoref{spec:updpol}, since action selection proceeds by decreasing bonus.

\begin{myalgorithm}[Decision List Policy]
    \label{algo:dec_pol}
    \makeatletter
    \@fleqntrue
    \makeatother
    \begin{alignat*}{4}
              & \updpol \varassign
              \sortpol((\bot_M, d, 0)~\cons~\ic{concat}([\pol_a ~|~ a \in A - \{d\}])\\
              & \quad \begin{alignedat}{3}
                \where 
~ \pol_a &\varassign [(x, a, \bonus_a(x)) \mid \bonus_a(x) > 0, x \in X_{\mid\T_a}]
&\quad&\text{(Positive bonus, states with domain $\T_a$)}\\
                          \bonus_a &\coloneqq Q^a_w - Q^d_w
                          &\quad&\text{(Bonus function)}\\
                          \T_a &\varassign
                          \scope(R^a) \cup \textstyle\bigcup_{i \in \I_a} \tscope^a_i \cup \tscope^d_i &\quad &\text{(Scope of the bonus function)}\\
                          \I_a &\varassign \{ i < m \mid \effects_a \cap \scope(\basis_i) \neq \emptyset \} &\qquad &\text{(Basis functions relevant to the action)}
                      \end{alignedat}
    \end{alignat*}
    \makeatletter
    \@fleqnfalse
    \makeatother
\end{myalgorithm}

\section{Factored Bellman Error}
The Bellman error $\| \LL_w - \etr_w \|$ is an indicator of the degree of optimality of a policy.
An inefficient computation would enumerate every state, and return the maximum error.
However, for a decision list policy, we can compute the error incurred by each branch separately.
The total error then equals the maximum error of any branch (\autoref{algo:ferr}, Listing~\ref{snip:err}).
For now, assume that we have a function $\brancherr$ that computes the error for a single branch, i.e. the maximum error for any state that selects the respective branch.
These are all states that are consistent with the current branch $t$, but did not match any prior branch $t' \in \ic{ts}$ of the policy, formally
         $\cstates \coloneqq \{ x \in \states.~t \consistent x \land \forall t' \in \ic{ts}.~t' \not\consistent x \}$.
Hence we also need to past the prefix of the decision list policy to the function that computes the error of a branch.
Note that if the branch is selected by no state its error is $-\infty$.
We show that if \autoref{spec:brancherr} is met and $\pol$ is a greedy policy w.r.t. $w$, then $\bellmanerr$ satisfies \autoref{spec:bellmanerr}~(see Listing~\ref{snip:errprop}).
\begin{myalgorithm}[Factored Bellman Error]
    \label{algo:ferr}
        $\bellmanerr \coloneqq \textstyle\sup_{i < \length(p)} \brancherr(p_i, \map(\fst, p_{:i}))$
\end{myalgorithm}
\begin{specification}[$\brancherr$]
    \label{spec:brancherr}
    Given a prefix of a policy $\pi$ with the current branch $(t, a)$ and a list of partial states $\ic{ts}$ from prior branches, $\brancherr(t, a, \ic{ts}) =
        \textstyle\sup_{x \in \cstates} | Q^a_w(x) - \etr_w(x)|$.
\end{specification}

\paragraph{Branch Error}
Take a branch, $(t, a)$, of the policy, for a partial state $t$ an action $a$.
The states of all prior branches form the list of partial states $\ic{ts}$.
To find the Bellman error of the current branch, we need to maximize $|\Q^a_w(x) - \etr_w(x)|$ w.r.t. states $x \in \cstates$.
Note that for all states $x$
\begin{alignat}{4}
    \label{eqn:errbr}
          Q^a_w(x) - \etr_w(x) = & \textstyle\sum_{i < r_a} \reward^a_i(x) + \textstyle\sum_{i < \dimh} w_i(h_i - \disc g_i)(x) \nonumber               \\
    {}={} & \textstyle\sum f \in [ \reward^a_0, \ldots, \reward^a_{\dimr_a}, w_0(h_0 - \disc g_0), \ldots, 
     w_m(h_m - \disc g_m)].~f(x). 
\end{alignat}

If we had an algorithm to efficiently compute the maximum sum of scoped functions, we could determine the error of a branch. 
Again, we specify an algorithm $\maxsum$ to do exactly that (\autoref{spec:maxsum}).
In \autoref{algo:brancherr} (see Listing~\ref{snip:brancherr}), we call $\maxsum$ with the functions from \autoref{eqn:errbr}.
To restrict the maximization to the states $\cstates$, we instantiate all functions with the partial state $t$. 
Additionally, we define the functions $\mathcal{I}'$ that evaluate to $-\infty$ on states that select a different branch of the policy.
Hence, these states are ignored in the error computation.
We also apply $\maxsum$ to the negated functions to compute the absolute value of the error.
Finally, we formally prove that $\brancherr$ satisfies \autoref{spec:brancherr}.

\begin{specification}[Variable Elimination]
    \label{spec:maxsum}
    For scoped functions $\ic{fs}$, $\maxsum(\ic{fs}) = \sup_{x \in \states} \textstyle\sum_{f \in \ic{fs}} f(x)$.
\end{specification}
\begin{myalgorithm}[Branch Error]
    \label{algo:brancherr}
    \makeatletter\@fleqntrue\makeatother
    \begin{alignat*}{2}
        & \brancherr \coloneqq
        \max(\maxsum(\ic{fs}~\append~\mathcal{I}'),
        \maxsum(-\ic{fs}~\append~\mathcal{I}'))
                                                       \\
              & \quad \begin{alignedat}{3}\where~
                          &\ic{rs} &&\coloneqq [ \reward^a_0, \ldots, \reward^a_{\dimr_a} ]               \\
                          &\ic{ws} &&\coloneqq [ w_0 (\h_0 - \disc \g_0), \ldots, w_\dimh (\h_\dimh - \disc \g_\dimh)]    \\
                          &\ic{fs} &&\coloneqq \map(\inst_t , \ic{rs}~\append~\ic{ws})
                      \end{alignedat}
    \end{alignat*}
    \makeatletter
    \@fleqnfalse
    \makeatother
\end{myalgorithm}

\paragraph{Variable Elimination}
The specification for $\maxsum$ can be efficiently implemented with a variable elimination algorithm~(\autoref{algo:varelim}, Listing~\ref{snip:varelim}).
In each iteration, the algorithm selects a dimension of the state space, collects all functions that depend on this dimension in a set $E$. 
It then creates a new function $e$ that maximizes all functions in $E$ over that dimension of the state space.
The algorithm keeps track of a set of functions to maximize and the number of the current iteration.
Since the number of operations performed by the algorithm varies greatly with the elimination order, the variables can be reordered with a bijection $\iO : \N_{<\dims} \to \N_{<\dims}$.
We formally prove that at any point during execution, the maximum of the current list of functions stays constant, thus $\maxsum$ meets~\autoref{spec:maxsum}.

\begin{myalgorithm}[Variable Elimination]%
    \label{algo:varelim}%
    \makeatletter\@fleqntrue\makeatother%
    \begin{alignat*}{2}
        & \elimstep(i, \ic{fs}) \coloneqq (i + 1, e~\cons~E')\\
              & \quad \begin{alignedat}{6}\where~~
                          & (E, E') \coloneqq
                          \ic{partition}(f \mapsto \iO(i) \in \ic{scope}(f), \ic{fs}) \ \ \text{(Partition based on $\iO(i) \in \scope(f)$?)}                             \\
                          & e \coloneqq (x \mapsto \textstyle\max_{y \in \doms_{\iO(i)}} \textstyle\sum_{f \in E} f(x_{\iO(i) \mapsto y}), \textstyle\bigcup_{f \in E} \ic{scope}(f) - \{ \iO(i) \})
                      \end{alignedat} \\[5pt]
        & \maxsum(\ic{fs}) \coloneqq \textstyle\sum_{f \in \ic{fs'}} f(\emptymap)\\
              & \quad \begin{alignedat}{6}
                          \where ~&(\_, \ic{fs'}) \coloneqq \elimstep^{\dims}(0, \ic{fs})
                      \end{alignedat}
    \end{alignat*}
    \makeatletter\@fleqnfalse\makeatother
\end{myalgorithm}

\section{Value Determination}
After a new candidate policy is found, it is evaluated in the value determination phase.
Since the exact value function cannot be represented in general as a linear combination of the basis functions, the aim is to find weights for the basis functions with the smallest approximation error (see \autoref{spec:updweights}).
This problem can be expressed using linear programming (LP).
The structure of the algorithm that finds optimal weights is analogous to the factored Bellman error computation.
For each branch of the policy, we generate a set of LP constraints (according to \autoref{spec:branchweights}, see Listing~\ref{snip:valdet}) that expresses the approximation error incurred by this branch.
The union of all constraints is then
\begin{alignat*}{2}
     & \weightslp \coloneqq \textstyle\bigcup_{i < \length(p)} \branchweights(p_i,\map(\fst, p_{:i})).
\end{alignat*}
Variables of the LP are the approximation error $\phi$ and the weights $w$.
The LP is optimized for minimal error $\phi$, the values of the variables $w$ in an optimal solution determine the new weights.
Given a set of LP constraints $\ic{cs}$, $\langle \ic{cs} \rangle_{\ic{LP}}$ denotes the set of feasible solutions.
We show that for any optimal solution $(\phi^*, w^*) \in \langle \weightslp \rangle_{\ic{LP}}$, setting $\updweights \coloneqq w^*$ satisfies \autoref{spec:updweights}.
We also formally prove that the LP always has an optimal solution, since the set of potentially optimal solutions is compact.

\begin{specification}[$\branchweights$]
    \label{spec:branchweights}
    Given a partial state $t$, an action $a$, a list of partial states $\ic{ts}$, $\branchweights$ constructs an LP that minimizes the approximation error over the states~$\cstates$:
    \begin{alignat*}{2}
         & (\phi, w) \in \langle\branchweights(t, a, ts)\rangle_{\ic{LP}} \iff \forall x \in \cstates. ~ \phi \ge | \Q^a_w(x) - \etr_w(x) |.
    \end{alignat*}
\end{specification}

\paragraph{LPs for Branches}
For each branch of the policy, we proceed similarly to the Bellman error computation:
we create two constraint sets, for positive and negative errors respectively~(\autoref{algo:branchlp}, Listing~\ref{snip:valdetbranch}).
We omit the scopes for brevity.
At this level, we make use of another algorithm $\minlp$. 
Its first input $C$ is a list of $m$ scoped functions, the second input $b$ is another list of scoped functions. Now $\minlp(C, b)$ creates an LP that minimizes $Cw - b$ w.r.t. $w$ over all states~ (see \autoref{spec:minlp}).
We can show that $\branchweights$ fulfills \autoref{spec:branchweights}, since the feasible solutions are upper bounds of the smallest approximation error.
The derivation of $C$ and $b$ follows the Bellman error computation. Interested readers should consult the formal proofs.

\begin{myalgorithm}[Branch LP]
    \label{algo:branchlp}
    \makeatletter\@fleqntrue\makeatother%
    \begin{alignat*}{6}
              & \branchweights \coloneqq \minlp(C, -b~\append~\mathcal{I}') \cup
              \minlp(-C, b~\append~\mathcal{I}')                                    \\
              & \quad \begin{alignedat}{4} \where~
                & b &&\coloneqq map(\inst_t, [\reward^a_i ~|~ i < \dimr_a]) &&
                \hfill\qquad\qquad\qquad\qquad\qquad\qquad\text{($Cw - b~\widehat{=}~\Q^a_w - \etr_w$)}    \\
                & C &&\coloneqq map(\inst_t, [h_i - \disc g_i^a ~|~ i < \dimh])
              \end{alignedat}
    \end{alignat*}
    \makeatletter\@fleqnfalse\makeatother%
\end{myalgorithm}

\begin{specification}[$\minlp$]
    \label{spec:minlp}
    $\minlp(C, b)$ generates an LP that minimizes $Cw - b$:
    \begin{alignat*}{2}
         & (\phi, w) \in \langle\minlp(C, b)\rangle_{\ic{LP}} \iff \forall x \in \states.~\phi \ge \textstyle\sum_{i < \length(C)} w_iC_i(x) + \textstyle\sum_{i < \length(b)} b_i(x).
    \end{alignat*}
\end{specification}

\paragraph{Factored LP Construction}
The algorithm $\minlp$ resembles $\maxsum$, so we only point out the challenges encountered during verification.
Full details can be found in the formalization (see Listing~\ref{snip:flp}).
There are two significant modifications we made to the algorithm to make verification feasible.
First, the original algorithm may create equality constraints that constrain variables to $-\infty$.
Since these constraints are not supported by the LP solvers we use, we formally prove that one can modify the algorithm to omit such constraints.
Second, the combination of LP constraints in the definition of $\weightslp$ requires some care,
to avoid interactions between LP variables created in different branches. 
The $\minlp$ algorithm creates new (private) LP variables, we need to make sure that the variables have distinct names for each branch.
This issue was not discussed by~\citeauthor{guestrin2003efficient}.
The problem can be solved by adding a tag to each generated variable.
The tags contain $t$, $a$, and a flag to differentiate the two invocations of $\minlp$ in each branch. For distinct tags $p$, $p'$ we can then show:
the solutions to the union of two constraint sets are equivalent to the intersection of the solution spaces of the individual constraint sets (concerning $\phi$ and $w$):
\begin{alignat*}{2}
     & \langle\minlp(p, C, b) \cup \minlp(p', C', b')\rangle_{\ic{LP}} = \langle\minlp(p, C, b)\rangle_{\ic{LP}} \cap \langle\minlp(p', C', b')\rangle_{\ic{LP}}.
\end{alignat*}
We show that $\minlp$ creates an LP that is equivalent to the explicit (potentially exponentially larger) LP that creates a constraint for each state of the MDP.
It immediately follows that $\minlp$ satisfies \autoref{spec:minlp}.
This completes the correctness proof of the abstract version of approximate policy iteration.
Due to our approach to the verification of algorithms using loosely coupled locales,
$\minlp$ and $\maxsum$ are not tied to MDPs and are thereby reusable components.

\section{Code Generation}
We now discuss the process of deriving a verified efficiently executable version from the verified abstract algorithm discussed above.
To do so, we follow the methodology of program refinement~\cite{refinementWirth}, where one starts with an abstract, potentially unexecutable, version of the algorithm and verifies it.
Then one devises more optimised versions of the algorithm, and only proves the optimizations correct in this latter step, thus separating mathematical reasoning from implementation specific reasoning.
This approach was used in most successful algorithm verification efforts~\cite{klein2009sel4,esparza2013fully,DBLP:conf/cav/KanavL014}.
In this work, the three most important stages are the initial abstract algorithm, an implementation with abstract data structures, and finally an implementation with concretized data structures.
As a last step, we export verified code in the programming language Scala for the approximate policy iteration algorithm.%

\subsection{Refinement using Locales}
Our implementation of step-wise refinement is based on Isabelle/HOL locales.
For each locale of the abstract algorithm, we define a corresponding locale where we define the executable version of the algorithm.
Finally, we relate the abstract version of the MDP to the concrete version.
For each definition, we then show that corresponding inputs lead to corresponding outputs, i.e. our abstract algorithm and the implementation behave the same.
At this point in the refinement, data structures remain abstract APIs, with the concrete implementations chosen only later.
We use the data structures provided by the Isabelle Collections Framework~\cite{lammichContainers} for code generation, that we extend with our own data structure for scoped functions, represented as a pair of a function and a set for its scope.
The data structure also provides an operation to evaluate a function on all of its scope for memoization.

\subsection{Certification of Linear Programming Solutions}
\begin{wraptable}{r}{65mm}
    \footnotesize
\centering
\begin{tabular}{@{}rrrrr@{}}
\toprule
\textbf{Nodes} & \textbf{Constrs} & \textbf{Vars} & $\mathbf{t (s)}$ & $\mathbf{t_{LP}(s)}$ \\ \midrule
1   & 74     & 41     & 0.27     & 0.02 \\
3   & 1258     & 693  & 0.89    & 0.15  \\
5   & 4378     & 2455     & 1.89     & 0.46  \\
7   & 9418     & 5305     & 3.78     & 0.98  \\
9   & 16378     & 9243     & 6.74     & 1.82  \\
11   & 25258     & 14269     & 12.44     & 3.36  \\
13   & 36058     & 20383     & 20.95     & 5.31 \\
15   & 48778     & 27585     & 34.69     & 8.67  \\
17   & 63418     & 35875     & 58.30     & 16.86  \\
19   & 79978     & 45253     & 92.19     & 30.25  \\
\end{tabular}
\caption{Evaluation on the ring domain, the columns denote the number of variables (number of states grows exponentially), the number of LP constraints and variables generated in the last iteration.
The last two columns give tlhe total running time and the time spent in the LP solver.}
\label{tab:performance}
\end{wraptable} %
An implementation of approximate policy iteration depends on efficient LP solvers. 
In our verified implementation, we use precise but unverified LP solvers and certify their results.
This avoids implementing a verified, optimized LP solver but retains formal guarantees.
The tradeoff here is that the unverified LP solver may now return solutions that cannot be certified.
At the cost of performance, it would also be possible to connect the formalization to an existing executable simplex implementation for Isabelle/HOL~\cite{IsabelleSimplex}.
To achieve formal guarantees, the LP has to be solved exactly, i.e. using rational numbers. 
Two potential candidates for precise LP solvers are \qsopt~\cite{qsoptex} and \soplex~\cite{soplex}.
For larger LPs, the performance of \soplex\ was more reliable in our setting.

We certify optimality using the dual solution and the strong duality of linear programming.
In our formalization we also formally prove that infeasibility and unboundedness can be certified similarly using farkas certificates or unbounded rays. 
The linear program is first preprocessed to a standard form: variable bounds and equality constraints are reduced to only inequality constraints. 
The constraints of the resulting LP are in the standard form $Ax \le b$, with no restrictions on $x$.
During benchmarking of the certification process, the normalization operation usually applied to rationals after each operation proved to be very costly. 
For certification, we represent rational numbers as pairs that are never normalized, which leads to faster certificate checking in our experiments.

The exported Scala program takes an arbitrary function from linear programs to their solutions as input.
If this function returns invalid solutions, they are rejected by the certificate checker, so there are no assumptions we need to place on the LP solver.
However, there is the implicit assumption that the LP solver is deterministic.
Since we are working in a fragment of a functional programming language, calling the LP solver twice on the same problem should lead to the same solution.
In theory, a nondeterministic LP solver could be misused to lead to inconsistencies.
As we do not compare LP solutions in our algorithm and SoPlex is actually deterministic, this problem does not impact our verified software.
A possible more general solution to the problem could be the use of memoization or to explicitly model the nondeterminism using monads.
\subsection{Experimental Evaluation}
We show the practicality of our verified implementation by applying it to the ring topology domain.
The run our implementation on an Intel i7-11800H CPU and we set the discount factor to $0.9$ in all our experiments.
We do not reorder variables.
In all runs, the weights converged after 5 iterations.
The results of the experiments (\autoref{tab:performance}, for full table see \autoref{tab:performancefull}) show that the algorithm can deal with systems of half a million states and 20 actions.
For larger systems, the precise mode of the LP solver Soplex cannot find a rational solution.
The experiment shows that our implementation of linear programming certification can process linear programs with tens of thousands of constraints. %
\section{Discussion}
Our work combines and integrates a wide range of tools and formalization efforts to produce a verified implementation of approximate policy iteration.
In Isabelle/HOL, we build on the formal libraries for LPs~\cite{LPDualityIsabelle}, MDPs~\cite{holzl2017markov}, linear algebra, analysis~\cite{holzl2013type}, probability theory~\cite{johannesMeasure}, and the collections framework~\cite{lammichContainers}.
Furthermore, we certify the results computed by precise LP solvers.
In total, our development is comprised of approximately 20,000 lines of code, two thirds of which are concerning code generation.
We show how to facilitate locales, powerful automation in Isabelle/HOL, and certification to develop complex formally verified software.
We also make the case that the process of algorithm verification provides a detailed understanding of the algorithm: all hidden assumptions are made explicit, while we modularize the algorithm into components with precisely specified behaviour.

The methodology presented here can be applied to a large number of algorithms since the algorithm we verified is a seminal algorithm for solving factored MDPs, combining a large number of concepts that are widely used in many contexts, like AI~\cite{sannerFactoredMDPs,RLFactoredMDPs,RLFactoredMDPsII} and model checking~\cite{PRISM2006,stormModelChecker}.

Applications of interactive theorem provers in verification and formalizing mathematics have been recently attracting a lot of attention~\cite{AvigadHarissonWhyFormal,avigadFormalMaths,MassotWhyFormal}.
In most applications, especially in computer science~\cite{esparza2013fully,klein2009sel4} and AI~\cite{CoqNNAAAI2019,SelsamNNVerification}, the emphasis is on the difficulty of the proofs, whether that is due to many cases or complex constructions, etc., and how theorem proving helped find mistakes or find missing cases in the proofs.
A distinct feature of this work is that its complexity comes from the large number of concepts it combines, shown in Fig.~\ref{fig:locales}, which is a more prevalent issue in formalizing pure mathematics.
Our project has contributed a better restructuring of the algorithm and untangling of the different concepts, leading to better understandability.

Multiple directions can be considered to extend and build on our work.
An alternative to using exact LP solvers is to use their highly optimized floating point counterparts that do not calculate exact solutions.
In this case, one needs to certify the error bound derived from a dual solution to the LP.
Then the error bound has to be incorporated in the error analysis of the algorithm to obtain formal guarantees.
Moreover, we may initialize the algorithm with weights computed by an unverified, floating point implementation of API to reduce number of iterations performed by the verified implementation.
In Isabelle/HOL, some of the correctness proofs for code generation can be automated using tools to transfer theorems.
Furthermore, the ergonomics for instantiating locales and inheriting from other locales could be improved for situations with many locale parameters.

\paragraph{Related Work}
Several formal treatments of MDPs have been developed in the theorem provers Isabelle/HOL~\cite{holzl2017markov,ValIterIsabelle,DBLP:journals/corr/abs-2112-05996,SCCsMDPs} and Coq~\cite{VajjhaSTPF21}.
All developments verify algorithms for explicitly represented MDPs, which limits their practical applicability to solve large MDPs.
We adapt and integrate the implementation of~\citeauthor{ValIterIsabelle} with our formalization.
The algorithm we verified in Isabelle/HOL was first presented by~\cite{guestrin2003efficient}.
An elementary approach to Linear Programming certification certification with Isabelle/HOL has been done before as part of the Flyspeck project \cite{LPsCertIsabelle}, where the feasibility of a solution to a linear program is checked by Isabelle/HOL.
The work presents a method that uses dual solutions produced by floating-point LP solvers to find bounds on the objective value of an LP.
The tool Marabou for neural network verfication uses Farkas vectors to produce proofs for its results~\cite{NNVerificationFarkas}.

\bibliographystyle{abbrvnat}
\bibliography{short_paper}

\begin{thebibliography}{38}
\providecommand{\natexlab}[1]{#1}
\providecommand{\url}[1]{\texttt{#1}}
\expandafter\ifx\csname urlstyle\endcsname\relax
  \providecommand{\doi}[1]{doi: #1}\else
  \providecommand{\doi}{doi: \begingroup \urlstyle{rm}\Url}\fi

\bibitem[Abdulaziz and Koller(2022)]{TemporalSemantics}
M.~Abdulaziz and L.~Koller.
\newblock Formal semantics and formally verified validation for temporal
  planning.
\newblock In \emph{The 36th {{AAAI Conference}} on {{Artificial Intelligence}}
  ({{AAAI}})}, 2022.
\newblock \doi{10.1609/aaai.v36i9.21197}.

\bibitem[Abdulaziz and Kurz(2023)]{verifiedSATPlan}
M.~Abdulaziz and F.~Kurz.
\newblock Formally verified {{SAT-Based AI}} planning.
\newblock In \emph{The 37th {{AAAI Conference}} on {{Artificial Intelligence}}
  ({{AAAI}})}, 2023.
\newblock \doi{10.1609/aaai.v37i12.26714}.

\bibitem[Abdulaziz and Lammich(2018)]{ictai2018}
M.~Abdulaziz and P.~Lammich.
\newblock A {{Formally Verified Validator}} for {{Classical Planning Problems}}
  and {{Solutions}}.
\newblock In \emph{The 30th {{International Conference}} on {{Tools}} with
  {{Artificial Intelligence}} ({{ICTAI}})}, 2018.
\newblock \doi{10.1109/ICTAI.2018.00079}.

\bibitem[Applegate et~al.(2007)Applegate, Cook, Dash, and Espinoza]{qsoptex}
D.~L. Applegate, W.~Cook, S.~Dash, and D.~G. Espinoza.
\newblock Exact solutions to linear programming problems.
\newblock \emph{Operations Research Letters}, Nov. 2007.
\newblock ISSN 01676377.
\newblock \doi{10.1016/j.orl.2006.12.010}.

\bibitem[Avigad(2023)]{avigadFormalMaths}
J.~Avigad.
\newblock Mathematics and the formal turn, Nov. 2023.

\bibitem[Avigad and Harrison(2014)]{AvigadHarissonWhyFormal}
J.~Avigad and J.~Harrison.
\newblock Formally verified mathematics.
\newblock \emph{Commun. ACM}, 2014.
\newblock \doi{10.1145/2591012}.

\bibitem[Bagnall and Stewart(2019)]{CoqNNAAAI2019}
A.~Bagnall and G.~Stewart.
\newblock Certifying the {{True Error}}: {{Machine Learning}} in {{Coq}} with
  {{Verified Generalization Guarantees}}.
\newblock In \emph{The 33rd {{AAAI Conference}} on {{Artificial Intelligence}}
  ({{AAAI}})}, 2019.
\newblock \doi{10.1609/aaai.v33i01.33012662}.

\bibitem[Ballarin(2014)]{LocalesBallarin}
C.~Ballarin.
\newblock Locales: {{A Module System}} for {{Mathematical Theories}}.
\newblock \emph{J. Autom. Reason.}, 2014.
\newblock \doi{10.1007/s10817-013-9284-7}.

\bibitem[Bestuzheva et~al.(2023)Bestuzheva, Besan{\c c}on, Chen, Chmiela,
  Donkiewicz, {van Doornmalen}, Eifler, Gaul, Gamrath, Gleixner, Gottwald,
  Graczyk, Halbig, Hoen, Hojny, {van der Hulst}, Koch, L{\"u}bbecke, Maher,
  Matter, M{\"u}hmer, M{\"u}ller, Pfetsch, Rehfeldt, Schlein, Schl{\"o}sser,
  Serrano, Shinano, Sofranac, Turner, Vigerske, Wegscheider, Wellner, Weninger,
  and Witzig]{soplex}
K.~Bestuzheva, M.~Besan{\c c}on, W.-K. Chen, A.~Chmiela, T.~Donkiewicz, J.~{van
  Doornmalen}, L.~Eifler, O.~Gaul, G.~Gamrath, A.~Gleixner, L.~Gottwald,
  C.~Graczyk, K.~Halbig, A.~Hoen, C.~Hojny, R.~{van der Hulst}, T.~Koch,
  M.~L{\"u}bbecke, S.~J. Maher, F.~Matter, E.~M{\"u}hmer, B.~M{\"u}ller, M.~E.
  Pfetsch, D.~Rehfeldt, S.~Schlein, F.~Schl{\"o}sser, F.~Serrano, Y.~Shinano,
  B.~Sofranac, M.~Turner, S.~Vigerske, F.~Wegscheider, P.~Wellner, D.~Weninger,
  and J.~Witzig.
\newblock Enabling {{Research}} through the {{SCIP Optimization Suite}} 8.0.
\newblock \emph{ACM Trans. Math. Softw.}, June 2023.
\newblock ISSN 0098-3500.
\newblock \doi{10.1145/3585516}.

\bibitem[Budde et~al.(2017)Budde, Dehnert, Hahn, Hartmanns, Junges, and
  Turrini]{jani}
C.~E. Budde, C.~Dehnert, E.~M. Hahn, A.~Hartmanns, S.~Junges, and A.~Turrini.
\newblock {{JANI}}: {{Quantitative Model}} and {{Tool Interaction}}.
\newblock In \emph{The 23rd {{International Conference}} on {{Tools}} and
  {{Algorithms}} for the {{Construction}} and {{Analysis}} of {{Systems}}},
  2017.
\newblock \doi{10.1007/978-3-662-54580-5_9}.

\bibitem[Chevallier and Fleuriot(2021)]{DBLP:journals/corr/abs-2112-05996}
M.~Chevallier and J.~D. Fleuriot.
\newblock Formalising the {{Foundations}} of {{Discrete Reinforcement
  Learning}} in {{Isabelle}}/{{HOL}}.
\newblock \emph{CoRR}, 2021.
\newblock URL \url{https://arxiv.org/abs/2112.05996}.

\bibitem[Dehnert et~al.(2017)Dehnert, Junges, Katoen, and
  Volk]{stormModelChecker}
C.~Dehnert, S.~Junges, J.-P. Katoen, and M.~Volk.
\newblock A {{Storm}} is {{Coming}}: {{A Modern Probabilistic Model Checker}}.
\newblock In \emph{The 29th {{International Conference}} on {{Computer Aided
  Verification}}}, 2017.
\newblock \doi{10.1007/978-3-319-63390-9_31}.

\bibitem[Delgado et~al.(2011)Delgado, Sanner, and
  De~Barros]{sannerFactoredMDPs}
K.~V. Delgado, S.~Sanner, and L.~N. De~Barros.
\newblock Efficient solutions to factored {{MDPs}} with imprecise transition
  probabilities.
\newblock \emph{Artificial Intelligence}, 2011.
\newblock \doi{10.1016/j.artint.2011.01.001}.

\bibitem[Deng et~al.(2022)Deng, Devic, and Juba]{RLFactoredMDPsII}
Z.~Deng, S.~Devic, and B.~Juba.
\newblock Polynomial {{Time Reinforcement Learning}} in {{Factored State MDPs}}
  with {{Linear Value Functions}}.
\newblock In \emph{The 25th {{International Conference}} on {{Artificial
  Intelligence}} and {{Statistics}}}, May 2022.
\newblock URL \url{https://proceedings.mlr.press/v151/deng22c.html}.

\bibitem[Esparza et~al.(2013)Esparza, Lammich, Neumann, Nipkow, Schimpf, and
  Smaus]{esparza2013fully}
J.~Esparza, P.~Lammich, R.~Neumann, T.~Nipkow, A.~Schimpf, and J.-G. Smaus.
\newblock A {{Fully Verified Executable LTL Model Checker}}.
\newblock In \emph{25th {{International Conference}} on {{Computer Aided
  Verification}} ({{CAV}})}, 2013.
\newblock \doi{10.1007/978-3-642-39799-8_31}.

\bibitem[Guestrin et~al.(2003)Guestrin, Koller, Parr, and
  Venkataraman]{guestrin2003efficient}
C.~Guestrin, D.~Koller, R.~Parr, and S.~Venkataraman.
\newblock Efficient solution algorithms for factored {{MDPs}}.
\newblock \emph{J. Artif. Intell. Res.}, 2003.
\newblock \doi{10.1613/jair.1000}.

\bibitem[Hartmanns et~al.(2023)Hartmanns, Kohlen, and Lammich]{SCCsMDPs}
A.~Hartmanns, B.~Kohlen, and P.~Lammich.
\newblock Fast {{Verified SCCs}} for~{{Probabilistic Model Checking}}.
\newblock In \emph{The 21st {{International Symposium}} on {{Automated
  Technology}} for {{Verification}} and {{Analysis}} ({{ATVA}})}, 2023.
\newblock \doi{10.1007/978-3-031-45329-8_9}.

\bibitem[Hinton et~al.(2006)Hinton, Kwiatkowska, Norman, and Parker]{PRISM2006}
A.~Hinton, M.~Z. Kwiatkowska, G.~Norman, and D.~Parker.
\newblock {{PRISM}}: {{A Tool}} for {{Automatic Verification}} of
  {{Probabilistic Systems}}.
\newblock In \emph{The 22nd {{International Conference}} on {{Tools}} and
  {{Algorithms}} for the {{Construction}} and {{Analysis}} of {{Systems}}
  ({{TACAS}})}, 2006.
\newblock \doi{10.1007/11691372_29}.

\bibitem[H{\"o}lzl(2017)]{holzl2017markov}
J.~H{\"o}lzl.
\newblock Markov {{Chains}} and {{Markov Decision Processes}} in
  {{Isabelle}}/{{HOL}}.
\newblock \emph{J. Autom. Reason.}, 2017.
\newblock \doi{10.1007/s10817-016-9401-5}.

\bibitem[H{\"o}lzl and Heller(2011)]{johannesMeasure}
J.~H{\"o}lzl and A.~Heller.
\newblock Three {{Chapters}} of {{Measure Theory}} in {{Isabelle}}/{{HOL}}.
\newblock In \emph{2nd {{International Conference}} on {{Interactive Theorem
  Proving}} ({{ITP}})}, 2011.
\newblock \doi{10.1007/978-3-642-22863-6_12}.

\bibitem[H{\"o}lzl et~al.(2013)H{\"o}lzl, Immler, and Huffman]{holzl2013type}
J.~H{\"o}lzl, F.~Immler, and B.~Huffman.
\newblock Type {{Classes}} and {{Filters}} for {{Mathematical Analysis}} in
  {{Isabelle}}/{{HOL}}.
\newblock In \emph{4th {{International Conference}} on {{Interactive Theorem
  Proving}} ({{ITP}})}, 2013.
\newblock \doi{10.1007/978-3-642-39634-2_21}.

\bibitem[Isac et~al.(2022)Isac, Barrett, Zhang, and Katz]{NNVerificationFarkas}
O.~Isac, C.~Barrett, M.~Zhang, and G.~Katz.
\newblock Neural {{Network Verification}} with {{Proof Production}}.
\newblock In \emph{The 22nd {{Conference}} on {{Formal Methods}} in
  {{Computer-Aided Design}} ({{FMCAD}})}, Oct. 2022.
\newblock \doi{10.34727/2022/isbn.978-3-85448-053-2_9}.

\bibitem[Kanav et~al.(2014)Kanav, Lammich, and
  Popescu]{DBLP:conf/cav/KanavL014}
S.~Kanav, P.~Lammich, and A.~Popescu.
\newblock A {{Conference Management System}} with {{Verified Document
  Confidentiality}}.
\newblock In \emph{The 26th {{International Conference}} on {{Computer Aided
  Verification}} ({{CAV}})}, 2014.
\newblock \doi{10.1007/978-3-319-08867-9_11}.

\bibitem[Klein et~al.(2009)Klein, Elphinstone, Heiser, Andronick, Cock, Derrin,
  Elkaduwe, Engelhardt, Kolanski, Norrish, Sewell, Tuch, and
  Winwood]{klein2009sel4}
G.~Klein, K.~Elphinstone, G.~Heiser, J.~Andronick, D.~Cock, P.~Derrin,
  D.~Elkaduwe, K.~Engelhardt, R.~Kolanski, M.~Norrish, T.~Sewell, H.~Tuch, and
  S.~Winwood.
\newblock {{seL4}}: Formal verification of an {{OS}} kernel.
\newblock In \emph{22nd {{ACM Symposium}} on {{Operating Systems Principles}}
  2009 ({{SOSP}})}, 2009.
\newblock \doi{10.1145/1629575.1629596}.

\bibitem[Lammich and Lochbihler(2019)]{lammichContainers}
P.~Lammich and A.~Lochbihler.
\newblock Automatic {{Refinement}} to {{Efficient Data Structures}}: {{A
  Comparison}} of {{Two Approaches}}.
\newblock \emph{J Autom Reasoning}, June 2019.
\newblock ISSN 1573-0670.
\newblock \doi{10.1007/s10817-018-9461-9}.

\bibitem[Leroy(2009)]{leroy2009formal}
X.~Leroy.
\newblock Formal verification of a realistic compiler.
\newblock \emph{Commun. ACM}, 2009.
\newblock \doi{10.1145/1538788.1538814}.

\bibitem[Massot(N/A)]{MassotWhyFormal}
P.~Massot.
\newblock Why formalize mathematics?
\newblock N/A.

\bibitem[Nipkow et~al.(2002)Nipkow, Paulson, and Wenzel]{IsabelleHOLRef}
T.~Nipkow, L.~C. Paulson, and M.~Wenzel.
\newblock \emph{Isabelle/{{HOL}} - {{A Proof Assistant}} for {{Higher-Order
  Logic}}}.
\newblock 2002.

\bibitem[Obua and Nipkow(2009)]{LPsCertIsabelle}
S.~Obua and T.~Nipkow.
\newblock Flyspeck {{II}}: The basic linear programs.
\newblock \emph{Ann Math Artif Intell}, Aug. 2009.
\newblock ISSN 1573-7470.
\newblock \doi{10.1007/s10472-009-9168-z}.

\bibitem[Osband and Roy()]{RLFactoredMDPs}
I.~Osband and B.~V. Roy.
\newblock Near-optimal {{Reinforcement Learning}} in {{Factored MDPs}}.

\bibitem[Sanner()]{RDDLref}
S.~Sanner.
\newblock Relational {{Dynamic Influence Diagram Language}} ({{RDDL}}):
  {{Language Description}}.

\bibitem[Sch{\"a}ffeler and Abdulaziz(2023)]{ValIterIsabelle}
M.~Sch{\"a}ffeler and M.~Abdulaziz.
\newblock Formally {{Verified Solution Methods}} for {{Infinite-Horizon Markov
  Decision Processes}}.
\newblock In \emph{The 37th {{AAAI Conference}} on {{Artificial Intelligence}}
  ({{AAAI}})}, 2023.
\newblock \doi{10.1609/aaai.v37i12.26759}.

\bibitem[Selsam et~al.(2017)Selsam, Liang, and Dill]{SelsamNNVerification}
D.~Selsam, P.~Liang, and D.~L. Dill.
\newblock Developing {{Bug-Free Machine Learning Systems With Formal
  Mathematics}}.
\newblock In \emph{The 34th {{International Conference}} on {{Machine
  Learning}} ({{ICML}})}, 2017.
\newblock URL \url{http://proceedings.mlr.press/v70/selsam17a.html}.

\bibitem[Spasi{\'c} and Mari{\'c}(2012)]{IsabelleSimplex}
M.~Spasi{\'c} and F.~Mari{\'c}.
\newblock Formalization of {{Incremental Simplex Algorithm}} by {{Stepwise
  Refinement}}.
\newblock In \emph{The 18th {{International Symposium}} on {{Formal Methods}}},
  2012.
\newblock \doi{10.1007/978-3-642-32759-9_35}.

\bibitem[Thiemann(2022)]{LPDualityIsabelle}
R.~Thiemann.
\newblock Duality of {{Linear Programming}}.
\newblock \emph{Arch. Formal Proofs}, 2022.
\newblock URL \url{https://www.isa-afp.org/entries/LP\_Duality.html}.

\bibitem[Vajjha et~al.(2021)Vajjha, Shinnar, Trager, Pestun, and
  Fulton]{VajjhaSTPF21}
K.~Vajjha, A.~Shinnar, B.~M. Trager, V.~Pestun, and N.~Fulton.
\newblock {{CertRL}}: Formalizing convergence proofs for value and policy
  iteration in {{Coq}}.
\newblock In \emph{The 10th {{International Conference}} on {{Certified
  Programs}} and {{Proofs}} ({{CPP}})}, 2021.
\newblock \doi{10.1145/3437992.3439927}.

\bibitem[Wirth(1971)]{refinementWirth}
N.~Wirth.
\newblock Program {{Development}} by {{Stepwise Refinement}}.
\newblock \emph{Commun. ACM}, 1971.
\newblock \doi{10.1145/362575.362577}.

\bibitem[Younes and Littman(2004)]{younes2004ppddl1}
H.~L. Younes and M.~L. Littman.
\newblock {{PPDDL1}}. 0: {{The}} language for the probabilistic part of
  {{IPC-4}}.
\newblock In \emph{Proc. {{International}} Planning Competition}, 2004.

\end{thebibliography}

\newpage
\appendix

\section{Appendix / supplemental material} 

\subsection{Experiments}
\begin{table}[h]
    \footnotesize
\centering
\begin{tabular}{@{}rrrrr@{}}
\toprule
\textbf{Computers} & \textbf{Constraints} & \textbf{Variables} & \textbf{Time (s)} & \textbf{Time in LP (s)} \\ \midrule
1   & 74     & 41     & 0.27     & 0.02 \\
2   & 498     & 268   & 0.58     & 0.08  \\
3   & 1258     & 693  & 0.89    & 0.15  \\
4   & 2578     & 1438     & 1.34    & 0.28  \\
5   & 4378     & 2455     & 1.89     & 0.46  \\
6   & 6658     & 3744     & 2.61     & 0.69 \\
7   & 9418     & 5305     & 3.78     & 0.98  \\
8   & 12658     & 7138     & 5.08     & 1.31  \\
9   & 16378     & 9243     & 6.74     & 1.82  \\
10   & 20578     & 11620     & 8.89     & 2.45  \\
11   & 25258     & 14269     & 12.44     & 3.36  \\
12   & 30418     & 17190     & 15.75     & 4.19  \\
13   & 36058     & 20383     & 20.95     & 5.31 \\
14   & 42178     & 23848    & 26.92     & 6.66  \\
15   & 48778     & 27585     & 34.69     & 8.67  \\
16   & 55858     & 31594     & 44.42     & 10.79 \\
17   & 63418     & 35875     & 58.30     & 16.86  \\
18   & 71458     & 40428     & n.a.     & n.a.  \\
19   & 79978     & 45253     & 92.19     & 30.25  \\
20   & 88978     & 50350     & n.a.     & n.a.
\end{tabular}
\caption{Evaluation on the ring domain, the columns denote the number of variables (number of states grows exponentially), the number of LP constraints and variables generated in the last iteration.
The last two columns give the total running time and the time spent in the LP solver.}
\label{tab:performancefull}
\end{table}
 
\subsection{Isabelle/HOL Snippets}

Here we show listings of the most important parts of the formalization.

\subsubsection{Factored MDPs}
\begin{isasnipenv}[snip:fmdp]{Definition of Factored MDP}
\isacommand{locale}\isamarkupfalse%
\ Factored{\isacharunderscore}{\kern0pt}MDP{\isacharunderscore}{\kern0pt}Consts\ {\isacharequal}{\kern0pt}\isanewline
\ \ \isakeyword{fixes}\isanewline
\ \ \ \ dims\ {\isacharcolon}{\kern0pt}{\isacharcolon}{\kern0pt}\ nat\ \isakeyword{and}\ %
\isamarkupcmt{Number of state variables%
}\isanewline
\ \ \ \ doms\ {\isacharcolon}{\kern0pt}{\isacharcolon}{\kern0pt}\ {\isacartoucheopen}nat\ {\isasymRightarrow}\ {\isacharprime}{\kern0pt}x\ {\isacharcolon}{\kern0pt}{\isacharcolon}{\kern0pt}\ countable\ set{\isacartoucheclose}\ \isakeyword{and}\ %
\isamarkupcmt{Domain of each state variable%
}\isanewline
\isanewline
reward{\isacharunderscore}{\kern0pt}dim\ {\isacharcolon}{\kern0pt}{\isacharcolon}{\kern0pt}\ {\isacartoucheopen}{\isacharprime}{\kern0pt}a\ {\isacharcolon}{\kern0pt}{\isacharcolon}{\kern0pt}countable\ {\isasymRightarrow}\ nat{\isacartoucheclose}\ \isakeyword{and}\isanewline
\isamarkupcmt{number of rewards for action%
}\isanewline
rewards\ {\isacharcolon}{\kern0pt}{\isacharcolon}{\kern0pt}\ {\isacartoucheopen}{\isacharprime}{\kern0pt}a\ {\isasymRightarrow}\ nat\ {\isasymRightarrow}\ {\isacharprime}{\kern0pt}x\ state\ {\isasymRightarrow}\ real{\isacartoucheclose}\ \isakeyword{and}\ \isanewline
\isamarkupcmt{rewards a i x: reward i for action a in state x%
}\isanewline
reward{\isacharunderscore}{\kern0pt}scope\ {\isacharcolon}{\kern0pt}{\isacharcolon}{\kern0pt}\ {\isacartoucheopen}{\isacharprime}{\kern0pt}a\ {\isasymRightarrow}\ nat\ {\isasymRightarrow}\ nat\ set{\isacartoucheclose}\ \isakeyword{and}\isanewline
\isamarkupcmt{variables the reward function depends upon%
}\isanewline
\isanewline
actions\ {\isacharcolon}{\kern0pt}{\isacharcolon}{\kern0pt}\ {\isacartoucheopen}{\isacharprime}{\kern0pt}a\ set{\isacartoucheclose}\ \isakeyword{and}\isanewline
\isamarkupcmt{finite, nonempty set of actions%
}\isanewline
\isanewline
transitions\ {\isacharcolon}{\kern0pt}{\isacharcolon}{\kern0pt}\ {\isacartoucheopen}{\isacharprime}{\kern0pt}a\ {\isasymRightarrow}\ nat\ {\isasymRightarrow}\ {\isacharprime}{\kern0pt}x\ state\ {\isasymRightarrow}\ {\isacharprime}{\kern0pt}x\ pmf{\isacartoucheclose}\ \isakeyword{and}\isanewline
\isamarkupcmt{pmf transitions a i x y: probability that next state at dim i will be y, for action x.%
}\isanewline
transitions{\isacharunderscore}{\kern0pt}scope\ {\isacharcolon}{\kern0pt}{\isacharcolon}{\kern0pt}\ {\isacartoucheopen}{\isacharprime}{\kern0pt}a\ {\isasymRightarrow}\ nat\ {\isasymRightarrow}\ nat\ set{\isacartoucheclose}\ \isakeyword{and}\isanewline
\isamarkupcmt{scopes of the transition functions%
}\isanewline
\isanewline
l\ {\isacharcolon}{\kern0pt}{\isacharcolon}{\kern0pt}\ real\ \isakeyword{and}\isanewline
\isamarkupcmt{discount factor < 1%
}\isanewline
\isanewline
h{\isacharunderscore}{\kern0pt}dim\ {\isacharcolon}{\kern0pt}{\isacharcolon}{\kern0pt}\ {\isacartoucheopen}nat{\isacartoucheclose}\ \isakeyword{and}\isanewline
\isamarkupcmt{number of basis functions%
}\isanewline
h\ {\isacharcolon}{\kern0pt}{\isacharcolon}{\kern0pt}\ {\isacartoucheopen}nat\ {\isasymRightarrow}\ {\isacharprime}{\kern0pt}x\ state\ {\isasymRightarrow}\ real{\isacartoucheclose}\ \isakeyword{and}\isanewline
\isamarkupcmt{basis functions, estimate value function%
}\isanewline
h{\isacharunderscore}{\kern0pt}scope\ {\isacharcolon}{\kern0pt}{\isacharcolon}{\kern0pt}\ {\isacartoucheopen}nat\ {\isasymRightarrow}\ nat\ set{\isacartoucheclose}\isanewline
\isamarkupcmt{scopes of the basis functions%
}\isanewline
\isanewline
\isacommand{locale}\isamarkupfalse%
\ Factored{\isacharunderscore}{\kern0pt}MDP\ {\isacharequal}{\kern0pt}\ Factored{\isacharunderscore}{\kern0pt}MDP{\isacharunderscore}{\kern0pt}Consts\isanewline
\ \ \isakeyword{where}\ \isanewline
\ \ \ \ rewards\ {\isacharequal}{\kern0pt}\ rewards\isanewline
\ \ \isakeyword{for}\ \isanewline
\ \ \ \ rewards\ {\isacharcolon}{\kern0pt}{\isacharcolon}{\kern0pt}\ {\isacartoucheopen}{\isacharprime}{\kern0pt}a{\isacharcolon}{\kern0pt}{\isacharcolon}{\kern0pt}countable\ {\isasymRightarrow}\ nat\ {\isasymRightarrow}\ {\isacharprime}{\kern0pt}x{\isacharcolon}{\kern0pt}{\isacharcolon}{\kern0pt}countable\ state\ {\isasymRightarrow}\ real{\isacartoucheclose}\ {\isacharplus}{\kern0pt}\isanewline
\ \ \isakeyword{assumes}\isanewline
\ \ \ \ transitions{\isacharunderscore}{\kern0pt}closed{\isacharcolon}{\kern0pt}\isanewline
\ \ \ \ {\isacartoucheopen}{\isasymAnd}a\ i\ x{\isachardot}{\kern0pt}\isanewline
\ \ \ \ \ \ \ \ a\ {\isasymin}\ actions\ {\isasymLongrightarrow}\isanewline
\ \ \ \ \ \ \ \ i\ {\isacharless}{\kern0pt}\ dims\ {\isasymLongrightarrow}\isanewline
\ \ \ \ \ \ \ \ x\ {\isasymin}\ partial{\isacharunderscore}{\kern0pt}states{\isacharunderscore}{\kern0pt}on\ {\isacharparenleft}{\kern0pt}transitions{\isacharunderscore}{\kern0pt}scope\ a\ i{\isacharparenright}{\kern0pt}\ {\isasymLongrightarrow}\isanewline
\ \ \ \ \ \ \ \ set{\isacharunderscore}{\kern0pt}pmf\ {\isacharparenleft}{\kern0pt}transitions\ a\ i\ x{\isacharparenright}{\kern0pt}\ {\isasymsubseteq}\ doms\ i{\isacartoucheclose}\ \isakeyword{and}\ %
\isamarkupcmt{transitions go to domains%
}\isanewline
\ \ \ \ transitions{\isacharunderscore}{\kern0pt}scope{\isacharcolon}{\kern0pt}\ \isanewline
\ \ \ \ {\isacartoucheopen}{\isasymAnd}a\ i{\isachardot}{\kern0pt}\isanewline
\ \ \ \ \ \ \ \ a\ {\isasymin}\ actions\ {\isasymLongrightarrow}\isanewline
\ \ \ \ \ \ \ \ i\ {\isacharless}{\kern0pt}\ dims\ {\isasymLongrightarrow}\isanewline
\ \ \ \ \ \ \ \ has{\isacharunderscore}{\kern0pt}scope{\isacharunderscore}{\kern0pt}on\ {\isacharparenleft}{\kern0pt}transitions\ a\ i{\isacharparenright}{\kern0pt}\ partial{\isacharunderscore}{\kern0pt}states\ {\isacharparenleft}{\kern0pt}transitions{\isacharunderscore}{\kern0pt}scope\ a\ i{\isacharparenright}{\kern0pt}{\isacartoucheclose}\ \isakeyword{and}\isanewline
\ \ \ \ %
\isamarkupcmt{the transition functions have the specified scopes%
}\isanewline
\ \ \ \ transitions{\isacharunderscore}{\kern0pt}scope{\isacharunderscore}{\kern0pt}dims{\isacharcolon}{\kern0pt}\isanewline
\ \ \ \ {\isacartoucheopen}{\isasymAnd}a\ i{\isachardot}{\kern0pt}\isanewline
\ \ \ \ \ \ \ \ a\ {\isasymin}\ actions\ {\isasymLongrightarrow}\isanewline
\ \ \ \ \ \ \ \ i\ {\isacharless}{\kern0pt}\ dims\ {\isasymLongrightarrow}\ transitions{\isacharunderscore}{\kern0pt}scope\ a\ i\ {\isasymsubseteq}\ {\isacharbraceleft}{\kern0pt}{\isachardot}{\kern0pt}{\isachardot}{\kern0pt}{\isacharless}{\kern0pt}dims{\isacharbraceright}{\kern0pt}{\isacartoucheclose}\ \isakeyword{and}\isanewline
\ \ \ \ %
\isamarkupcmt{the scopes are valid%
}\isanewline
\ \ \ \ actions{\isacharunderscore}{\kern0pt}fin{\isacharcolon}{\kern0pt}\ {\isacartoucheopen}finite\ actions{\isacartoucheclose}\ \isakeyword{and}\isanewline
\ \ \ \ actions{\isacharunderscore}{\kern0pt}ne{\isacharcolon}{\kern0pt}\ {\isacartoucheopen}actions\ {\isasymnoteq}\ {\isacharbraceleft}{\kern0pt}{\isacharbraceright}{\kern0pt}{\isacartoucheclose}\ \isakeyword{and}\isanewline
\ \ \ \ %
\isamarkupcmt{finite, nonempty action sets%
}\isanewline
\isanewline
doms{\isacharunderscore}{\kern0pt}fin{\isacharcolon}{\kern0pt}\ {\isacartoucheopen}{\isasymAnd}i{\isachardot}{\kern0pt}\ i\ {\isacharless}{\kern0pt}\ dims\ {\isasymLongrightarrow}\ finite\ {\isacharparenleft}{\kern0pt}doms\ i{\isacharparenright}{\kern0pt}{\isacartoucheclose}\ \isakeyword{and}\isanewline
doms{\isacharunderscore}{\kern0pt}ne{\isacharcolon}{\kern0pt}\ {\isacartoucheopen}{\isasymAnd}i{\isachardot}{\kern0pt}\ i\ {\isacharless}{\kern0pt}\ dims\ {\isasymLongrightarrow}\ doms\ i\ {\isasymnoteq}\ {\isacharbraceleft}{\kern0pt}{\isacharbraceright}{\kern0pt}{\isacartoucheclose}\ \isakeyword{and}\isanewline
\isamarkupcmt{finite, nonempty domains%
}\isanewline
\isanewline
dims{\isacharunderscore}{\kern0pt}pos{\isacharcolon}{\kern0pt}\ {\isacartoucheopen}dims\ {\isachargreater}{\kern0pt}\ {\isadigit{0}}{\isacartoucheclose}\ \isakeyword{and}\isanewline
\isamarkupcmt{there exists a domain%
}\isanewline
\isanewline
reward{\isacharunderscore}{\kern0pt}scope{\isacharcolon}{\kern0pt}\ \isanewline
{\isacartoucheopen}{\isasymAnd}a\ i{\isachardot}{\kern0pt}\ \isanewline
\ \ \ \ \ \ \ \ a\ {\isasymin}\ actions\ {\isasymLongrightarrow}\isanewline
\ \ \ \ \ \ \ \ i\ {\isacharless}{\kern0pt}\ reward{\isacharunderscore}{\kern0pt}dim\ a\ {\isasymLongrightarrow}\isanewline
\ \ \ \ \ \ \ \ has{\isacharunderscore}{\kern0pt}scope{\isacharunderscore}{\kern0pt}on\ {\isacharparenleft}{\kern0pt}rewards\ a\ i{\isacharparenright}{\kern0pt}\ partial{\isacharunderscore}{\kern0pt}states\ {\isacharparenleft}{\kern0pt}reward{\isacharunderscore}{\kern0pt}scope\ a\ i{\isacharparenright}{\kern0pt}{\isacartoucheclose}\ \isakeyword{and}\isanewline
reward{\isacharunderscore}{\kern0pt}scope{\isacharunderscore}{\kern0pt}dims{\isacharcolon}{\kern0pt}\isanewline
{\isacartoucheopen}{\isasymAnd}a\ i{\isachardot}{\kern0pt}\isanewline
\ \ \ \ \ \ \ \ a\ {\isasymin}\ actions\ {\isasymLongrightarrow}\isanewline
\ \ \ \ \ \ \ \ i\ {\isacharless}{\kern0pt}\ reward{\isacharunderscore}{\kern0pt}dim\ a\ {\isasymLongrightarrow}\ reward{\isacharunderscore}{\kern0pt}scope\ a\ i\ {\isasymsubseteq}\ {\isacharbraceleft}{\kern0pt}{\isachardot}{\kern0pt}{\isachardot}{\kern0pt}{\isacharless}{\kern0pt}dims{\isacharbraceright}{\kern0pt}{\isacartoucheclose}\ \isakeyword{and}\isanewline
\isamarkupcmt{the reward functions are proper scoped functions%
}\isanewline
\isanewline
h{\isacharunderscore}{\kern0pt}scope{\isacharcolon}{\kern0pt}\isanewline
{\isacartoucheopen}{\isasymAnd}i{\isachardot}{\kern0pt}\ \isanewline
\ \ \ \ \ \ \ \ i\ {\isacharless}{\kern0pt}\ h{\isacharunderscore}{\kern0pt}dim\ {\isasymLongrightarrow}\ \isanewline
\ \ \ \ \ \ \ \ has{\isacharunderscore}{\kern0pt}scope{\isacharunderscore}{\kern0pt}on\ {\isacharparenleft}{\kern0pt}h\ i{\isacharparenright}{\kern0pt}\ partial{\isacharunderscore}{\kern0pt}states\ {\isacharparenleft}{\kern0pt}h{\isacharunderscore}{\kern0pt}scope\ i{\isacharparenright}{\kern0pt}{\isacartoucheclose}\ \isakeyword{and}\isanewline
h{\isacharunderscore}{\kern0pt}scope{\isacharunderscore}{\kern0pt}dims{\isacharcolon}{\kern0pt}\isanewline
{\isacartoucheopen}{\isasymAnd}i{\isachardot}{\kern0pt}\ i\ {\isacharless}{\kern0pt}\ h{\isacharunderscore}{\kern0pt}dim\ {\isasymLongrightarrow}\ h{\isacharunderscore}{\kern0pt}scope\ i\ {\isasymsubseteq}\ {\isacharbraceleft}{\kern0pt}{\isachardot}{\kern0pt}{\isachardot}{\kern0pt}{\isacharless}{\kern0pt}dims{\isacharbraceright}{\kern0pt}{\isacartoucheclose}\ \isakeyword{and}\isanewline
\isamarkupcmt{the basis functions are proper scoped functions%
}\isanewline
\isanewline
disc{\isacharunderscore}{\kern0pt}lt{\isacharunderscore}{\kern0pt}one{\isacharcolon}{\kern0pt}\ {\isacartoucheopen}l\ {\isacharless}{\kern0pt}\ {\isadigit{1}}{\isacartoucheclose}\ \isakeyword{and}\isanewline
disc{\isacharunderscore}{\kern0pt}nonneg{\isacharcolon}{\kern0pt}\ {\isacartoucheopen}l\ {\isasymge}\ {\isadigit{0}}{\isacartoucheclose}\ %
\isamarkupcmt{valid discount factor%
}\isanewline
\end{isasnipenv}

\begin{isasnipenv}[snip:fmdpd]{Factored MDPs with a Default Action}
\isacommand{locale}\isamarkupfalse%
\ Factored{\isacharunderscore}{\kern0pt}MDP{\isacharunderscore}{\kern0pt}Default{\isacharunderscore}{\kern0pt}Consts\ {\isacharequal}{\kern0pt}\ Factored{\isacharunderscore}{\kern0pt}MDP{\isacharunderscore}{\kern0pt}Consts\isanewline
\ \ \isakeyword{where}\ \isanewline
\ \ \ \ transitions\ {\isacharequal}{\kern0pt}\ transitions\isanewline
\ \ \isakeyword{for}\isanewline
\ \ \ \ transitions\ {\isacharcolon}{\kern0pt}{\isacharcolon}{\kern0pt}\ {\isacartoucheopen}{\isacharprime}{\kern0pt}a{\isacharcolon}{\kern0pt}{\isacharcolon}{\kern0pt}{\isacharbraceleft}{\kern0pt}countable{\isacharcomma}{\kern0pt}\ linorder{\isacharbraceright}{\kern0pt}\ {\isasymRightarrow}\ nat\ {\isasymRightarrow}\ {\isacharprime}{\kern0pt}x{\isacharcolon}{\kern0pt}{\isacharcolon}{\kern0pt}{\isacharbraceleft}{\kern0pt}countable{\isacharcomma}{\kern0pt}\ linorder{\isacharbraceright}{\kern0pt}\ state\ {\isasymRightarrow}\ {\isacharprime}{\kern0pt}x\ pmf{\isacartoucheclose}\ {\isacharplus}{\kern0pt}\isanewline
\ \ \isakeyword{fixes}\isanewline
\ \ \ \ d\ {\isacharcolon}{\kern0pt}{\isacharcolon}{\kern0pt}\ {\isacartoucheopen}{\isacharprime}{\kern0pt}a{\isacartoucheclose}\ \isakeyword{and}\isanewline
\ \ \ \ effects\ {\isacharcolon}{\kern0pt}{\isacharcolon}{\kern0pt}\ {\isacartoucheopen}{\isacharprime}{\kern0pt}a\ {\isasymRightarrow}\ nat\ set{\isacartoucheclose}\isanewline
\isanewline
\isacommand{locale}\isamarkupfalse%
\ Factored{\isacharunderscore}{\kern0pt}MDP{\isacharunderscore}{\kern0pt}Default\ {\isacharequal}{\kern0pt}\ \isanewline
\ \ Factored{\isacharunderscore}{\kern0pt}MDP{\isacharunderscore}{\kern0pt}Default{\isacharunderscore}{\kern0pt}Consts\isanewline
\ \ \isakeyword{where}\ \isanewline
\ \ \ \ transitions\ {\isacharequal}{\kern0pt}\ transitions\ {\isacharplus}{\kern0pt}\isanewline
\ \ Factored{\isacharunderscore}{\kern0pt}MDP\isanewline
\ \ \isakeyword{where}\isanewline
\ \ \ \ transitions\ {\isacharequal}{\kern0pt}\ transitions\isanewline
\ \ \isakeyword{for}\isanewline
\ \ \ \ transitions\ {\isacharcolon}{\kern0pt}{\isacharcolon}{\kern0pt}\ {\isacartoucheopen}{\isacharprime}{\kern0pt}a{\isacharcolon}{\kern0pt}{\isacharcolon}{\kern0pt}{\isacharbraceleft}{\kern0pt}countable{\isacharcomma}{\kern0pt}\ linorder{\isacharbraceright}{\kern0pt}\ {\isasymRightarrow}\ nat\ {\isasymRightarrow}\ {\isacharprime}{\kern0pt}x{\isacharcolon}{\kern0pt}{\isacharcolon}{\kern0pt}{\isacharbraceleft}{\kern0pt}countable{\isacharcomma}{\kern0pt}\ linorder{\isacharbraceright}{\kern0pt}\ state\ {\isasymRightarrow}\ {\isacharprime}{\kern0pt}x\ pmf{\isacartoucheclose}\ {\isacharplus}{\kern0pt}\isanewline
\ \ \isakeyword{assumes}\isanewline
\ \ \ \ default{\isacharunderscore}{\kern0pt}act{\isacharcolon}{\kern0pt}\ {\isacartoucheopen}d\ {\isasymin}\ actions{\isacartoucheclose}\ \isakeyword{and}\isanewline
\ \ \ \ effects{\isacharcolon}{\kern0pt}\ {\isacartoucheopen}{\isasymAnd}i\ a{\isachardot}{\kern0pt}\isanewline
\ \ \ \ \ \ i\ {\isacharless}{\kern0pt}\ dims\ {\isasymLongrightarrow}\isanewline
\ \ \ \ \ \ a\ {\isasymin}\ actions\ {\isasymLongrightarrow}\ \isanewline
\ \ \ \ \ \ {\isacharparenleft}{\kern0pt}{\isasymexists}x\ {\isasymin}\ partial{\isacharunderscore}{\kern0pt}states{\isachardot}{\kern0pt}\ transitions\ a\ i\ x\ {\isasymnoteq}\ transitions\ d\ i\ x{\isacharparenright}{\kern0pt}\ {\isasymLongrightarrow}\isanewline
\ \ \ \ \ \ i\ {\isasymin}\ effects\ a{\isacartoucheclose}\ \isakeyword{and}\isanewline
\ \ \ \ effects{\isacharunderscore}{\kern0pt}default{\isacharcolon}{\kern0pt}\isanewline
\ \ \ \ \ \ {\isacartoucheopen}effects\ d\ {\isacharequal}{\kern0pt}\ {\isacharbraceleft}{\kern0pt}{\isacharbraceright}{\kern0pt}{\isacartoucheclose}\ \isakeyword{and}\isanewline
\ \ \ \ \isanewline
\ \ \ \ rewards{\isacharunderscore}{\kern0pt}default{\isacharunderscore}{\kern0pt}dim{\isacharcolon}{\kern0pt}\ {\isacartoucheopen}{\isasymAnd}a{\isachardot}{\kern0pt}\ a\ {\isasymin}\ actions\ {\isasymLongrightarrow}\ reward{\isacharunderscore}{\kern0pt}dim\ a\ {\isasymge}\ reward{\isacharunderscore}{\kern0pt}dim\ d{\isacartoucheclose}\ \isakeyword{and}\isanewline
\ \ \ \ rewards{\isacharunderscore}{\kern0pt}eq{\isacharcolon}{\kern0pt}\ {\isacartoucheopen}{\isasymAnd}a\ i{\isachardot}{\kern0pt}\ a\ {\isasymin}\ actions\ {\isasymLongrightarrow}\ i\ {\isacharless}{\kern0pt}\ reward{\isacharunderscore}{\kern0pt}dim\ d\ {\isasymLongrightarrow}\ rewards\ a\ i\ {\isacharequal}{\kern0pt}\ rewards\ d\ i{\isacartoucheclose}\ \isakeyword{and}\isanewline
\ \ \ \ reward{\isacharunderscore}{\kern0pt}scope{\isacharunderscore}{\kern0pt}eq{\isacharcolon}{\kern0pt}\ {\isacartoucheopen}{\isasymAnd}a\ i{\isachardot}{\kern0pt}\ a\ {\isasymin}\ actions\ {\isasymLongrightarrow}\ i\ {\isacharless}{\kern0pt}\ reward{\isacharunderscore}{\kern0pt}dim\ d\ {\isasymLongrightarrow}\ reward{\isacharunderscore}{\kern0pt}scope\ a\ i\ {\isacharequal}{\kern0pt}\ reward{\isacharunderscore}{\kern0pt}scope\ d\ i{\isacartoucheclose}\isanewline
\ \ \ \ %
\isamarkupcmt{The first \isa{reward{\isacharunderscore}{\kern0pt}dim\ d} reward functions coincide for all actions.%
}
\end{isasnipenv}

\begin{isasnipenv}[snip:states]{(Partial) States}
\isacommand{definition}\isamarkupfalse%
\ {\isacartoucheopen}partial{\isacharunderscore}{\kern0pt}states\ {\isacharequal}{\kern0pt}\ {\isacharbraceleft}{\kern0pt}x{\isachardot}{\kern0pt}\ {\isasymforall}{\isacharparenleft}{\kern0pt}i{\isacharcomma}{\kern0pt}\ j{\isacharparenright}{\kern0pt}\ {\isasymin}\ Map{\isachardot}{\kern0pt}graph\ x{\isachardot}{\kern0pt}\ i\ {\isasymin}\ vars\ {\isasymand}\ j\ {\isasymin}\ doms\ i{\isacharbraceright}{\kern0pt}{\isacartoucheclose}\isanewline
\isacommand{definition}\isamarkupfalse%
\ {\isacartoucheopen}states\ {\isacharequal}{\kern0pt}\ partial{\isacharunderscore}{\kern0pt}states\ {\isasyminter}\ {\isacharbraceleft}{\kern0pt}x{\isachardot}{\kern0pt}\ dom\ x\ {\isacharequal}{\kern0pt}\ vars{\isacharbraceright}{\kern0pt}{\isacartoucheclose}\isanewline
\isacommand{definition}\isamarkupfalse%
\ {\isacartoucheopen}partial{\isacharunderscore}{\kern0pt}states{\isacharunderscore}{\kern0pt}on\ X\ {\isacharequal}{\kern0pt}\ partial{\isacharunderscore}{\kern0pt}states\ {\isasyminter}\ {\isacharbraceleft}{\kern0pt}x{\isachardot}{\kern0pt}\ X\ {\isasymsubseteq}\ dom\ x{\isacharbraceright}{\kern0pt}{\isacartoucheclose}\isanewline
\isacommand{definition}\isamarkupfalse%
\ {\isacartoucheopen}partial{\isacharunderscore}{\kern0pt}states{\isacharunderscore}{\kern0pt}on{\isacharunderscore}{\kern0pt}eq\ X\ {\isacharequal}{\kern0pt}\ partial{\isacharunderscore}{\kern0pt}states\ {\isasyminter}\ {\isacharbraceleft}{\kern0pt}x{\isachardot}{\kern0pt}\ X\ {\isacharequal}{\kern0pt}\ dom\ x{\isacharbraceright}{\kern0pt}{\isacartoucheclose}\isanewline
\isanewline
\isacommand{definition}\isamarkupfalse%
\ {\isacartoucheopen}consistent\ x\ x{\isacharprime}{\kern0pt}\ {\isasymlongleftrightarrow}\ x\ {\isacharbar}{\kern0pt}{\isacharbackquote}{\kern0pt}\ dom\ x{\isacharprime}{\kern0pt}\ {\isacharequal}{\kern0pt}\ x{\isacharprime}{\kern0pt}{\isacartoucheclose}
\end{isasnipenv}

\begin{isasnipenv}[snip:scoped]{Scoped Functions}
\isacommand{definition}\isamarkupfalse%
\ {\isacartoucheopen}has{\isacharunderscore}{\kern0pt}scope{\isacharunderscore}{\kern0pt}on\ f\ D\ R\ {\isacharequal}{\kern0pt}\isanewline
\ \ {\isacharparenleft}{\kern0pt}{\isasymforall}d\ {\isasymin}\ D{\isachardot}{\kern0pt}\ {\isasymforall}d{\isacharprime}{\kern0pt}\ {\isasymin}\ D{\isachardot}{\kern0pt}\ restrict\ d\ R\ {\isacharequal}{\kern0pt}\ restrict\ d{\isacharprime}{\kern0pt}\ R\ {\isasymlongrightarrow}\ f\ d\ {\isacharequal}{\kern0pt}\ f\ d{\isacharprime}{\kern0pt}{\isacharparenright}{\kern0pt}{\isacartoucheclose}\isanewline
\isanewline
\isacommand{definition}\isamarkupfalse%
\ {\isacartoucheopen}instantiate\ f\ t\ x\ {\isacharequal}{\kern0pt}\ f\ {\isacharparenleft}{\kern0pt}x\ {\isacharplus}{\kern0pt}{\isacharplus}{\kern0pt}\ t{\isacharparenright}{\kern0pt}{\isacartoucheclose}%
\end{isasnipenv}

\begin{isasnipenv}[snip:trans]{Basic Definitions on Factored MDPs}
\isacommand{definition}\isamarkupfalse%
\ p\isactrlsub a\ {\isacharcolon}{\kern0pt}{\isacharcolon}{\kern0pt}\ {\isacartoucheopen}{\isacharprime}{\kern0pt}a\ {\isasymRightarrow}\ {\isacharprime}{\kern0pt}x\ state\ {\isasymRightarrow}\ {\isacharprime}{\kern0pt}x\ state\ {\isasymRightarrow}\ real{\isacartoucheclose}\ \isakeyword{where}\isanewline
{\isachardoublequoteopen}p\isactrlsub a\ a\ x\ x{\isacharprime}{\kern0pt}\ {\isacharequal}{\kern0pt}\ {\isacharparenleft}{\kern0pt}{\isasymProd}i\ {\isacharless}{\kern0pt}\ dims{\isachardot}{\kern0pt}\ pmf\ {\isacharparenleft}{\kern0pt}transitions\ a\ i\ x{\isacharparenright}{\kern0pt}\ {\isacharparenleft}{\kern0pt}the\ {\isacharparenleft}{\kern0pt}x{\isacharprime}{\kern0pt}\ i{\isacharparenright}{\kern0pt}{\isacharparenright}{\kern0pt}{\isacharparenright}{\kern0pt}{\isachardoublequoteclose}%
\isanewline
\isanewline
\isacommand{definition}\isamarkupfalse%
\ {\isacartoucheopen}reward\ a\ x\ {\isacharequal}{\kern0pt}\ {\isacharparenleft}{\kern0pt}{\isasymSum}i\ {\isacharless}{\kern0pt}\ reward{\isacharunderscore}{\kern0pt}dim\ a{\isachardot}{\kern0pt}\ rewards\ a\ i\ x{\isacharparenright}{\kern0pt}{\isacartoucheclose}\isanewline
\end{isasnipenv}

\begin{isasnipenv}[snip:linval]{Linear Value Functions}
\isacommand{definition}\isamarkupfalse
\ {\isacartoucheopen}{\isasymnu}\isactrlsub w\ w\ x\ {\isacharequal}{\kern0pt}\ {\isacharparenleft}{\kern0pt}if\ x\ {\isasymin}\ partial{\isacharunderscore}{\kern0pt}states\ then\ {\isacharparenleft}{\kern0pt}{\isasymSum}i\ {\isacharless}{\kern0pt}\ h{\isacharunderscore}{\kern0pt}dim{\isachardot}{\kern0pt}\ w\ i\ {\isacharasterisk}{\kern0pt}\ h\ i\ x{\isacharparenright}{\kern0pt}\ else\ {\isadigit{0}}{\isacharparenright}{\kern0pt}{\isacartoucheclose}%

\isanewline
\isacommand{definition}\isamarkupfalse%
\ g\ {\isacharcolon}{\kern0pt}{\isacharcolon}{\kern0pt}\ {\isacartoucheopen}nat\ {\isasymRightarrow}\ {\isacharprime}{\kern0pt}a\ {\isasymRightarrow}\ {\isacharprime}{\kern0pt}x\ state\ {\isasymRightarrow}\ real{\isacartoucheclose}\ \isakeyword{where}\isanewline
\ \ {\isacartoucheopen}g\ i\ a\ x\ {\isacharequal}{\kern0pt}\ {\isacharparenleft}{\kern0pt}{\isasymSum}x{\isacharprime}{\kern0pt}\ {\isasymin}\ states{\isachardot}{\kern0pt}\ p\isactrlsub a\ a\ x\ x{\isacharprime}{\kern0pt}\ {\isacharasterisk}{\kern0pt}\ h\ i\ x{\isacharprime}{\kern0pt}{\isacharparenright}{\kern0pt}{\isacartoucheclose}%
\isanewline

\isacommand{definition}\isamarkupfalse%
\ {\isasymGamma}\isactrlsub a\ {\isacharcolon}{\kern0pt}{\isacharcolon}{\kern0pt}\ {\isacartoucheopen}{\isacharprime}{\kern0pt}a\ {\isasymRightarrow}\ nat\ set\ {\isasymRightarrow}\ nat\ set{\isacartoucheclose}\ \isakeyword{where}\isanewline
\ \ {\isacartoucheopen}{\isasymGamma}\isactrlsub a\ a\ I\ {\isacharequal}{\kern0pt}\ {\isacharparenleft}{\kern0pt}{\isasymUnion}i\ {\isasymin}\ I{\isachardot}{\kern0pt}\ transitions{\isacharunderscore}{\kern0pt}scope\ a\ i{\isacharparenright}{\kern0pt}{\isacartoucheclose}

\isanewline
\isacommand{lemma}\isamarkupfalse%
\ scope{\isacharunderscore}{\kern0pt}g{\isacharcolon}{\kern0pt}\isanewline
\ \ \isakeyword{assumes}\ a{\isacharcolon}{\kern0pt}\ {\isacartoucheopen}a\ {\isasymin}\ actions{\isacartoucheclose}\isanewline
\ \ \isakeyword{assumes}\ i{\isacharcolon}{\kern0pt}\ {\isacartoucheopen}i\ {\isacharless}{\kern0pt}\ h{\isacharunderscore}{\kern0pt}dim{\isacartoucheclose}\isanewline
\ \ \isakeyword{shows}\ {\isacartoucheopen}has{\isacharunderscore}{\kern0pt}scope{\isacharunderscore}{\kern0pt}on\ {\isacharparenleft}{\kern0pt}g\ i\ a{\isacharparenright}{\kern0pt}\ states\ {\isacharparenleft}{\kern0pt}{\isasymGamma}\isactrlsub a\ a\ {\isacharparenleft}{\kern0pt}h{\isacharunderscore}{\kern0pt}scope\ i{\isacharparenright}{\kern0pt}{\isacharparenright}{\kern0pt}{\isacartoucheclose}\isanewline
\isanewline
\isacommand{lemma}\isamarkupfalse%
\ Q{\isacharunderscore}{\kern0pt}g{\isacharcolon}{\kern0pt}\isanewline
\ \ \isakeyword{assumes}\ {\isacartoucheopen}a\ {\isasymin}\ actions{\isacartoucheclose}\ {\isacartoucheopen}x\ {\isasymin}\ partial{\isacharunderscore}{\kern0pt}states{\isacartoucheclose}\isanewline
\ \ \isakeyword{shows}\ {\isacartoucheopen}Q\ w\ x\ a\ {\isacharequal}{\kern0pt}\ reward\ a\ x\ {\isacharplus}{\kern0pt}\ l\ {\isacharasterisk}{\kern0pt}\ {\isacharparenleft}{\kern0pt}{\isasymSum}i\ {\isacharless}{\kern0pt}\ h{\isacharunderscore}{\kern0pt}dim{\isachardot}{\kern0pt}\ w\ i\ {\isacharasterisk}{\kern0pt}\ g\ i\ a\ x{\isacharparenright}{\kern0pt}{\isacartoucheclose}
\end{isasnipenv}

\subsubsection{Approximate Policy Iteration}
\begin{isasnipenv}[snip:api]{Approximate Policy Iteration}
\isacommand{locale}\isamarkupfalse%
\ API{\isacharunderscore}{\kern0pt}Interface{\isacharunderscore}{\kern0pt}Consts\ {\isacharequal}{\kern0pt}\isanewline
\ \ Factored{\isacharunderscore}{\kern0pt}MDP{\isacharunderscore}{\kern0pt}Default{\isacharunderscore}{\kern0pt}Consts\ {\isacharplus}{\kern0pt}\isanewline
\ \ \isakeyword{fixes}\isanewline
\ \ \ \ {\isasymepsilon}\ {\isacharcolon}{\kern0pt}{\isacharcolon}{\kern0pt}\ real\ \isakeyword{and}\ %
\isamarkupcmt{desired precision%
}\isanewline
\ \ \ \ t{\isacharunderscore}{\kern0pt}max\ {\isacharcolon}{\kern0pt}{\isacharcolon}{\kern0pt}\ nat\ \isakeyword{and}\ %
\isamarkupcmt{max number of iterations%
}\isanewline
\ \ \ \ update{\isacharunderscore}{\kern0pt}dec{\isacharunderscore}{\kern0pt}list\ {\isacharcolon}{\kern0pt}{\isacharcolon}{\kern0pt}\ {\isacartoucheopen}weights\ {\isasymRightarrow}\ {\isacharparenleft}{\kern0pt}{\isacharprime}{\kern0pt}a\ state\ {\isasymtimes}\ {\isacharprime}{\kern0pt}b{\isacharparenright}{\kern0pt}\ list{\isacartoucheclose}\ \isakeyword{and}\isanewline
\ \ \ \ update{\isacharunderscore}{\kern0pt}weights\ {\isacharcolon}{\kern0pt}{\isacharcolon}{\kern0pt}\ {\isacartoucheopen}nat\ {\isasymRightarrow}\ {\isacharparenleft}{\kern0pt}{\isacharprime}{\kern0pt}a\ state\ {\isasymtimes}\ {\isacharprime}{\kern0pt}b{\isacharparenright}{\kern0pt}\ list\ {\isasymRightarrow}\ weights{\isacartoucheclose}\isanewline
\isakeyword{begin}\isanewline
\isanewline
\isacommand{definition}\isamarkupfalse%
\ {\isacartoucheopen}is{\isacharunderscore}{\kern0pt}greedy{\isacharunderscore}{\kern0pt}dec{\isacharunderscore}{\kern0pt}list{\isacharunderscore}{\kern0pt}pol\ w\ p\ {\isasymlongleftrightarrow}\isanewline
\ \ is{\isacharunderscore}{\kern0pt}dec{\isacharunderscore}{\kern0pt}list{\isacharunderscore}{\kern0pt}pol\ p\ {\isasymand}\isanewline
\ \ {\isacharparenleft}{\kern0pt}is{\isacharunderscore}{\kern0pt}greedy{\isacharunderscore}{\kern0pt}pol{\isacharunderscore}{\kern0pt}w\ w\ {\isacharparenleft}{\kern0pt}dec{\isacharunderscore}{\kern0pt}list{\isacharunderscore}{\kern0pt}to{\isacharunderscore}{\kern0pt}pol\ p{\isacharparenright}{\kern0pt}{\isacharparenright}{\kern0pt}\ %
\isamarkupcmt{turned into a policy, it is optimal wrt the weights%
}{\isacartoucheclose}%
\isanewline
\isanewline
\isacommand{definition}\isamarkupfalse%
\ {\isacartoucheopen}opt{\isacharunderscore}{\kern0pt}weights{\isacharunderscore}{\kern0pt}pol\ p\ w\ {\isacharequal}{\kern0pt}\ is{\isacharunderscore}{\kern0pt}arg{\isacharunderscore}{\kern0pt}min\ {\isacharparenleft}{\kern0pt}proj{\isacharunderscore}{\kern0pt}err{\isacharunderscore}{\kern0pt}w\ p{\isacharparenright}{\kern0pt}\ {\isacharparenleft}{\kern0pt}{\isasymlambda}{\isacharunderscore}{\kern0pt}{\isachardot}{\kern0pt}\ True{\isacharparenright}{\kern0pt}\ w{\isacartoucheclose}\isanewline
\isanewline
\isacommand{definition}\isamarkupfalse%
\ {\isacartoucheopen}update{\isacharunderscore}{\kern0pt}weights{\isacharunderscore}{\kern0pt}spec\ {\isasymlongleftrightarrow}\ {\isacharparenleft}{\kern0pt}{\isasymforall}p\ i{\isachardot}{\kern0pt}\ is{\isacharunderscore}{\kern0pt}dec{\isacharunderscore}{\kern0pt}list{\isacharunderscore}{\kern0pt}pol\ p\ {\isasymlongrightarrow}\ opt{\isacharunderscore}{\kern0pt}weights{\isacharunderscore}{\kern0pt}pol\ {\isacharparenleft}{\kern0pt}dec{\isacharunderscore}{\kern0pt}list{\isacharunderscore}{\kern0pt}to{\isacharunderscore}{\kern0pt}pol\ p{\isacharparenright}{\kern0pt}\ {\isacharparenleft}{\kern0pt}update{\isacharunderscore}{\kern0pt}weights\ i\ p{\isacharparenright}{\kern0pt}{\isacharparenright}{\kern0pt}{\isacartoucheclose}\isanewline
\isanewline
\isacommand{definition}\isamarkupfalse%
\ {\isacartoucheopen}dec{\isacharunderscore}{\kern0pt}list{\isacharunderscore}{\kern0pt}pol{\isacharunderscore}{\kern0pt}spec\ {\isasymlongleftrightarrow}\ {\isacharparenleft}{\kern0pt}{\isasymforall}w{\isachardot}{\kern0pt}\ is{\isacharunderscore}{\kern0pt}greedy{\isacharunderscore}{\kern0pt}dec{\isacharunderscore}{\kern0pt}list{\isacharunderscore}{\kern0pt}pol\ w\ {\isacharparenleft}{\kern0pt}update{\isacharunderscore}{\kern0pt}dec{\isacharunderscore}{\kern0pt}list\ w{\isacharparenright}{\kern0pt}{\isacharparenright}{\kern0pt}{\isacartoucheclose}%
\isanewline\isanewline
\isacommand{definition}\isamarkupfalse%
\ {\isacartoucheopen}proj{\isacharunderscore}{\kern0pt}err{\isacharunderscore}{\kern0pt}w\ p\ w\ {\isacharequal}{\kern0pt}\ {\isacharparenleft}{\kern0pt}{\isasymSqunion}x\ {\isasymin}\ states{\isachardot}{\kern0pt}\ dist\ {\isacharparenleft}{\kern0pt}{\isasymnu}\isactrlsub w\ w\ x{\isacharparenright}{\kern0pt}\ {\isacharparenleft}{\kern0pt}Q\ w\ x\ {\isacharparenleft}{\kern0pt}p\ x{\isacharparenright}{\kern0pt}{\isacharparenright}{\kern0pt}{\isacharparenright}{\kern0pt}{\isacartoucheclose}%
\isanewline\isanewline
\isacommand{function}\isamarkupfalse%
\ api{\isacharunderscore}{\kern0pt}aux\ \isakeyword{where}\isanewline
{\isacartoucheopen}api{\isacharunderscore}{\kern0pt}aux\ pw\ t\ {\isacharequal}{\kern0pt}\ {\isacharparenleft}{\kern0pt}\isanewline
\ \ let\isanewline
\ \ \ \ {\isacharparenleft}{\kern0pt}p{\isacharcomma}{\kern0pt}\ w{\isacharparenright}{\kern0pt}\ {\isacharequal}{\kern0pt}\ pw{\isacharsemicolon}{\kern0pt}\isanewline
\ \ \ \ w{\isacharprime}{\kern0pt}\ {\isacharequal}{\kern0pt}\ update{\isacharunderscore}{\kern0pt}weights\ t\ p{\isacharsemicolon}{\kern0pt}\isanewline
\ \ \ \ p{\isacharprime}{\kern0pt}\ {\isacharequal}{\kern0pt}\ update{\isacharunderscore}{\kern0pt}dec{\isacharunderscore}{\kern0pt}list\ w{\isacharprime}{\kern0pt}{\isacharsemicolon}{\kern0pt}\isanewline
\ \ \ \ err\ {\isacharequal}{\kern0pt}\ proj{\isacharunderscore}{\kern0pt}err{\isacharunderscore}{\kern0pt}w\ {\isacharparenleft}{\kern0pt}dec{\isacharunderscore}{\kern0pt}list{\isacharunderscore}{\kern0pt}to{\isacharunderscore}{\kern0pt}pol\ p{\isacharprime}{\kern0pt}{\isacharparenright}{\kern0pt}\ w{\isacharprime}{\kern0pt}{\isacharsemicolon}{\kern0pt}\isanewline
\ \ \ \ t{\isacharprime}{\kern0pt}\ {\isacharequal}{\kern0pt}\ t\ {\isacharplus}{\kern0pt}\ {\isadigit{1}}{\isacharsemicolon}{\kern0pt}\isanewline
\ \ \ \ w{\isacharunderscore}{\kern0pt}eq\ {\isacharequal}{\kern0pt}\ {\isacharparenleft}{\kern0pt}{\isasymforall}i\ {\isacharless}{\kern0pt}\ h{\isacharunderscore}{\kern0pt}dim{\isachardot}{\kern0pt}\ w{\isacharprime}{\kern0pt}\ i\ {\isacharequal}{\kern0pt}\ w\ i{\isacharparenright}{\kern0pt}{\isacharsemicolon}{\kern0pt}\isanewline
\ \ \ \ err{\isacharunderscore}{\kern0pt}le\ {\isacharequal}{\kern0pt}\ {\isacharparenleft}{\kern0pt}err\ {\isasymle}\ {\isasymepsilon}{\isacharparenright}{\kern0pt}{\isacharsemicolon}{\kern0pt}\isanewline
\ \ \ \ timeout\ {\isacharequal}{\kern0pt}\ {\isacharparenleft}{\kern0pt}t{\isacharprime}{\kern0pt}\ {\isasymge}\ t{\isacharunderscore}{\kern0pt}max{\isacharparenright}{\kern0pt}\ in\isanewline
\ \ \ \ {\isacharparenleft}{\kern0pt}if\ w{\isacharunderscore}{\kern0pt}eq\ {\isasymor}\ err{\isacharunderscore}{\kern0pt}le\ {\isasymor}\ timeout\ then\ {\isacharparenleft}{\kern0pt}w{\isacharprime}{\kern0pt}{\isacharcomma}{\kern0pt}\ p{\isacharprime}{\kern0pt}{\isacharcomma}{\kern0pt}\ err{\isacharcomma}{\kern0pt}\ t{\isacharcomma}{\kern0pt}\ w{\isacharunderscore}{\kern0pt}eq{\isacharcomma}{\kern0pt}\ err{\isacharunderscore}{\kern0pt}le{\isacharcomma}{\kern0pt}\ timeout{\isacharparenright}{\kern0pt}\isanewline
\ \ \ \ \ else\ api{\isacharunderscore}{\kern0pt}aux\ {\isacharparenleft}{\kern0pt}p{\isacharprime}{\kern0pt}{\isacharcomma}{\kern0pt}\ w{\isacharprime}{\kern0pt}{\isacharparenright}{\kern0pt}\ t{\isacharprime}{\kern0pt}{\isacharparenright}{\kern0pt}{\isacharparenright}{\kern0pt}{\isacartoucheclose}\isanewline
\isanewline
\isacommand{definition}\isamarkupfalse%
\ {\isacartoucheopen}api\ {\isacharequal}{\kern0pt}\isanewline
\ \ {\isacharparenleft}{\kern0pt}let\ \isanewline
\ \ \ \ w{\isadigit{0}}\ {\isacharequal}{\kern0pt}\ {\isacharparenleft}{\kern0pt}{\isasymlambda}{\isacharunderscore}{\kern0pt}{\isachardot}{\kern0pt}\ {\isadigit{0}}{\isacharparenright}{\kern0pt}{\isacharsemicolon}{\kern0pt}\isanewline
\ \ \ \ p{\isadigit{0}}\ {\isacharequal}{\kern0pt}\ update{\isacharunderscore}{\kern0pt}dec{\isacharunderscore}{\kern0pt}list\ w{\isadigit{0}}\isanewline
\ \ in\ api{\isacharunderscore}{\kern0pt}aux\ {\isacharparenleft}{\kern0pt}p{\isadigit{0}}{\isacharcomma}{\kern0pt}\ w{\isadigit{0}}{\isacharparenright}{\kern0pt}\ {\isadigit{0}}{\isacharparenright}{\kern0pt}{\isacartoucheclose}
\isanewline
\isanewline
\isakeyword{end}
\isanewline
\isanewline
\isacommand{locale}\isamarkupfalse%
\ API{\isacharunderscore}{\kern0pt}Interface\ {\isacharequal}{\kern0pt}\ \isanewline
\ \ API{\isacharunderscore}{\kern0pt}Interface{\isacharunderscore}{\kern0pt}Consts\ {\isacharplus}{\kern0pt}\isanewline
\ \ Factored{\isacharunderscore}{\kern0pt}MDP{\isacharunderscore}{\kern0pt}Default\ {\isacharplus}{\kern0pt}\isanewline
\ \ \isakeyword{assumes}\isanewline
\ \ \ \ update{\isacharunderscore}{\kern0pt}weights{\isacharunderscore}{\kern0pt}spec{\isacharcolon}{\kern0pt}\ {\isacartoucheopen}update{\isacharunderscore}{\kern0pt}weights{\isacharunderscore}{\kern0pt}spec{\isacartoucheclose}\ \isakeyword{and}\isanewline
\ \ \ \ dec{\isacharunderscore}{\kern0pt}list{\isacharunderscore}{\kern0pt}pol{\isacharunderscore}{\kern0pt}spec{\isacharcolon}{\kern0pt}\ {\isacartoucheopen}dec{\isacharunderscore}{\kern0pt}list{\isacharunderscore}{\kern0pt}pol{\isacharunderscore}{\kern0pt}spec{\isacartoucheclose}\isanewline
\isakeyword{begin}
\isanewline
\isanewline
\isacommand{theorem}\isamarkupfalse%
\ api{\isacharunderscore}{\kern0pt}correct{\isacharcolon}{\kern0pt}\isanewline
\ \ \isakeyword{assumes}\ {\isacartoucheopen}api\ {\isacharequal}{\kern0pt}\ {\isacharparenleft}{\kern0pt}w{\isacharcomma}{\kern0pt}\ p{\isacharcomma}{\kern0pt}\ err{\isacharcomma}{\kern0pt}\ t{\isacharcomma}{\kern0pt}\ True{\isacharcomma}{\kern0pt}\ err{\isacharunderscore}{\kern0pt}le{\isacharcomma}{\kern0pt}\ timeout{\isacharparenright}{\kern0pt}{\isacartoucheclose}\isanewline
\ \ \isakeyword{shows}\ {\isacartoucheopen}{\isacharparenleft}{\kern0pt}{\isadigit{1}}\ {\isacharminus}{\kern0pt}\ l{\isacharparenright}{\kern0pt}\ {\isacharasterisk}{\kern0pt}\ pol{\isacharunderscore}{\kern0pt}err\ {\isacharparenleft}{\kern0pt}dec{\isacharunderscore}{\kern0pt}list{\isacharunderscore}{\kern0pt}to{\isacharunderscore}{\kern0pt}pol\ p{\isacharparenright}{\kern0pt}\ {\isasymle}\ {\isadigit{2}}\ {\isacharasterisk}{\kern0pt}\ l\ {\isacharasterisk}{\kern0pt}\ err{\isacartoucheclose}\isanewline
\isanewline
\isakeyword{end}
\end{isasnipenv}

\subsubsection{Decision List Policy}
\begin{isasnipenv}[snip:decpoldef]{Decision List Policy Definitions}
\isacommand{definition}\isamarkupfalse%
\ U\isactrlsub a\ {\isacharcolon}{\kern0pt}{\isacharcolon}{\kern0pt}\ {\isacartoucheopen}{\isacharprime}{\kern0pt}a\ {\isasymRightarrow}\ nat\ set{\isacartoucheclose}\ \isakeyword{where}\isanewline
\ \ {\isacartoucheopen}U\isactrlsub a\ a\ {\isacharequal}{\kern0pt}\ {\isacharparenleft}{\kern0pt}{\isasymUnion}i\ {\isasymin}\ {\isacharbraceleft}{\kern0pt}reward{\isacharunderscore}{\kern0pt}dim\ d{\isachardot}{\kern0pt}{\isachardot}{\kern0pt}{\isacharless}{\kern0pt}reward{\isacharunderscore}{\kern0pt}dim\ a{\isacharbraceright}{\kern0pt}{\isachardot}{\kern0pt}\ reward{\isacharunderscore}{\kern0pt}scope\ a\ i{\isacharparenright}{\kern0pt}{\isacartoucheclose}\isanewline
\isanewline
\isacommand{definition}\isamarkupfalse%
\ {\isacartoucheopen}I\isactrlsub a\ a\ {\isacharequal}{\kern0pt}\ {\isacharbraceleft}{\kern0pt}i{\isachardot}{\kern0pt}\ i\ {\isacharless}{\kern0pt}\ h{\isacharunderscore}{\kern0pt}dim\ {\isasymand}\ effects\ a\ {\isasyminter}\ h{\isacharunderscore}{\kern0pt}scope\ i\ {\isasymnoteq}\ {\isacharbraceleft}{\kern0pt}{\isacharbraceright}{\kern0pt}{\isacharbraceright}{\kern0pt}{\isacartoucheclose}\isanewline
\isanewline
\isacommand{definition}\isamarkupfalse%
\ {\isasymGamma}\isactrlsub a\ {\isacharcolon}{\kern0pt}{\isacharcolon}{\kern0pt}\ {\isacartoucheopen}{\isacharprime}{\kern0pt}a\ {\isasymRightarrow}\ nat\ set\ {\isasymRightarrow}\ nat\ set{\isacartoucheclose}\ \isakeyword{where}\isanewline
\ \ {\isacartoucheopen}{\isasymGamma}\isactrlsub a\ a\ I\ {\isacharequal}{\kern0pt}\ {\isacharparenleft}{\kern0pt}{\isasymUnion}i\ {\isasymin}\ I{\isachardot}{\kern0pt}\ transitions{\isacharunderscore}{\kern0pt}scope\ a\ i{\isacharparenright}{\kern0pt}{\isacartoucheclose}\isanewline
\isanewline
\isacommand{definition}\isamarkupfalse%
\ {\isasymGamma}\isactrlsub a{\isacharprime}{\kern0pt}\ {\isacharcolon}{\kern0pt}{\isacharcolon}{\kern0pt}\ {\isacartoucheopen}{\isacharprime}{\kern0pt}a\ {\isasymRightarrow}\ nat\ set\ {\isasymRightarrow}\ nat\ set{\isacartoucheclose}\ \isakeyword{where}\isanewline
\ \ {\isacartoucheopen}{\isasymGamma}\isactrlsub a{\isacharprime}{\kern0pt}\ a\ X\ {\isacharequal}{\kern0pt}\ {\isasymGamma}\isactrlsub a\ a\ X\ {\isasymunion}\ {\isasymGamma}\isactrlsub a\ d\ X{\isacartoucheclose}\isanewline
\isanewline
\isacommand{definition}\isamarkupfalse%
\ T\isactrlsub a\ {\isacharcolon}{\kern0pt}{\isacharcolon}{\kern0pt}\ {\isacartoucheopen}{\isacharprime}{\kern0pt}a\ {\isasymRightarrow}\ nat\ set{\isacartoucheclose}\ \isakeyword{where}\isanewline
\ {\isacartoucheopen}T\isactrlsub a\ a\ {\isacharequal}{\kern0pt}\ U\isactrlsub a\ a\ {\isasymunion}\ {\isacharparenleft}{\kern0pt}{\isasymUnion}i\ {\isasymin}\ I\isactrlsub a\ a{\isachardot}{\kern0pt}\ {\isasymGamma}\isactrlsub a{\isacharprime}{\kern0pt}\ a\ {\isacharparenleft}{\kern0pt}h{\isacharunderscore}{\kern0pt}scope\ i{\isacharparenright}{\kern0pt}{\isacharparenright}{\kern0pt}{\isacartoucheclose}\isanewline
\isanewline
\isacommand{definition}\isamarkupfalse%
\ bonus\ {\isacharcolon}{\kern0pt}{\isacharcolon}{\kern0pt}\ {\isacartoucheopen}weights\ {\isasymRightarrow}\ {\isacharprime}{\kern0pt}a\ {\isasymRightarrow}\ {\isacharprime}{\kern0pt}x\ state\ {\isasymRightarrow}\ real{\isacartoucheclose}\ \isakeyword{where}\isanewline
\ \ {\isacartoucheopen}bonus\ w\ a\ x\ {\isacharequal}{\kern0pt}\ Q\ w\ x\ a\ {\isacharminus}{\kern0pt}\ Q\ w\ x\ d{\isacartoucheclose}%
\begin{isamarkuptext}%
Computationally efficient version of \isa{bonus}.%
\end{isamarkuptext}\isamarkuptrue%
\isacommand{definition}\isamarkupfalse%
\ {\isacartoucheopen}bonus{\isacharprime}{\kern0pt}\ w\ a\ x\ {\isacharequal}{\kern0pt}\ R\isactrlsub a\ a\ x\ {\isacharplus}{\kern0pt}\ l\ {\isacharasterisk}{\kern0pt}\ {\isacharparenleft}{\kern0pt}{\isasymSum}i\ {\isasymin}\ I\isactrlsub a\ a{\isachardot}{\kern0pt}\ w\ i\ {\isacharasterisk}{\kern0pt}\ {\isacharparenleft}{\kern0pt}g{\isacharprime}{\kern0pt}\ i\ a\ x\ {\isacharminus}{\kern0pt}\ g{\isacharprime}{\kern0pt}\ i\ d\ x{\isacharparenright}{\kern0pt}{\isacharparenright}{\kern0pt}{\isacartoucheclose}%
\isanewline
\isanewline
\isacommand{definition}\isamarkupfalse%
\ dec{\isacharunderscore}{\kern0pt}list{\isacharunderscore}{\kern0pt}act\ {\isacharcolon}{\kern0pt}{\isacharcolon}{\kern0pt}\ {\isacartoucheopen}weights\ {\isasymRightarrow}\ {\isacharprime}{\kern0pt}a\ {\isasymRightarrow}\ {\isacharparenleft}{\kern0pt}{\isacharprime}{\kern0pt}x\ state\ {\isasymtimes}\ {\isacharprime}{\kern0pt}a\ {\isasymtimes}\ real{\isacharparenright}{\kern0pt}\ list{\isacartoucheclose}\ \isakeyword{where}\isanewline
\ \ {\isacartoucheopen}dec{\isacharunderscore}{\kern0pt}list{\isacharunderscore}{\kern0pt}act\ w\ a\ {\isacharequal}{\kern0pt}\ {\isacharparenleft}{\kern0pt}\isanewline
\ \ let\isanewline
\ \ \ \ ts\ {\isacharequal}{\kern0pt}\ assignment{\isacharunderscore}{\kern0pt}list\ {\isacharparenleft}{\kern0pt}T\isactrlsub a\ a{\isacharparenright}{\kern0pt}{\isacharsemicolon}{\kern0pt}\isanewline
\ \ \ \ ts{\isacharprime}{\kern0pt}\ {\isacharequal}{\kern0pt}\ filter\ {\isacharparenleft}{\kern0pt}{\isasymlambda}t{\isachardot}{\kern0pt}\ bonus{\isacharprime}{\kern0pt}\ w\ a\ t\ {\isachargreater}{\kern0pt}\ {\isadigit{0}}{\isacharparenright}{\kern0pt}\ ts{\isacharsemicolon}{\kern0pt}\isanewline
\ \ \ \ ts{\isacharprime}{\kern0pt}{\isacharprime}{\kern0pt}\ {\isacharequal}{\kern0pt}\ map\ {\isacharparenleft}{\kern0pt}{\isasymlambda}t{\isachardot}{\kern0pt}\ {\isacharparenleft}{\kern0pt}t{\isacharcomma}{\kern0pt}\ a{\isacharcomma}{\kern0pt}\ bonus{\isacharprime}{\kern0pt}\ w\ a\ t{\isacharparenright}{\kern0pt}{\isacharparenright}{\kern0pt}\ ts{\isacharprime}{\kern0pt}\isanewline
\ \ in\isanewline
\ \ \ \ ts{\isacharprime}{\kern0pt}{\isacharprime}{\kern0pt}\isanewline
{\isacharparenright}{\kern0pt}{\isacartoucheclose}\isanewline
\isanewline
\isacommand{abbreviation}\isamarkupfalse%
\ {\isacartoucheopen}{\isasympi}{\isacharunderscore}{\kern0pt}unsorted\ w\ {\isasymequiv}\ {\isacharparenleft}{\kern0pt}{\isasymlambda}{\isacharunderscore}{\kern0pt}{\isachardot}{\kern0pt}\ None{\isacharcomma}{\kern0pt}\ d{\isacharcomma}{\kern0pt}\ {\isadigit{0}}{\isacharparenright}{\kern0pt}\ {\isacharhash}{\kern0pt}\ concat\ {\isacharparenleft}{\kern0pt}map\ {\isacharparenleft}{\kern0pt}dec{\isacharunderscore}{\kern0pt}list{\isacharunderscore}{\kern0pt}act\ w{\isacharparenright}{\kern0pt}\ actions{\isacharunderscore}{\kern0pt}nondef{\isacharparenright}{\kern0pt}{\isacartoucheclose}\isanewline
\isacommand{abbreviation}\isamarkupfalse%
\ {\isacartoucheopen}{\isasympi}{\isacharunderscore}{\kern0pt}sorted\ w\ {\isasymequiv}\ sort{\isacharunderscore}{\kern0pt}dec{\isacharunderscore}{\kern0pt}list\ {\isacharparenleft}{\kern0pt}{\isasympi}{\isacharunderscore}{\kern0pt}unsorted\ w{\isacharparenright}{\kern0pt}{\isacartoucheclose}\isanewline
\isanewline
\isacommand{definition}\isamarkupfalse%
\ dec{\isacharunderscore}{\kern0pt}list{\isacharunderscore}{\kern0pt}pol\ {\isacharcolon}{\kern0pt}{\isacharcolon}{\kern0pt}\ {\isacartoucheopen}weights\ {\isasymRightarrow}\ {\isacharparenleft}{\kern0pt}{\isacharprime}{\kern0pt}x\ state\ {\isasymtimes}\ {\isacharprime}{\kern0pt}a{\isacharparenright}{\kern0pt}\ list{\isacartoucheclose}\ \isakeyword{where}\ \isanewline
\ \ {\isacartoucheopen}dec{\isacharunderscore}{\kern0pt}list{\isacharunderscore}{\kern0pt}pol\ w\ {\isacharequal}{\kern0pt}\ map\ {\isacharparenleft}{\kern0pt}{\isasymlambda}{\isacharparenleft}{\kern0pt}t{\isacharcomma}{\kern0pt}\ a{\isacharcomma}{\kern0pt}\ {\isacharunderscore}{\kern0pt}{\isacharparenright}{\kern0pt}{\isachardot}{\kern0pt}\ {\isacharparenleft}{\kern0pt}t{\isacharcomma}{\kern0pt}\ a{\isacharparenright}{\kern0pt}{\isacharparenright}{\kern0pt}\ {\isacharparenleft}{\kern0pt}{\isasympi}{\isacharunderscore}{\kern0pt}sorted\ w{\isacharparenright}{\kern0pt}{\isacartoucheclose}\isanewline
\isanewline
\isacommand{definition}\isamarkupfalse%
\ dec{\isacharunderscore}{\kern0pt}list{\isacharunderscore}{\kern0pt}pol{\isacharunderscore}{\kern0pt}sel\ {\isacharcolon}{\kern0pt}{\isacharcolon}{\kern0pt}\ {\isacartoucheopen}weights\ {\isasymRightarrow}\ {\isacharprime}{\kern0pt}x\ state\ {\isasymRightarrow}\ {\isacharprime}{\kern0pt}a{\isacartoucheclose}\ \isakeyword{where}\isanewline
\ \ {\isacartoucheopen}dec{\isacharunderscore}{\kern0pt}list{\isacharunderscore}{\kern0pt}pol{\isacharunderscore}{\kern0pt}sel\ w\ {\isacharequal}{\kern0pt}\ dec{\isacharunderscore}{\kern0pt}list{\isacharunderscore}{\kern0pt}to{\isacharunderscore}{\kern0pt}pol\ {\isacharparenleft}{\kern0pt}dec{\isacharunderscore}{\kern0pt}list{\isacharunderscore}{\kern0pt}pol\ w{\isacharparenright}{\kern0pt}{\isacartoucheclose}
\end{isasnipenv}

\begin{isasnipenv}[snip:polprop]{Properties of the Decision List Policy}
\begin{isamarkuptext}%
The main property of \isa{dec{\isacharunderscore}{\kern0pt}list{\isacharunderscore}{\kern0pt}pol} is: Given a state, the first element of the policy it is
consistent with is the greedy action in that state.%
\end{isamarkuptext}\isamarkuptrue%
\isacommand{lemma}\isamarkupfalse%
\ dec{\isacharunderscore}{\kern0pt}list{\isacharunderscore}{\kern0pt}pol{\isacharunderscore}{\kern0pt}sel{\isacharunderscore}{\kern0pt}greedy{\isacharcolon}{\kern0pt}\isanewline
\ \ \isakeyword{assumes}\ {\isacartoucheopen}x\ {\isasymin}\ states{\isacartoucheclose}\isanewline
\ \ \isakeyword{shows}\ {\isacartoucheopen}is{\isacharunderscore}{\kern0pt}greedy{\isacharunderscore}{\kern0pt}act\ w\ x\ {\isacharparenleft}{\kern0pt}dec{\isacharunderscore}{\kern0pt}list{\isacharunderscore}{\kern0pt}pol{\isacharunderscore}{\kern0pt}sel\ w\ x{\isacharparenright}{\kern0pt}{\isacartoucheclose}\isanewline
\isanewline
\isacommand{lemma}\isamarkupfalse%
\ dec{\isacharunderscore}{\kern0pt}list{\isacharunderscore}{\kern0pt}pol{\isacharunderscore}{\kern0pt}is{\isacharunderscore}{\kern0pt}pol{\isacharcolon}{\kern0pt}\ {\isacartoucheopen}is{\isacharunderscore}{\kern0pt}dec{\isacharunderscore}{\kern0pt}list{\isacharunderscore}{\kern0pt}pol\ {\isacharparenleft}{\kern0pt}dec{\isacharunderscore}{\kern0pt}list{\isacharunderscore}{\kern0pt}pol\ w{\isacharparenright}{\kern0pt}{\isacartoucheclose}\isanewline
\end{isasnipenv}

\subsubsection{Bellman Error}
\begin{isasnipenv}[snip:err]{Bellman Error}
\isacommand{locale}\isamarkupfalse%
\ Bellman{\isacharunderscore}{\kern0pt}Err{\isacharunderscore}{\kern0pt}Consts\ {\isacharequal}{\kern0pt}\isanewline
\ \ Factored{\isacharunderscore}{\kern0pt}MDP{\isacharunderscore}{\kern0pt}Default{\isacharunderscore}{\kern0pt}Consts\ \isanewline
\ \ \isakeyword{where}\ rewards\ {\isacharequal}{\kern0pt}\ rewards\ \isakeyword{for}\isanewline
\ \ rewards\ {\isacharcolon}{\kern0pt}{\isacharcolon}{\kern0pt}\ {\isachardoublequoteopen}{\isacharprime}{\kern0pt}a{\isacharcolon}{\kern0pt}{\isacharcolon}{\kern0pt}{\isacharbraceleft}{\kern0pt}countable{\isacharcomma}{\kern0pt}\ linorder{\isacharbraceright}{\kern0pt}\ {\isasymRightarrow}\ nat\ {\isasymRightarrow}\ {\isacharprime}{\kern0pt}x{\isacharcolon}{\kern0pt}{\isacharcolon}{\kern0pt}{\isacharbraceleft}{\kern0pt}countable{\isacharcomma}{\kern0pt}\ linorder{\isacharbraceright}{\kern0pt}\ state\ {\isasymRightarrow}\ real{\isachardoublequoteclose}\ {\isacharplus}{\kern0pt}\ \isanewline
\isakeyword{fixes}\isanewline
\ \ error{\isacharunderscore}{\kern0pt}branch\ {\isacharcolon}{\kern0pt}{\isacharcolon}{\kern0pt}\ {\isacartoucheopen}{\isacharprime}{\kern0pt}x\ state\ {\isasymRightarrow}\ {\isacharprime}{\kern0pt}a\ {\isasymRightarrow}\ {\isacharparenleft}{\kern0pt}{\isacharprime}{\kern0pt}x\ state{\isacharparenright}{\kern0pt}\ list\ {\isasymRightarrow}\ ereal{\isacartoucheclose}\ \isakeyword{and}\isanewline
\ \ dec{\isacharunderscore}{\kern0pt}pol\ {\isacharcolon}{\kern0pt}{\isacharcolon}{\kern0pt}\ {\isacartoucheopen}{\isacharparenleft}{\kern0pt}{\isacharprime}{\kern0pt}x\ state\ {\isasymtimes}\ {\isacharprime}{\kern0pt}a{\isacharparenright}{\kern0pt}\ list{\isacartoucheclose}\ \isakeyword{and}\isanewline
\ \ w\ {\isacharcolon}{\kern0pt}{\isacharcolon}{\kern0pt}\ {\isacartoucheopen}weights{\isacartoucheclose}\isanewline
\isakeyword{begin}\isanewline
\isanewline
\isacommand{definition}\isamarkupfalse%
\ {\isacartoucheopen}greedy{\isacharunderscore}{\kern0pt}spec\ {\isacharequal}\isanewline\ {\kern0pt}\ {\isacharparenleft}{\kern0pt}{\isasymforall}x\ {\isasymin}\ states{\isachardot}{\kern0pt}\ is{\isacharunderscore}{\kern0pt}greedy{\isacharunderscore}{\kern0pt}act\ w\ x\ {\isacharparenleft}{\kern0pt}dec{\isacharunderscore}{\kern0pt}list{\isacharunderscore}{\kern0pt}to{\isacharunderscore}{\kern0pt}pol\ dec{\isacharunderscore}{\kern0pt}pol\ x{\isacharparenright}{\kern0pt}{\isacharparenright}{\kern0pt}{\isacartoucheclose}\isanewline
\isanewline
\isacommand{definition}\isamarkupfalse%
\ {\isacartoucheopen}dec{\isacharunderscore}{\kern0pt}pol{\isacharunderscore}{\kern0pt}spec\ {\isasymlongleftrightarrow}\ is{\isacharunderscore}{\kern0pt}dec{\isacharunderscore}{\kern0pt}list{\isacharunderscore}{\kern0pt}pol\ dec{\isacharunderscore}{\kern0pt}pol\ {\isasymand}\ greedy{\isacharunderscore}{\kern0pt}spec{\isacartoucheclose}\isanewline
\isanewline
\isacommand{definition}\isamarkupfalse%
\ {\isacartoucheopen}error{\isacharunderscore}{\kern0pt}branch{\isacharunderscore}{\kern0pt}spec\ {\isacharequal}{\kern0pt}\ {\isacharparenleft}{\kern0pt}{\isasymforall}t\ a\ ts{\isachardot}{\kern0pt}\ \isanewline
\ \ a\ {\isasymin}\ actions\ {\isasymlongrightarrow}\ t\ {\isasymin}\ partial{\isacharunderscore}{\kern0pt}states\ {\isasymlongrightarrow}\ set\ ts\ {\isasymsubseteq}\ partial{\isacharunderscore}{\kern0pt}states\ {\isasymlongrightarrow}\isanewline
\ \ error{\isacharunderscore}{\kern0pt}branch\ t\ a\ ts\ {\isacharequal}{\kern0pt}
\isanewline
\ \ {\isacharparenleft}{\kern0pt}{\isasymSqunion}\ x\ {\isasymin}\ {\isacharbraceleft}{\kern0pt}x{\isachardot}{\kern0pt}\ x\ {\isasymin}\ states\ {\isasymand}\ consistent\ x\ t\ {\isasymand}\ list{\isacharunderscore}{\kern0pt}all\ {\isacharparenleft}{\kern0pt}{\isasymlambda}t{\isacharprime}{\kern0pt}{\isachardot}{\kern0pt}\ {\isasymnot}consistent\ x\ t{\isacharprime}{\kern0pt}{\isacharparenright}{\kern0pt}\ ts{\isacharbraceright}{\kern0pt}{\isachardot}{\kern0pt}\isanewline
\ \ \ \ ereal\ {\isacharparenleft}{\kern0pt}dist\ {\isacharparenleft}{\kern0pt}Q\ w\ x\ a{\isacharparenright}{\kern0pt}\ {\isacharparenleft}{\kern0pt}{\isasymnu}\isactrlsub w\ w\ x{\isacharparenright}{\kern0pt}{\isacharparenright}{\kern0pt}{\isacharparenright}{\kern0pt}{\isacharparenright}{\kern0pt}{\isacartoucheclose}\isanewline
\isanewline
\isacommand{definition}\isamarkupfalse%
\ {\isacartoucheopen}update{\isacharunderscore}{\kern0pt}err{\isacharunderscore}{\kern0pt}iter\ {\isacharequal}{\kern0pt}
\isanewline
\ \ {\isacharparenleft}{\kern0pt}{\isasymlambda}{\isacharparenleft}{\kern0pt}t{\isacharcomma}{\kern0pt}\ a{\isacharparenright}{\kern0pt}\ {\isacharparenleft}{\kern0pt}ts{\isacharcomma}{\kern0pt}\ err{\isacharparenright}{\kern0pt}{\isachardot}{\kern0pt}\ {\isacharparenleft}{\kern0pt}t{\isacharhash}{\kern0pt}ts{\isacharcomma}{\kern0pt}\ sup\ {\isacharparenleft}{\kern0pt}error{\isacharunderscore}{\kern0pt}branch\ t\ a\ ts{\isacharparenright}{\kern0pt}\ err{\isacharparenright}{\kern0pt}{\isacharparenright}{\kern0pt}{\isacartoucheclose}\isanewline
\isanewline
\isacommand{definition}\isamarkupfalse%
\ {\isacartoucheopen}err{\isacharunderscore}{\kern0pt}list\ xs\ {\isacharequal}{\kern0pt}\ snd\ {\isacharparenleft}{\kern0pt}fold\ update{\isacharunderscore}{\kern0pt}err{\isacharunderscore}{\kern0pt}iter\ xs\ {\isacharparenleft}{\kern0pt}{\isacharbrackleft}{\kern0pt}{\isacharbrackright}{\kern0pt}{\isacharcomma}{\kern0pt}\ {\isadigit{0}}{\isacharparenright}{\kern0pt}{\isacharparenright}{\kern0pt}{\isacartoucheclose}\isanewline
\isanewline
\isacommand{definition}\isamarkupfalse%
\ {\isacartoucheopen}factored{\isacharunderscore}{\kern0pt}bellman{\isacharunderscore}{\kern0pt}err\ {\isacharequal}{\kern0pt}\ real{\isacharunderscore}{\kern0pt}of{\isacharunderscore}{\kern0pt}ereal\ {\isacharparenleft}{\kern0pt}err{\isacharunderscore}{\kern0pt}list\ dec{\isacharunderscore}{\kern0pt}pol{\isacharparenright}{\kern0pt}{\isacartoucheclose}\isanewline
\isanewline
\isacommand{end}\isamarkupfalse%
\end{isasnipenv}

\begin{isasnipenv}[snip:errprop]{Factored Bellmann Error Correctness}
\isacommand{lemma}\isamarkupfalse%
\ bellman{\isacharunderscore}{\kern0pt}err{\isacharunderscore}{\kern0pt}w{\isacharunderscore}{\kern0pt}eq{\isacharunderscore}{\kern0pt}Q{\isacharcolon}{\kern0pt}\isanewline
\ \ \isakeyword{shows}\ {\isacartoucheopen}bellman{\isacharunderscore}{\kern0pt}err{\isacharunderscore}{\kern0pt}w\ w\ {\isacharequal}\isanewline
\ \ {\kern0pt}\ {\isacharparenleft}{\kern0pt}{\isasymSqunion}x{\isasymin}states{\isachardot}{\kern0pt}\ dist\ {\isacharparenleft}{\kern0pt}Q\ w\ x\ {\isacharparenleft}{\kern0pt}dec{\isacharunderscore}{\kern0pt}list{\isacharunderscore}{\kern0pt}to{\isacharunderscore}{\kern0pt}pol\ dec{\isacharunderscore}{\kern0pt}pol\ x{\isacharparenright}{\kern0pt}{\isacharparenright}{\kern0pt}\ {\isacharparenleft}{\kern0pt}{\isasymnu}\isactrlsub w\ w\ x{\isacharparenright}{\kern0pt}{\isacharparenright}{\kern0pt}{\isacartoucheclose}
\end{isasnipenv}

\begin{isasnipenv}[snip:brancherr]{Error of a Single Branch}
\isacommand{locale}\isamarkupfalse%
\ Bellman{\isacharunderscore}{\kern0pt}Err{\isacharunderscore}{\kern0pt}Branch{\isacharunderscore}{\kern0pt}Consts\ {\isacharequal}{\kern0pt}\ Factored{\isacharunderscore}{\kern0pt}MDP{\isacharunderscore}{\kern0pt}Default{\isacharunderscore}{\kern0pt}Consts\isanewline
\ \ \isakeyword{where}\ \isanewline
\ \ \ \ rewards\ {\isacharequal}{\kern0pt}\ rewards\isanewline
\ \ \isakeyword{for}\isanewline
\ \ \ \ rewards\ {\isacharcolon}{\kern0pt}{\isacharcolon}{\kern0pt}\ {\isacartoucheopen}{\isacharprime}{\kern0pt}a{\isacharcolon}{\kern0pt}{\isacharcolon}{\kern0pt}{\isacharbraceleft}{\kern0pt}countable{\isacharcomma}{\kern0pt}\ linorder{\isacharbraceright}{\kern0pt}\ {\isasymRightarrow}\ nat
\isanewline
\ \ \ \ \ \ {\isasymRightarrow}\ {\isacharparenleft}{\kern0pt}{\isacharprime}{\kern0pt}x{\isacharcolon}{\kern0pt}{\isacharcolon}{\kern0pt}{\isacharbraceleft}{\kern0pt}countable{\isacharcomma}{\kern0pt}\ linorder{\isacharbraceright}{\kern0pt}\ state{\isacharparenright}{\kern0pt}\ {\isasymRightarrow}\ real{\isacartoucheclose}\ {\isacharplus}{\kern0pt}\isanewline
\ \ \isakeyword{fixes}\isanewline
\ \ \ \ variable{\isacharunderscore}{\kern0pt}elim\ {\isacharcolon}{\kern0pt}{\isacharcolon}{\kern0pt}\ {\isacartoucheopen}{\isacharparenleft}{\kern0pt}{\isacharparenleft}{\kern0pt}{\isacharprime}{\kern0pt}x\ state\ {\isasymRightarrow}\ ereal{\isacharparenright}{\kern0pt}\ {\isasymtimes}\ {\isacharparenleft}{\kern0pt}nat\ set{\isacharparenright}{\kern0pt}{\isacharparenright}{\kern0pt}\ list\ {\isasymRightarrow}\ ereal{\isacartoucheclose}\isanewline
\ \ \ \ %
\isamarkupcmt{Variable elimination algorithm, maximizes sum of input functions%
}\isanewline
\isakeyword{begin}%
\begin{isamarkuptext}%
Specification of the variable elimination algorithm.%
\end{isamarkuptext}\isamarkuptrue%
\isacommand{definition}\isamarkupfalse%
\ {\isacartoucheopen}variable{\isacharunderscore}{\kern0pt}elim{\isacharunderscore}{\kern0pt}correct\ fs\ {\isasymlongleftrightarrow}\ \isanewline
\ \ variable{\isacharunderscore}{\kern0pt}elim\ fs\ {\isacharequal}{\kern0pt}\ {\isacharparenleft}{\kern0pt}MAX\ x\ {\isasymin}\ states{\isachardot}{\kern0pt}\ sum{\isacharunderscore}{\kern0pt}list\ {\isacharparenleft}{\kern0pt}map\ {\isacharparenleft}{\kern0pt}{\isasymlambda}{\isacharparenleft}{\kern0pt}f{\isacharcomma}{\kern0pt}\ s{\isacharparenright}{\kern0pt}{\isachardot}{\kern0pt}\ f\ x{\isacharparenright}{\kern0pt}\ fs{\isacharparenright}{\kern0pt}{\isacharparenright}{\kern0pt}{\isacartoucheclose}%
\begin{isamarkuptext}%
Valid input functions: don't take (positive) infinite values + scopes are valid.%
\end{isamarkuptext}\isamarkuptrue%
\isacommand{definition}\isamarkupfalse%
\ {\isacartoucheopen}scope{\isacharunderscore}{\kern0pt}inv\ b\ {\isasymlongleftrightarrow}\ {\isacharparenleft}{\kern0pt}{\isasymforall}{\isacharparenleft}{\kern0pt}b{\isacharcomma}{\kern0pt}\ scope{\isacharunderscore}{\kern0pt}b{\isacharparenright}{\kern0pt}\ {\isasymin}\ set\ b{\isachardot}{\kern0pt}\ \isanewline
\ \ {\isacharparenleft}{\kern0pt}{\isasymforall}x{\isachardot}{\kern0pt}\ b\ x\ {\isasymnoteq}\ {\isasyminfinity}{\isacharparenright}{\kern0pt}\ {\isasymand}\ scope{\isacharunderscore}{\kern0pt}b\ {\isasymsubseteq}\ vars\ {\isasymand}\ has{\isacharunderscore}{\kern0pt}scope{\isacharunderscore}{\kern0pt}on\ b\ partial{\isacharunderscore}{\kern0pt}states\ scope{\isacharunderscore}{\kern0pt}b{\isacharparenright}{\kern0pt}{\isacartoucheclose}\isanewline
\isanewline
\isacommand{definition}\isamarkupfalse%
\ {\isacartoucheopen}scope{\isacharunderscore}{\kern0pt}inv{\isacharprime}{\kern0pt}\ b\ {\isasymlongleftrightarrow}\ {\isacharparenleft}{\kern0pt}{\isasymforall}{\isacharparenleft}{\kern0pt}b{\isacharcomma}{\kern0pt}\ scope{\isacharunderscore}{\kern0pt}b{\isacharparenright}{\kern0pt}\ {\isasymin}\ set\ b{\isachardot}{\kern0pt}\ \isanewline
\ \ {\isacharparenleft}{\kern0pt}{\isasymforall}x{\isachardot}{\kern0pt}\ b\ x\ {\isasymnoteq}\ {\isacharminus}{\kern0pt}{\isasyminfinity}\ {\isasymand}\ b\ x\ {\isasymnoteq}\ {\isasyminfinity}{\isacharparenright}{\kern0pt}\ {\isasymand}\ scope{\isacharunderscore}{\kern0pt}b\ {\isasymsubseteq}\ vars\ {\isasymand}\ has{\isacharunderscore}{\kern0pt}scope{\isacharunderscore}{\kern0pt}on\ b\ partial{\isacharunderscore}{\kern0pt}states\ scope{\isacharunderscore}{\kern0pt}b{\isacharparenright}{\kern0pt}{\isacartoucheclose}\isanewline
\isanewline
\isacommand{definition}\isamarkupfalse%
\ {\isacartoucheopen}variable{\isacharunderscore}{\kern0pt}elim{\isacharunderscore}{\kern0pt}spec\ {\isasymlongleftrightarrow}\ {\isacharparenleft}{\kern0pt}{\isasymforall}fs{\isachardot}{\kern0pt}\ scope{\isacharunderscore}{\kern0pt}inv\ fs\ {\isasymlongrightarrow}\ variable{\isacharunderscore}{\kern0pt}elim{\isacharunderscore}{\kern0pt}correct\ fs{\isacharparenright}{\kern0pt}{\isacartoucheclose}%

\begin{isamarkuptext}%
Functions to maximize that depend on the weights.%
\end{isamarkuptext}\isamarkuptrue%
\isacommand{definition}\isamarkupfalse%
\ {\isacartoucheopen}hg{\isacharunderscore}{\kern0pt}scope\ t\ a\ i\ {\isacharequal}{\kern0pt}\ h{\isacharunderscore}{\kern0pt}scope\ i\ {\isasymunion}\ {\isasymGamma}\isactrlsub a\ a\ {\isacharparenleft}{\kern0pt}h{\isacharunderscore}{\kern0pt}scope\ i{\isacharparenright}{\kern0pt}\ {\isacharminus}{\kern0pt}\ dom\ t{\isacartoucheclose}\isanewline
\isacommand{definition}\isamarkupfalse%
\ {\isacartoucheopen}hg{\isacharunderscore}{\kern0pt}inst\ w\ t\ a\ i\ {\isacharequal}{\kern0pt}\ instantiate\ {\isacharparenleft}{\kern0pt}{\isasymlambda}x{\isachardot}{\kern0pt}\ w\ i\ {\isacharasterisk}{\kern0pt}\ {\isacharparenleft}{\kern0pt}h\ i\ x\ {\isacharminus}{\kern0pt}\ l\ {\isacharasterisk}{\kern0pt}\ g{\isacharprime}{\kern0pt}\ i\ a\ x{\isacharparenright}{\kern0pt}{\isacharparenright}{\kern0pt}\ t{\isacartoucheclose}\isanewline
\isanewline
\isacommand{definition}\isamarkupfalse%
\ {\isacartoucheopen}C\ w\ t\ a\ {\isacharequal}{\kern0pt}\ {\isacharparenleft}{\kern0pt}map\ {\isacharparenleft}{\kern0pt}{\isasymlambda}i{\isachardot}{\kern0pt}\ {\isacharparenleft}{\kern0pt}hg{\isacharunderscore}{\kern0pt}inst\ w\ t\ a\ i{\isacharcomma}{\kern0pt}\ hg{\isacharunderscore}{\kern0pt}scope\ t\ a\ i{\isacharparenright}{\kern0pt}{\isacharparenright}{\kern0pt}\ {\isacharbrackleft}{\kern0pt}{\isadigit{0}}{\isachardot}{\kern0pt}{\isachardot}{\kern0pt}{\isacharless}{\kern0pt}h{\isacharunderscore}{\kern0pt}dim{\isacharbrackright}{\kern0pt}{\isacharparenright}{\kern0pt}{\isacartoucheclose}\isanewline
\isacommand{definition}\isamarkupfalse%
\ {\isacartoucheopen}neg{\isacharunderscore}{\kern0pt}C\ w\ t\ a\ {\isacharequal}{\kern0pt}\ map\ neg{\isacharunderscore}{\kern0pt}scoped\ {\isacharparenleft}{\kern0pt}C\ w\ t\ a{\isacharparenright}{\kern0pt}{\isacartoucheclose}%
\begin{isamarkuptext}%
Functions to maximize that independent of the weights.%
\end{isamarkuptext}\isamarkuptrue%
\isacommand{definition}\isamarkupfalse%
\ {\isacartoucheopen}r{\isacharunderscore}{\kern0pt}act{\isacharunderscore}{\kern0pt}dim\ a\ {\isacharequal}{\kern0pt}\ reward{\isacharunderscore}{\kern0pt}dim\ a\ {\isacharminus}{\kern0pt}\ reward{\isacharunderscore}{\kern0pt}dim\ d{\isacartoucheclose}\isanewline
\isacommand{definition}\isamarkupfalse%
\ {\isacartoucheopen}r{\isacharunderscore}{\kern0pt}scope\ t\ a\ i\ {\isacharequal}{\kern0pt}\ reward{\isacharunderscore}{\kern0pt}scope\ a\ i\ {\isacharminus}{\kern0pt}\ dom\ t{\isacartoucheclose}\isanewline
\isacommand{definition}\isamarkupfalse%
\ {\isacartoucheopen}r{\isacharunderscore}{\kern0pt}inst\ t\ a\ i\ {\isacharequal}{\kern0pt}\ instantiate\ {\isacharparenleft}{\kern0pt}rewards\ a\ i{\isacharparenright}{\kern0pt}\ t{\isacartoucheclose}\isanewline
\isanewline
\isacommand{definition}\isamarkupfalse%
\ {\isacartoucheopen}b\ t\ a\ {\isacharequal}{\kern0pt}\ {\isacharparenleft}{\kern0pt}map\ {\isacharparenleft}{\kern0pt}{\isasymlambda}i{\isachardot}{\kern0pt}\ {\isacharparenleft}{\kern0pt}r{\isacharunderscore}{\kern0pt}inst\ t\ a\ i{\isacharcomma}{\kern0pt}\ r{\isacharunderscore}{\kern0pt}scope\ t\ a\ i{\isacharparenright}{\kern0pt}{\isacharparenright}{\kern0pt}\ {\isacharbrackleft}{\kern0pt}{\isadigit{0}}{\isachardot}{\kern0pt}{\isachardot}{\kern0pt}{\isacharless}{\kern0pt}reward{\isacharunderscore}{\kern0pt}dim\ a{\isacharbrackright}{\kern0pt}{\isacharparenright}{\kern0pt}{\isacartoucheclose}\isanewline
\isacommand{definition}\isamarkupfalse%
\ {\isacartoucheopen}neg{\isacharunderscore}{\kern0pt}b\ t\ a\ {\isacharequal}{\kern0pt}\ map\ neg{\isacharunderscore}{\kern0pt}scoped\ {\isacharparenleft}{\kern0pt}b\ t\ a{\isacharparenright}{\kern0pt}{\isacartoucheclose}%
\begin{isamarkuptext}%
Indicator functions that make us ignore states that choose earlier branches.%
\end{isamarkuptext}\isamarkuptrue%
\isacommand{definition}\isamarkupfalse%
\ {\isacartoucheopen}{\isasymI}\ t\ x\ {\isacharequal}{\kern0pt}\ {\isacharparenleft}{\kern0pt}if\ consistent\ x\ t\ then\ {\isacharminus}{\kern0pt}{\isasyminfinity}{\isacharcolon}{\kern0pt}{\isacharcolon}{\kern0pt}ereal\ else\ {\isadigit{0}}{\isacharparenright}{\kern0pt}{\isacartoucheclose}\isanewline
\isacommand{definition}\isamarkupfalse%
\ {\isacartoucheopen}{\isasymI}{\isacharprime}{\kern0pt}\ ts\ t{\isacharprime}{\kern0pt}\ {\isacharequal}{\kern0pt}\ map\ {\isacharparenleft}{\kern0pt}{\isasymlambda}t{\isachardot}{\kern0pt}\ {\isacharparenleft}{\kern0pt}instantiate\ {\isacharparenleft}{\kern0pt}{\isasymI}\ t{\isacharparenright}{\kern0pt}\ t{\isacharprime}{\kern0pt}{\isacharcomma}{\kern0pt}\ dom\ t\ {\isacharminus}{\kern0pt}\ dom\ t{\isacharprime}{\kern0pt}{\isacharparenright}{\kern0pt}{\isacharparenright}{\kern0pt}\ ts{\isacartoucheclose}\isanewline
\isacommand{definition}\isamarkupfalse%
\ {\isacartoucheopen}scope{\isacharunderscore}{\kern0pt}{\isasymI}{\isacharprime}{\kern0pt}\ ts\ t{\isacharprime}{\kern0pt}\ {\isacharequal}{\kern0pt}\ {\isasymUnion}{\isacharparenleft}{\kern0pt}set\ {\isacharparenleft}{\kern0pt}map\ snd\ {\isacharparenleft}{\kern0pt}{\isasymI}{\isacharprime}{\kern0pt}\ ts\ t{\isacharprime}{\kern0pt}{\isacharparenright}{\kern0pt}{\isacharparenright}{\kern0pt}{\isacharparenright}{\kern0pt}{\isacartoucheclose}\isanewline
\isanewline
\isacommand{definition}\isamarkupfalse%
\ {\isacartoucheopen}b{\isacharprime}{\kern0pt}\ t\ a\ ts\ {\isacharequal}{\kern0pt}\ {\isacharparenleft}{\kern0pt}b\ t\ a{\isacharparenright}{\kern0pt}\ {\isacharat}{\kern0pt}\ \ {\isasymI}{\isacharprime}{\kern0pt}\ ts\ t{\isacartoucheclose}\isanewline
\isacommand{definition}\isamarkupfalse%
\ {\isacartoucheopen}neg{\isacharunderscore}{\kern0pt}b{\isacharprime}{\kern0pt}\ t\ a\ ts\ {\isacharequal}{\kern0pt}\ {\isacharparenleft}{\kern0pt}neg{\isacharunderscore}{\kern0pt}b\ t\ a{\isacharparenright}{\kern0pt}\ {\isacharat}{\kern0pt}\ \ {\isasymI}{\isacharprime}{\kern0pt}\ ts\ t{\isacartoucheclose}%
\begin{isamarkuptext}%
The maximum of the negative + positive error is the absolute error.%
\end{isamarkuptext}\isamarkuptrue%
\isacommand{definition}\isamarkupfalse%
\ {\isacartoucheopen}{\isasymepsilon}{\isacharunderscore}{\kern0pt}pos\ w\ t\ a\ ts\ {\isacharequal}{\kern0pt}\ variable{\isacharunderscore}{\kern0pt}elim\ {\isacharparenleft}{\kern0pt}C\ w\ t\ a\ {\isacharat}{\kern0pt}\ neg{\isacharunderscore}{\kern0pt}b{\isacharprime}{\kern0pt}\ t\ a\ ts{\isacharparenright}{\kern0pt}{\isacartoucheclose}\isanewline
\isacommand{definition}\isamarkupfalse%
\ {\isacartoucheopen}{\isasymepsilon}{\isacharunderscore}{\kern0pt}neg\ w\ t\ a\ ts\ {\isacharequal}{\kern0pt}\ variable{\isacharunderscore}{\kern0pt}elim\ {\isacharparenleft}{\kern0pt}neg{\isacharunderscore}{\kern0pt}C\ w\ t\ a\ {\isacharat}{\kern0pt}\ b{\isacharprime}{\kern0pt}\ t\ a\ ts{\isacharparenright}{\kern0pt}{\isacartoucheclose}\isanewline
\isanewline
\isacommand{definition}\isamarkupfalse%
\ {\isacartoucheopen}{\isasymepsilon}{\isacharunderscore}{\kern0pt}max\ w\ t\ a\ ts\ {\isacharequal}{\kern0pt}\ max\ {\isacharparenleft}{\kern0pt}{\isasymepsilon}{\isacharunderscore}{\kern0pt}pos\ w\ t\ a\ ts{\isacharparenright}{\kern0pt}\ {\isacharparenleft}{\kern0pt}{\isasymepsilon}{\isacharunderscore}{\kern0pt}neg\ w\ t\ a\ ts{\isacharparenright}{\kern0pt}{\isacartoucheclose}\isanewline
\isanewline
\isacommand{end}\isamarkupfalse%
\end{isasnipenv}

\begin{isasnipenv}[snip:brancherrprop]{Branch Error Correctness}
\isacommand{lemma}\isamarkupfalse%
\ {\isasymepsilon}{\isacharunderscore}{\kern0pt}max{\isacharunderscore}{\kern0pt}correct{\isacharprime}{\kern0pt}{\isacharprime}{\kern0pt}{\isacharcolon}{\kern0pt}\isanewline
\ \ \isakeyword{assumes}\ {\isacartoucheopen}t\ {\isasymin}\ partial{\isacharunderscore}{\kern0pt}states{\isacartoucheclose}\ {\isacartoucheopen}a\ {\isasymin}\ actions{\isacartoucheclose}\ {\isacartoucheopen}set\ ts\ {\isasymsubseteq}\ partial{\isacharunderscore}{\kern0pt}states{\isacartoucheclose}\isanewline
\ \ \isakeyword{shows}\ {\isacartoucheopen}{\isasymepsilon}{\isacharunderscore}{\kern0pt}max\ w\ t\ a\ ts\ {\isacharequal}{\kern0pt}\isanewline
\ \ {\isacharparenleft}{\kern0pt}{\isasymSqunion}x\ {\isasymin}\ {\isacharbraceleft}{\kern0pt}x{\isachardot}{\kern0pt}\ x\ {\isasymin}\ states\ {\isasymand}\ consistent\ x\ t\ {\isasymand}\ {\isacharparenleft}{\kern0pt}{\isasymforall}t{\isacharprime}{\kern0pt}\ {\isasymin}\ set\ ts{\isachardot}{\kern0pt}\ {\isasymnot}\ consistent\ x\ t{\isacharprime}{\kern0pt}{\isacharparenright}{\kern0pt}{\isacharbraceright}{\kern0pt}{\isachardot}{\kern0pt}\ \isanewline
\ \ \ \ \ \ ereal\ {\isacharparenleft}{\kern0pt}dist\ {\isacharparenleft}{\kern0pt}Q\ w\ x\ a{\isacharparenright}{\kern0pt}\ {\isacharparenleft}{\kern0pt}{\isasymnu}\isactrlsub w\ w\ x{\isacharparenright}{\kern0pt}{\isacharparenright}{\kern0pt}{\isacharparenright}{\kern0pt}{\isacartoucheclose}
\end{isasnipenv}

\begin{isasnipenv}[snip:varelim]{Variable Elimination Definitions}
\isacommand{locale}\isamarkupfalse%
\ Variable{\isacharunderscore}{\kern0pt}Elimination\ {\isacharequal}{\kern0pt}\ Variable{\isacharunderscore}{\kern0pt}Elimination{\isacharunderscore}{\kern0pt}Consts\ {\isasymF}\ dims\ {\isasymO}\ doms\isanewline
\ \ \isakeyword{for}\isanewline
\ \ \ \ {\isasymF}\ {\isacharcolon}{\kern0pt}{\isacharcolon}{\kern0pt}\ {\isacartoucheopen}{\isacharparenleft}{\kern0pt}{\isacharparenleft}{\kern0pt}{\isacharparenleft}{\kern0pt}nat\ {\isasymrightharpoonup}\ {\isacharprime}{\kern0pt}a{\isacharparenright}{\kern0pt}\ {\isasymRightarrow}\ ereal{\isacharparenright}{\kern0pt}\ {\isasymtimes}\ nat\ set{\isacharparenright}{\kern0pt}\ list{\isacartoucheclose}\ \isakeyword{and}\ %
\isamarkupcmt{Functions to be maximized%
}\isanewline
\ \ \ \ dims\ {\isacharcolon}{\kern0pt}{\isacharcolon}{\kern0pt}\ {\isacartoucheopen}nat{\isacartoucheclose}\ \isakeyword{and}\ %
\isamarkupcmt{Number of variables%
}\isanewline
\ \ \ \ {\isasymO}\ {\isacharcolon}{\kern0pt}{\isacharcolon}{\kern0pt}\ {\isacartoucheopen}nat\ {\isasymRightarrow}\ nat{\isacartoucheclose}\ \isakeyword{and}\ %
\isamarkupcmt{Elimination order%
}\isanewline
\ \ \ \ doms\ {\isacharcolon}{\kern0pt}{\isacharcolon}{\kern0pt}\ {\isacartoucheopen}nat\ {\isasymRightarrow}\ {\isacharprime}{\kern0pt}a\ set{\isacartoucheclose}\ %
\isamarkupcmt{Domains of the dimensions%
}\ {\isacharplus}{\kern0pt}\isanewline
\ \ \isakeyword{assumes}\ \isanewline
\ \ \ \ {\isasymO}{\isacharunderscore}{\kern0pt}bij{\isacharcolon}{\kern0pt}\ {\isacartoucheopen}bij{\isacharunderscore}{\kern0pt}betw\ {\isasymO}\ {\isacharbraceleft}{\kern0pt}{\isachardot}{\kern0pt}{\isachardot}{\kern0pt}{\isacharless}{\kern0pt}dims{\isacharbraceright}{\kern0pt}\ {\isacharbraceleft}{\kern0pt}{\isachardot}{\kern0pt}{\isachardot}{\kern0pt}{\isacharless}{\kern0pt}dims{\isacharbraceright}{\kern0pt}{\isacartoucheclose}\ \isakeyword{and}\isanewline
\ \ \ \ {\isasymF}{\isacharunderscore}{\kern0pt}scopes{\isacharcolon}{\kern0pt}\ {\isacartoucheopen}{\isasymforall}{\isacharparenleft}{\kern0pt}f{\isacharcomma}{\kern0pt}\ s{\isacharparenright}{\kern0pt}\ {\isasymin}\ set\ {\isasymF}{\isachardot}{\kern0pt}\ s\ {\isasymsubseteq}\ {\isacharbraceleft}{\kern0pt}{\isachardot}{\kern0pt}{\isachardot}{\kern0pt}{\isacharless}{\kern0pt}dims{\isacharbraceright}{\kern0pt}\ {\isasymand}\isanewline
\ \ \ \ has{\isacharunderscore}{\kern0pt}scope{\isacharunderscore}{\kern0pt}on\ f\ {\isacharbraceleft}{\kern0pt}x{\isachardot}{\kern0pt}\ {\isasymforall}{\isacharparenleft}{\kern0pt}i{\isacharcomma}{\kern0pt}\ j{\isacharparenright}{\kern0pt}\ {\isasymin}\ Map{\isachardot}{\kern0pt}graph\ x{\isachardot}{\kern0pt}\ i\ {\isasymin}\ {\isacharbraceleft}{\kern0pt}{\isachardot}{\kern0pt}{\isachardot}{\kern0pt}{\isacharless}{\kern0pt}dims{\isacharbraceright}{\kern0pt}\ {\isasymand}\ j\ {\isasymin}\ doms\ i{\isacharbraceright}{\kern0pt}\ s{\isacartoucheclose}\ \isakeyword{and}\isanewline
\ \ \ \ domains{\isacharunderscore}{\kern0pt}ne{\isacharcolon}{\kern0pt}\ {\isacartoucheopen}{\isasymAnd}i{\isachardot}{\kern0pt}\ i\ {\isacharless}{\kern0pt}\ dims\ {\isasymLongrightarrow}\ doms\ i\ {\isasymnoteq}\ {\isacharbraceleft}{\kern0pt}{\isacharbraceright}{\kern0pt}{\isacartoucheclose}\ \isakeyword{and}\isanewline
\ \ \ \ domains{\isacharunderscore}{\kern0pt}fin{\isacharcolon}{\kern0pt}\ {\isacartoucheopen}{\isasymAnd}i{\isachardot}{\kern0pt}\ i\ {\isacharless}{\kern0pt}\ dims\ {\isasymLongrightarrow}\ finite\ {\isacharparenleft}{\kern0pt}doms\ i{\isacharparenright}{\kern0pt}{\isacartoucheclose}\ \isakeyword{and}\isanewline
\ \ \ \ {\isasymF}{\isacharunderscore}{\kern0pt}not{\isacharunderscore}{\kern0pt}inf{\isacharcolon}{\kern0pt}\ {\isacartoucheopen}{\isasymAnd}f\ x{\isachardot}{\kern0pt}\ f\ {\isasymin}\ set\ {\isasymF}\ {\isasymLongrightarrow}\ {\isacharparenleft}{\kern0pt}fst\ f{\isacharparenright}{\kern0pt}\ x\ {\isasymnoteq}\ {\isasyminfinity}{\isacartoucheclose}\isanewline
\isakeyword{begin}%
\begin{isamarkuptext}%
The body of the algorithm, in a single iteration, we first select the variable to eliminate as
  \isa{{\isasymO}\ i}, then partition the work set of functions \isa{F} based on whether \isa{{\isasymO}\ i}
  is in their scope.
  Next we build a new function that maximizes all functions that contain \isa{{\isacharat}{\kern0pt}{\isacharbraceleft}{\kern0pt}term\ {\isasymopen}{\isasymO}\ i{\isasymclose}{\isacharbraceright}{\kern0pt}} over this
  variable, and add it to the working set.%
\end{isamarkuptext}\isamarkuptrue%
\isacommand{definition}\isamarkupfalse%
\ {\isacartoucheopen}elim{\isacharunderscore}{\kern0pt}step\ iF\ {\isacharequal}{\kern0pt}\ {\isacharparenleft}{\kern0pt}\isanewline
\ \ let\isanewline
\ \ \ \ {\isacharparenleft}{\kern0pt}i{\isacharcomma}{\kern0pt}\ F{\isacharparenright}{\kern0pt}\ {\isacharequal}{\kern0pt}\ iF{\isacharsemicolon}{\kern0pt}\isanewline
\ \ \ \ l\ {\isacharequal}{\kern0pt}\ {\isasymO}\ i{\isacharsemicolon}{\kern0pt}\isanewline
\ \ \ \ E\ {\isacharequal}{\kern0pt}\ filter\ {\isacharparenleft}{\kern0pt}{\isasymlambda}f{\isachardot}{\kern0pt}\ l\ {\isasymin}\ scope\ f{\isacharparenright}{\kern0pt}\ F{\isacharsemicolon}{\kern0pt}\isanewline
\ \ \ \ e\ {\isacharequal}{\kern0pt}\ {\isacharparenleft}{\kern0pt}{\isasymlambda}x{\isachardot}{\kern0pt}\ MAX\ x\isactrlsub l\ {\isasymin}\ doms\ l{\isachardot}{\kern0pt}\ {\isasymSum}f\ {\isasymleftarrow}\ E{\isachardot}{\kern0pt}\ {\isacharparenleft}{\kern0pt}fn\ f\ {\isacharparenleft}{\kern0pt}x{\isacharparenleft}{\kern0pt}l\ {\isasymmapsto}\ x\isactrlsub l{\isacharparenright}{\kern0pt}{\isacharparenright}{\kern0pt}{\isacharparenright}{\kern0pt}{\isacharparenright}{\kern0pt}{\isacharsemicolon}{\kern0pt}\isanewline
\ \ \ \ scope{\isacharunderscore}{\kern0pt}e\ {\isacharequal}{\kern0pt}\ {\isacharparenleft}{\kern0pt}{\isasymUnion}f{\isasymin}set\ E{\isachardot}{\kern0pt}\ scope\ f{\isacharparenright}{\kern0pt}\ {\isacharminus}{\kern0pt}\ {\isacharbraceleft}{\kern0pt}l{\isacharbraceright}{\kern0pt}\isanewline
\ \ in\isanewline
\ \ \ \ {\isacharparenleft}{\kern0pt}i\ {\isacharplus}{\kern0pt}\ {\isadigit{1}}{\isacharcomma}{\kern0pt}\ {\isacharparenleft}{\kern0pt}e{\isacharcomma}{\kern0pt}\ scope{\isacharunderscore}{\kern0pt}e{\isacharparenright}{\kern0pt}\ {\isacharhash}{\kern0pt}\ filter\ {\isacharparenleft}{\kern0pt}{\isasymlambda}f{\isachardot}{\kern0pt}\ l\ {\isasymnotin}\ scope\ f{\isacharparenright}{\kern0pt}\ F{\isacharparenright}{\kern0pt}\isanewline
\ \ {\isacharparenright}{\kern0pt}{\isacartoucheclose}%
\begin{isamarkuptext}%
The body of the algorithms needs to be iterated \isa{dims} times.%
\end{isamarkuptext}\isamarkuptrue%
\isacommand{definition}\isamarkupfalse%
\ {\isacartoucheopen}elim{\isacharunderscore}{\kern0pt}aux\ {\isacharequal}{\kern0pt}\ {\isacharparenleft}{\kern0pt}elim{\isacharunderscore}{\kern0pt}step\ {\isacharcircum}{\kern0pt}{\isacharcircum}{\kern0pt}\ dims{\isacharparenright}{\kern0pt}\ {\isacharparenleft}{\kern0pt}{\isadigit{0}}{\isacharcomma}{\kern0pt}\ {\isasymF}{\isacharparenright}{\kern0pt}{\isacartoucheclose}%
\begin{isamarkuptext}%
Finally, we are left with only functions with empty scopes, and we return their sum.%
\end{isamarkuptext}\isamarkuptrue%
\isacommand{definition}\isamarkupfalse%
\ {\isacartoucheopen}elim{\isacharunderscore}{\kern0pt}max\ {\isacharequal}{\kern0pt}\ {\isacharparenleft}{\kern0pt}{\isasymSum}{\isacharparenleft}{\kern0pt}f{\isacharcomma}{\kern0pt}\ f{\isacharunderscore}{\kern0pt}scope{\isacharparenright}{\kern0pt}\ {\isasymleftarrow}\ snd\ elim{\isacharunderscore}{\kern0pt}aux{\isachardot}{\kern0pt}\ f\ Map{\isachardot}{\kern0pt}empty{\isacharparenright}{\kern0pt}{\isacartoucheclose}%
\begin{isamarkuptext}%
\isa{expl{\isacharunderscore}{\kern0pt}max} is the inefficient variant of the algorithm that enumerates all states.%
\end{isamarkuptext}\isamarkuptrue%
\isacommand{definition}\isamarkupfalse%
\ {\isacartoucheopen}expl{\isacharunderscore}{\kern0pt}max{\isacharunderscore}{\kern0pt}{\isasymF}\ {\isasymF}{\isacharprime}{\kern0pt}\ {\isacharequal}{\kern0pt}\ {\isacharparenleft}{\kern0pt}MAX\ x\ {\isasymin}\ full{\isacharunderscore}{\kern0pt}vecs{\isachardot}{\kern0pt}\ {\isasymSum}f\ {\isasymleftarrow}\ {\isasymF}{\isacharprime}{\kern0pt}{\isachardot}{\kern0pt}\ fst\ f\ x{\isacharparenright}{\kern0pt}{\isacartoucheclose}\isanewline
\isacommand{definition}\isamarkupfalse%
\ {\isacartoucheopen}expl{\isacharunderscore}{\kern0pt}max\ {\isacharequal}{\kern0pt}\ expl{\isacharunderscore}{\kern0pt}max{\isacharunderscore}{\kern0pt}{\isasymF}\ {\isasymF}{\isacartoucheclose}%
\begin{isamarkuptext}%
The invariant \isa{invar{\isacharunderscore}{\kern0pt}max} can be used to show the algorithm \isa{elim{\isacharunderscore}{\kern0pt}max} correct.%
\end{isamarkuptext}\isamarkuptrue%
\isacommand{definition}\isamarkupfalse%
\ {\isacartoucheopen}invar{\isacharunderscore}{\kern0pt}max\ {\isasymF}{\isacharprime}{\kern0pt}\ i\ {\isasymlongleftrightarrow}\isanewline
\ \ {\isacharparenleft}{\kern0pt}{\isasymforall}f\ {\isasymin}\ set\ {\isasymF}{\isacharprime}{\kern0pt}{\isachardot}{\kern0pt}\ {\isasymforall}x{\isachardot}{\kern0pt}\ fst\ f\ x\ {\isasymnoteq}\ {\isasyminfinity}{\isacharparenright}{\kern0pt}\ {\isasymand}\isanewline
\ \ {\isacharparenleft}{\kern0pt}{\isasymforall}f\ {\isasymin}\ set\ {\isasymF}{\isacharprime}{\kern0pt}{\isachardot}{\kern0pt}\ has{\isacharunderscore}{\kern0pt}scope{\isacharunderscore}{\kern0pt}on\ {\isacharparenleft}{\kern0pt}fst\ f{\isacharparenright}{\kern0pt}\ partial{\isacharunderscore}{\kern0pt}vecs\ {\isacharparenleft}{\kern0pt}snd\ f{\isacharparenright}{\kern0pt}{\isacharparenright}{\kern0pt}\ {\isasymand}\isanewline
\ \ {\isacharparenleft}{\kern0pt}{\isasymforall}f\ {\isasymin}\ set\ {\isasymF}{\isacharprime}{\kern0pt}{\isachardot}{\kern0pt}\ snd\ f\ {\isasymsubseteq}\ {\isacharbraceleft}{\kern0pt}{\isachardot}{\kern0pt}{\isachardot}{\kern0pt}{\isacharless}{\kern0pt}dims{\isacharbraceright}{\kern0pt}\ {\isacharminus}{\kern0pt}\ {\isasymO}\ {\isacharbackquote}{\kern0pt}\ {\isacharbraceleft}{\kern0pt}{\isachardot}{\kern0pt}{\isachardot}{\kern0pt}{\isacharless}{\kern0pt}i{\isacharbraceright}{\kern0pt}{\isacharparenright}{\kern0pt}\ {\isasymand}\isanewline
\ \ {\isacharparenleft}{\kern0pt}MAX\ x{\isasymin}vecs{\isacharunderscore}{\kern0pt}on\ {\isacharparenleft}{\kern0pt}{\isacharbraceleft}{\kern0pt}{\isachardot}{\kern0pt}{\isachardot}{\kern0pt}{\isacharless}{\kern0pt}dims{\isacharbraceright}{\kern0pt}\ {\isacharminus}{\kern0pt}\ {\isasymO}\ {\isacharbackquote}{\kern0pt}\ {\isacharbraceleft}{\kern0pt}{\isachardot}{\kern0pt}{\isachardot}{\kern0pt}{\isacharless}{\kern0pt}i{\isacharbraceright}{\kern0pt}{\isacharparenright}{\kern0pt}{\isachardot}{\kern0pt}\ {\isasymSum}f{\isasymleftarrow}{\isasymF}{\isacharprime}{\kern0pt}{\isachardot}{\kern0pt}\ fst\ f\ x{\isacharparenright}{\kern0pt}\ {\isacharequal}{\kern0pt}\ expl{\isacharunderscore}{\kern0pt}max{\isacharunderscore}{\kern0pt}{\isasymF}\ {\isasymF}{\isacartoucheclose}\isanewline
\isanewline
\isacommand{lemma}\isamarkupfalse%
\ elim{\isacharunderscore}{\kern0pt}max{\isacharunderscore}{\kern0pt}correct{\isacharcolon}{\kern0pt}\ {\isacartoucheopen}elim{\isacharunderscore}{\kern0pt}max\ {\isacharequal}{\kern0pt}\ expl{\isacharunderscore}{\kern0pt}max{\isacartoucheclose}\isanewline
\isanewline
\isacommand{end}\isamarkupfalse%
\end{isasnipenv}

\subsection{Policy Evaluation}
\begin{isasnipenv}[snip:valdet]{Policy Evaluation / Weight Update}
\isacommand{locale}\isamarkupfalse%
\ ValueDet{\isacharunderscore}{\kern0pt}API{\isacharunderscore}{\kern0pt}Consts{\kern0pt}\ {\isacharequal}{\kern0pt} \ldots \isanewline
\isakeyword{fixes}\isanewline
\ \ min{\isacharunderscore}{\kern0pt}sol\ {\isacharcolon}{\kern0pt}{\isacharcolon}{\kern0pt}\ {\isacartoucheopen}real\ {\isasymtimes}\ {\isacharparenleft}{\kern0pt}nat\ {\isasymRightarrow}\ real{\isacharparenright}{\kern0pt}{\isacartoucheclose}\isanewline
\ \ dec{\isacharunderscore}{\kern0pt}pol\ {\isacharcolon}{\kern0pt}{\isacharcolon}{\kern0pt}\ {\isacartoucheopen}{\isacharparenleft}{\kern0pt}{\isacharprime}{\kern0pt}x\ state\ {\isasymtimes}\ {\isacharprime}{\kern0pt}a{\isacharparenright}{\kern0pt}\ list{\isacartoucheclose}\ \isanewline
\isakeyword{begin}\isanewline
\isacommand{definition}\isamarkupfalse%
\ {\isacartoucheopen}dec{\isacharunderscore}{\kern0pt}pol{\isacharunderscore}{\kern0pt}spec\ {\isasymlongleftrightarrow}\ fst\ {\isacharbackquote}{\kern0pt}\ set\ dec{\isacharunderscore}{\kern0pt}pol\ {\isasymsubseteq}\ partial{\isacharunderscore}{\kern0pt}states\ {\isasymand}\ snd\ {\isacharbackquote}{\kern0pt}\ set\ dec{\isacharunderscore}{\kern0pt}pol\ {\isasymsubseteq}\ actions\ {\isasymand}\ distinct\ dec{\isacharunderscore}{\kern0pt}pol\ {\isasymand}\isanewline
\ \ {\isacharparenleft}{\kern0pt}{\isasymforall}x\ {\isasymin}\ states{\isachardot}{\kern0pt}\ {\isasymexists}y\ {\isasymin}\ fst\ {\isacharbackquote}{\kern0pt}\ set\ dec{\isacharunderscore}{\kern0pt}pol{\isachardot}{\kern0pt}\ consistent\ x\ y{\isacharparenright}{\kern0pt}{\isacartoucheclose}\isanewline
\isanewline
\isacommand{definition}\isamarkupfalse%
\ {\isacartoucheopen}update{\isacharunderscore}{\kern0pt}weights{\isacharunderscore}{\kern0pt}iter\ {\isacharequal}{\kern0pt}\isanewline
\ \  {\isacharparenleft}{\kern0pt}{\isasymlambda}{\isacharparenleft}{\kern0pt}t{\isacharcomma}{\kern0pt}\ a{\isacharparenright}{\kern0pt}\ {\isacharparenleft}{\kern0pt}ts{\isacharcomma}{\kern0pt}\ cs{\isacharparenright}{\kern0pt}{\isachardot}{\kern0pt}\ {\isacharparenleft}{\kern0pt}t{\isacharhash}{\kern0pt}ts{\isacharcomma}{\kern0pt}\ union{\isacharunderscore}{\kern0pt}constr\ {\isacharparenleft}{\kern0pt}gen{\isacharunderscore}{\kern0pt}constr\ t\ a\ ts{\isacharparenright}{\kern0pt}\ cs{\isacharparenright}{\kern0pt}{\isacharparenright}{\kern0pt}{\isacartoucheclose}\isanewline
\isanewline
\isacommand{definition}\isamarkupfalse%
\ {\isacartoucheopen}constrs{\isacharunderscore}{\kern0pt}list\ xs\ {\isacharequal}{\kern0pt}\ snd\ {\isacharparenleft}{\kern0pt}fold\ update{\isacharunderscore}{\kern0pt}weights{\isacharunderscore}{\kern0pt}iter\ xs\ {\isacharparenleft}{\kern0pt}{\isacharbrackleft}{\kern0pt}{\isacharbrackright}{\kern0pt}{\isacharcomma}{\kern0pt}\ empty{\isacharunderscore}{\kern0pt}constr{\isacharparenright}{\kern0pt}{\isacharparenright}{\kern0pt}{\isacartoucheclose}\isanewline
\isanewline
\isacommand{definition}\isamarkupfalse%
\ {\isacartoucheopen}constrs\ {\isacharequal}{\kern0pt}\ constrs{\isacharunderscore}{\kern0pt}list\ dec{\isacharunderscore}{\kern0pt}pol{\isacartoucheclose}\isanewline
\isanewline
\isacommand{definition}\isamarkupfalse%
\ {\isacartoucheopen}update{\isacharunderscore}{\kern0pt}weights\ {\isacharequal}{\kern0pt}\ snd\ min{\isacharunderscore}{\kern0pt}sol{\isacartoucheclose}\isanewline
\isacommand{end}\isamarkupfalse%
\end{isasnipenv}

\begin{isasnipenv}[snip:valdetprop]{Policy Evaluation Correctness}
\isacommand{lemma}\isamarkupfalse%
\ proj{\isacharunderscore}{\kern0pt}err{\isacharunderscore}{\kern0pt}upd{\isacharunderscore}{\kern0pt}eq{\isacharunderscore}{\kern0pt}pol{\isacharcolon}{\kern0pt}\ {\isacartoucheopen}proj{\isacharunderscore}{\kern0pt}err{\isacharunderscore}{\kern0pt}w\ {\isacharparenleft}{\kern0pt}dec{\isacharunderscore}{\kern0pt}list{\isacharunderscore}{\kern0pt}to{\isacharunderscore}{\kern0pt}pol\ dec{\isacharunderscore}{\kern0pt}pol{\isacharparenright}{\kern0pt}\ update{\isacharunderscore}{\kern0pt}weights\ {\isacharequal}{\kern0pt}\ \isanewline
\ \ proj{\isacharunderscore}{\kern0pt}err{\isacharunderscore}{\kern0pt}pol\ {\isacharparenleft}{\kern0pt}dec{\isacharunderscore}{\kern0pt}list{\isacharunderscore}{\kern0pt}to{\isacharunderscore}{\kern0pt}pol\ dec{\isacharunderscore}{\kern0pt}pol{\isacharparenright}{\kern0pt}{\isacartoucheclose}
\end{isasnipenv}

\begin{isasnipenv}[snip:valdetbranch]{Linear Program for a Branch}
\isacommand{locale}\isamarkupfalse%
\ Process{\isacharunderscore}{\kern0pt}Branch{\isacharunderscore}{\kern0pt}Consts\ {\isacharequal}{\kern0pt} \ldots \isanewline
\isakeyword{fixes}\isanewline
\ \ t\ {\isacharcolon}{\kern0pt}{\isacharcolon}{\kern0pt}\ {\isacartoucheopen}{\isacharprime}{\kern0pt}x\ state{\isacartoucheclose}\ \isakeyword{and}\isanewline
\ \ a\ {\isacharcolon}{\kern0pt}{\isacharcolon}{\kern0pt}\ {\isacartoucheopen}{\isacharprime}{\kern0pt}a{\isacartoucheclose}\ \isakeyword{and}\isanewline
\ \ ts\ {\isacharcolon}{\kern0pt}{\isacharcolon}{\kern0pt}\ {\isacartoucheopen}{\isacharprime}{\kern0pt}x\ state\ list{\isacartoucheclose}\ \isakeyword{and}\isanewline
\ \ factored{\isacharunderscore}{\kern0pt}lp\ {\isacharcolon}{\kern0pt}{\isacharcolon}{\kern0pt}\ {\isacartoucheopen}\isanewline
\ \ \ \ \ \ bool\ {\isasymRightarrow}\isanewline
\ \ \ \ \ \ {\isacharparenleft}{\kern0pt}{\isacharparenleft}{\kern0pt}{\isacharprime}{\kern0pt}x\ state\ {\isasymRightarrow}\ real{\isacharparenright}{\kern0pt}\ {\isasymtimes}\ {\isacharparenleft}{\kern0pt}nat\ set{\isacharparenright}{\kern0pt}{\isacharparenright}{\kern0pt}\ list\ {\isasymRightarrow}\ %
\isamarkupcmt{C%
}\isanewline
\ \ \ \ \ \ {\isacharparenleft}{\kern0pt}{\isacharparenleft}{\kern0pt}{\isacharprime}{\kern0pt}x\ state\ {\isasymRightarrow}\ ereal{\isacharparenright}{\kern0pt}\ {\isasymtimes}\ {\isacharparenleft}{\kern0pt}nat\ set{\isacharparenright}{\kern0pt}{\isacharparenright}{\kern0pt}\ list\ {\isasymRightarrow}\ %
\isamarkupcmt{b%
}\isanewline
\ \ \ \ \ \ {\isacharprime}{\kern0pt}c{\isacartoucheclose}\isanewline
\isakeyword{begin}%
\begin{isamarkuptext}%
Here we define the specification of the Factored LP algorithm:
\end{isamarkuptext}\isamarkuptrue%
\isacommand{definition}\isamarkupfalse%
\ {\isacartoucheopen}factored{\isacharunderscore}{\kern0pt}lp{\isacharunderscore}{\kern0pt}constrs\ pos\ C\ b\ {\isasymlongleftrightarrow}\isanewline
\ constr{\isacharunderscore}{\kern0pt}set\ {\isacharparenleft}{\kern0pt}factored{\isacharunderscore}{\kern0pt}lp\ pos\ C\ b{\isacharparenright}{\kern0pt}\ {\isacharequal}{\kern0pt}\ {\isacharbraceleft}{\kern0pt}{\isacharparenleft}{\kern0pt}{\isasymphi}{\isacharcomma}{\kern0pt}\ w{\isacharparenright}{\kern0pt}{\isachardot}{\kern0pt}\ {\isasymforall}x\ {\isasymin}\ states{\isachardot}{\kern0pt}\ \isanewline
\ \ \ \ ereal\ {\isasymphi}\ {\isasymge}\ ereal\ {\isacharparenleft}{\kern0pt}{\isasymSum}i\ {\isacharless}{\kern0pt}\ length\ C{\isachardot}{\kern0pt}\ w\ i\ {\isacharasterisk}{\kern0pt}\ {\isacharparenleft}{\kern0pt}fst\ {\isacharparenleft}{\kern0pt}C\ {\isacharbang}{\kern0pt}\ i{\isacharparenright}{\kern0pt}\ x{\isacharparenright}{\kern0pt}{\isacharparenright}{\kern0pt}\isanewline
\ \ \ \ \ \  {\isacharplus}{\kern0pt}\ {\isacharparenleft}{\kern0pt}{\isasymSum}i\ {\isacharless}{\kern0pt}\ length\ b{\isachardot}{\kern0pt}\ fst\ {\isacharparenleft}{\kern0pt}b\ {\isacharbang}{\kern0pt}\ i{\isacharparenright}{\kern0pt}\ x{\isacharparenright}{\kern0pt}{\isacharbraceright}{\kern0pt}{\isacartoucheclose}\isanewline
\isanewline
\isacommand{definition}\isamarkupfalse%
\ {\isacartoucheopen}factored{\isacharunderscore}{\kern0pt}lp{\isacharunderscore}{\kern0pt}inv\ pos\ C\ b\ {\isasymlongleftrightarrow}\isanewline
\ inv{\isacharunderscore}{\kern0pt}constr\ {\isacharparenleft}{\kern0pt}factored{\isacharunderscore}{\kern0pt}lp\ pos\ C\ b{\isacharparenright}{\kern0pt}{\isacartoucheclose}\isanewline
\isanewline
\isacommand{definition}\isamarkupfalse%
\ {\isacartoucheopen}factored{\isacharunderscore}{\kern0pt}lp{\isacharunderscore}{\kern0pt}privs\ pos\ C\ b\ {\isasymlongleftrightarrow}\isanewline
\ privates\ {\isacharparenleft}{\kern0pt}factored{\isacharunderscore}{\kern0pt}lp\ pos\ C\ b{\isacharparenright}{\kern0pt}\ {\isasymsubseteq}\ {\isacharbraceleft}{\kern0pt}{\isacharparenleft}{\kern0pt}{\isacharparenleft}{\kern0pt}t{\isacharcomma}{\kern0pt}\ a{\isacharparenright}{\kern0pt}{\isacharcomma}{\kern0pt}\ pos{\isacharparenright}{\kern0pt}{\isacharbraceright}{\kern0pt}{\isacartoucheclose}\isanewline
\isanewline
\isacommand{definition}\isamarkupfalse%
\ {\isacartoucheopen}inv{\isacharunderscore}{\kern0pt}b\ b\ {\isasymlongleftrightarrow}\isanewline
\ \ {\isacharparenleft}{\kern0pt}{\isasymforall}{\isacharparenleft}{\kern0pt}b{\isacharcomma}{\kern0pt}\ scope{\isacharunderscore}{\kern0pt}b{\isacharparenright}{\kern0pt}\ {\isasymin}\ set\ b{\isachardot}{\kern0pt}\ has{\isacharunderscore}{\kern0pt}scope{\isacharunderscore}{\kern0pt}on\ b\ partial{\isacharunderscore}{\kern0pt}states\ scope{\isacharunderscore}{\kern0pt}b\ {\isasymand}\ scope{\isacharunderscore}{\kern0pt}b\ {\isasymsubseteq}\ vars\ {\isasymand}\ {\isacharparenleft}{\kern0pt}{\isasymforall}x{\isachardot}{\kern0pt}\ b\ x\ {\isasymnoteq}\ {\isasyminfinity}{\isacharparenright}{\kern0pt}{\isacharparenright}{\kern0pt}{\isacartoucheclose}\isanewline
\isanewline
\isacommand{definition}\isamarkupfalse%
\ {\isacartoucheopen}inv{\isacharunderscore}{\kern0pt}C\ b\ {\isasymlongleftrightarrow}\isanewline
\ \ {\isacharparenleft}{\kern0pt}{\isasymforall}{\isacharparenleft}{\kern0pt}b{\isacharcomma}{\kern0pt}\ scope{\isacharunderscore}{\kern0pt}b{\isacharparenright}{\kern0pt}\ {\isasymin}\ set\ b{\isachardot}{\kern0pt}\ has{\isacharunderscore}{\kern0pt}scope{\isacharunderscore}{\kern0pt}on\ b\ partial{\isacharunderscore}{\kern0pt}states\ scope{\isacharunderscore}{\kern0pt}b\ {\isasymand}\ scope{\isacharunderscore}{\kern0pt}b\ {\isasymsubseteq}\ vars{\isacharparenright}{\kern0pt}{\isacartoucheclose}\isanewline
\isanewline
\isacommand{definition}\isamarkupfalse%
\ {\isacartoucheopen}factored{\isacharunderscore}{\kern0pt}lp{\isacharunderscore}{\kern0pt}spec\ {\isasymlongleftrightarrow}\isanewline
\ \ {\isacharparenleft}{\kern0pt}{\isasymforall}pos\ C\ b{\isachardot}{\kern0pt}\ inv{\isacharunderscore}{\kern0pt}C\ C\ {\isasymlongrightarrow}\ inv{\isacharunderscore}{\kern0pt}b\ b\ {\isasymlongrightarrow}\isanewline
\ \ factored{\isacharunderscore}{\kern0pt}lp{\isacharunderscore}{\kern0pt}constrs\ pos\ C\ b\ {\isasymand}\ factored{\isacharunderscore}{\kern0pt}lp{\isacharunderscore}{\kern0pt}inv\ pos\ C\ b\ {\isasymand}\isanewline 
\ \ \ \ factored{\isacharunderscore}{\kern0pt}lp{\isacharunderscore}{\kern0pt}privs\ pos\ C\ b{\isacharparenright}{\kern0pt}{\isacartoucheclose}\isanewline
\isanewline
\isacommand{definition}\isamarkupfalse%
\ {\isacartoucheopen}neg{\isacharunderscore}{\kern0pt}scoped\ {\isacharequal}{\kern0pt}\ {\isacharparenleft}{\kern0pt}{\isasymlambda}{\isacharparenleft}{\kern0pt}f{\isacharcomma}{\kern0pt}\ s{\isacharparenright}{\kern0pt}{\isachardot}{\kern0pt}\ {\isacharparenleft}{\kern0pt}{\isacharminus}{\kern0pt}f{\isacharcomma}{\kern0pt}\ s{\isacharparenright}{\kern0pt}{\isacharparenright}{\kern0pt}{\isacartoucheclose}\isanewline
\isanewline
\isacommand{definition}\isamarkupfalse%
\ {\isacartoucheopen}hg{\isacharunderscore}{\kern0pt}scope\ i\ {\isacharequal}{\kern0pt}\ h{\isacharunderscore}{\kern0pt}scope\ i\ {\isasymunion}\ {\isasymGamma}\isactrlsub a\ a\ {\isacharparenleft}{\kern0pt}h{\isacharunderscore}{\kern0pt}scope\ i{\isacharparenright}{\kern0pt}\ {\isacharminus}{\kern0pt}\ dom\ t{\isacartoucheclose}\isanewline
\isacommand{definition}\isamarkupfalse%
\ {\isacartoucheopen}hg{\isacharunderscore}{\kern0pt}inst\ i\ {\isacharequal}{\kern0pt}\ instantiate\ {\isacharparenleft}{\kern0pt}{\isasymlambda}x{\isachardot}{\kern0pt}\ h\ i\ x\ {\isacharminus}{\kern0pt}\ l\ {\isacharasterisk}{\kern0pt}\ g{\isacharprime}{\kern0pt}\ i\ a\ x{\isacharparenright}{\kern0pt}\ t{\isacartoucheclose}\isanewline
\isanewline
\isacommand{definition}\isamarkupfalse%
\ {\isacartoucheopen}C\ {\isacharequal}{\kern0pt}\ {\isacharparenleft}{\kern0pt}map\ {\isacharparenleft}{\kern0pt}{\isasymlambda}i{\isachardot}{\kern0pt}\ {\isacharparenleft}{\kern0pt}hg{\isacharunderscore}{\kern0pt}inst\ i{\isacharcomma}{\kern0pt}\ hg{\isacharunderscore}{\kern0pt}scope\ i{\isacharparenright}{\kern0pt}{\isacharparenright}{\kern0pt}\ {\isacharbrackleft}{\kern0pt}{\isadigit{0}}{\isachardot}{\kern0pt}{\isachardot}{\kern0pt}{\isacharless}{\kern0pt}h{\isacharunderscore}{\kern0pt}dim{\isacharbrackright}{\kern0pt}{\isacharparenright}{\kern0pt}{\isacartoucheclose}\isanewline
\isacommand{definition}\isamarkupfalse%
\ {\isacartoucheopen}neg{\isacharunderscore}{\kern0pt}C\ {\isacharequal}{\kern0pt}\ map\ neg{\isacharunderscore}{\kern0pt}scoped\ C{\isacartoucheclose}\isanewline
\isanewline
\isamarkupcmt{the function \isa{r\ {\isacharminus}{\kern0pt}\ {\isasymI}} becomes \isa{b}%
}\isanewline
\isacommand{definition}\isamarkupfalse%
\ {\isacartoucheopen}r{\isacharunderscore}{\kern0pt}act{\isacharunderscore}{\kern0pt}dim\ {\isacharequal}{\kern0pt}\ reward{\isacharunderscore}{\kern0pt}dim\ a\ {\isacharminus}{\kern0pt}\ reward{\isacharunderscore}{\kern0pt}dim\ d{\isacartoucheclose}\isanewline
\isacommand{definition}\isamarkupfalse%
\ {\isacartoucheopen}r{\isacharunderscore}{\kern0pt}scope\ i\ {\isacharequal}{\kern0pt}\ reward{\isacharunderscore}{\kern0pt}scope\ a\ i\ {\isacharminus}{\kern0pt}\ dom\ t{\isacartoucheclose}\isanewline
\isacommand{definition}\isamarkupfalse%
\ {\isacartoucheopen}r{\isacharunderscore}{\kern0pt}inst\ i\ {\isacharequal}{\kern0pt}\ instantiate\ {\isacharparenleft}{\kern0pt}rewards\ a\ i{\isacharparenright}{\kern0pt}\ t{\isacartoucheclose}\isanewline
\isanewline
\isacommand{definition}\isamarkupfalse%
\ {\isacartoucheopen}b\ {\isacharequal}{\kern0pt}\ {\isacharparenleft}{\kern0pt}map\ {\isacharparenleft}{\kern0pt}{\isasymlambda}i{\isachardot}{\kern0pt}\ {\isacharparenleft}{\kern0pt}r{\isacharunderscore}{\kern0pt}inst\ i{\isacharcomma}{\kern0pt}\ r{\isacharunderscore}{\kern0pt}scope\ i{\isacharparenright}{\kern0pt}{\isacharparenright}{\kern0pt}\ {\isacharbrackleft}{\kern0pt}{\isadigit{0}}{\isachardot}{\kern0pt}{\isachardot}{\kern0pt}{\isacharless}{\kern0pt}reward{\isacharunderscore}{\kern0pt}dim\ a{\isacharbrackright}{\kern0pt}{\isacharparenright}{\kern0pt}{\isacartoucheclose}\isanewline
\isacommand{definition}\isamarkupfalse%
\ {\isacartoucheopen}neg{\isacharunderscore}{\kern0pt}b\ {\isacharequal}{\kern0pt}\ map\ neg{\isacharunderscore}{\kern0pt}scoped\ b{\isacartoucheclose}\isanewline
\isanewline
\isacommand{definition}\isamarkupfalse%
\ {\isacartoucheopen}{\isasymI}\ t{\isacharprime}{\kern0pt}\ x\ {\isacharequal}{\kern0pt}\ {\isacharparenleft}{\kern0pt}if\ consistent\ x\ t{\isacharprime}{\kern0pt}\ then\ {\isacharminus}{\kern0pt}{\isasyminfinity}{\isacharcolon}{\kern0pt}{\isacharcolon}{\kern0pt}ereal\ else\ {\isadigit{0}}{\isacharparenright}{\kern0pt}{\isacartoucheclose}\isanewline
\isacommand{definition}\isamarkupfalse%
\ {\isacartoucheopen}{\isasymI}{\isacharprime}{\kern0pt}\ {\isacharequal}{\kern0pt}\ map\ {\isacharparenleft}{\kern0pt}{\isasymlambda}t{\isacharprime}{\kern0pt}{\isachardot}{\kern0pt}\ {\isacharparenleft}{\kern0pt}instantiate\ {\isacharparenleft}{\kern0pt}{\isasymI}\ t{\isacharprime}{\kern0pt}{\isacharparenright}{\kern0pt}\ t{\isacharcomma}{\kern0pt}\ dom\ t{\isacharprime}{\kern0pt}\ {\isacharminus}{\kern0pt}\ dom\ t{\isacharparenright}{\kern0pt}{\isacharparenright}{\kern0pt}\ ts{\isacartoucheclose}\isanewline
\isanewline
\isacommand{definition}\isamarkupfalse%
\ {\isacartoucheopen}scope{\isacharunderscore}{\kern0pt}{\isasymI}{\isacharprime}{\kern0pt}\ {\isacharequal}{\kern0pt}\ {\isasymUnion}{\isacharparenleft}{\kern0pt}set\ {\isacharparenleft}{\kern0pt}map\ snd\ {\isacharparenleft}{\kern0pt}{\isasymI}{\isacharprime}{\kern0pt}{\isacharparenright}{\kern0pt}{\isacharparenright}{\kern0pt}{\isacharparenright}{\kern0pt}{\isacartoucheclose}\isanewline
\isacommand{definition}\isamarkupfalse%
\ {\isacartoucheopen}b{\isacharprime}{\kern0pt}\ {\isacharequal}{\kern0pt}\ b\ {\isacharat}{\kern0pt}\ \ {\isasymI}{\isacharprime}{\kern0pt}{\isacartoucheclose}\isanewline
\isacommand{definition}\isamarkupfalse%
\ {\isacartoucheopen}neg{\isacharunderscore}{\kern0pt}b{\isacharprime}{\kern0pt}\ {\isacharequal}{\kern0pt}\ neg{\isacharunderscore}{\kern0pt}b\ {\isacharat}{\kern0pt}\ {\isasymI}{\isacharprime}{\kern0pt}{\isacartoucheclose}\isanewline
\isanewline
\isacommand{definition}\isamarkupfalse%
\ {\isacartoucheopen}{\isasymOmega}{\isacharunderscore}{\kern0pt}pos\ {\isacharequal}{\kern0pt}\ factored{\isacharunderscore}{\kern0pt}lp\ True\ C\ neg{\isacharunderscore}{\kern0pt}b{\isacharprime}{\kern0pt}{\isacartoucheclose}\isanewline
\isacommand{definition}\isamarkupfalse%
\ {\isacartoucheopen}{\isasymOmega}{\isacharunderscore}{\kern0pt}neg\ {\isacharequal}{\kern0pt}\ factored{\isacharunderscore}{\kern0pt}lp\ False\ neg{\isacharunderscore}{\kern0pt}C\ b{\isacharprime}{\kern0pt}{\isacartoucheclose}\isanewline
\isacommand{definition}\isamarkupfalse%
\ {\isacartoucheopen}{\isasymOmega}\ {\isacharequal}{\kern0pt}\ union{\isacharunderscore}{\kern0pt}constr\ {\isasymOmega}{\isacharunderscore}{\kern0pt}pos\ {\isasymOmega}{\isacharunderscore}{\kern0pt}neg{\isacartoucheclose}\isanewline
\isacommand{end}\isamarkupfalse%
\end{isasnipenv}

\begin{isasnipenv}[snip:valdetbranchprop]{Linear Program for a Branch -- Correctness}
\isacommand{lemma}\isamarkupfalse%
\ {\isasymOmega}{\isacharunderscore}{\kern0pt}set{\isacharunderscore}{\kern0pt}correct{\isacharprime}{\kern0pt}{\isacharcolon}{\kern0pt}\isanewline
\ \ \isakeyword{shows}\ {\isacartoucheopen}constr{\isacharunderscore}{\kern0pt}set\ {\isasymOmega}\ {\isacharequal}{\kern0pt}\ {\isacharbraceleft}{\kern0pt}{\isacharparenleft}{\kern0pt}{\isasymphi}{\isacharcomma}{\kern0pt}\ w{\isacharparenright}{\kern0pt}{\isachardot}{\kern0pt}\ \isanewline
\ \ \ \ {\isasymforall}x\ {\isasymin}\ {\isacharbraceleft}{\kern0pt}x{\isachardot}{\kern0pt}\ x\ {\isasymin}\ states\ {\isasymand}\ consistent\ x\ t\ {\isasymand}\ {\isacharparenleft}{\kern0pt}{\isasymforall}t{\isacharprime}{\kern0pt}\ {\isasymin}\ set\ ts{\isachardot}{\kern0pt}\ {\isasymnot}\ consistent\ x\ t{\isacharprime}{\kern0pt}{\isacharparenright}{\kern0pt}{\isacharbraceright}{\kern0pt}{\isachardot}{\kern0pt}
\end{isasnipenv}

\begin{isasnipenv}[snip:flp]{Factored Linear Program}
\isacommand{locale}\isamarkupfalse%
\ factored{\isacharunderscore}{\kern0pt}lp{\isacharunderscore}{\kern0pt}consts\ {\isacharequal}{\kern0pt}\ \isakeyword{fixes}\isanewline
\ \ C\ {\isacharcolon}{\kern0pt}{\isacharcolon}{\kern0pt}\ {\isachardoublequoteopen}{\isacharparenleft}{\kern0pt}{\isacharparenleft}{\kern0pt}{\isacharparenleft}{\kern0pt}nat\ {\isasymrightharpoonup}\ {\isacharprime}{\kern0pt}a{\isacharparenright}{\kern0pt}\ {\isasymRightarrow}\ real{\isacharparenright}{\kern0pt}\ {\isasymtimes}\ nat\ set{\isacharparenright}{\kern0pt}\ list{\isachardoublequoteclose}\ \isakeyword{and}\isanewline
\ \ B\ {\isacharcolon}{\kern0pt}{\isacharcolon}{\kern0pt}\ {\isachardoublequoteopen}{\isacharparenleft}{\kern0pt}{\isacharparenleft}{\kern0pt}{\isacharparenleft}{\kern0pt}nat\ {\isasymrightharpoonup}\ {\isacharprime}{\kern0pt}a{\isacharparenright}{\kern0pt}\ {\isasymRightarrow}\ ereal{\isacharparenright}{\kern0pt}\ {\isasymtimes}\ nat\ set{\isacharparenright}{\kern0pt}\ list{\isachardoublequoteclose}\ \isakeyword{and}\isanewline
\isanewline
doms\ {\isacharcolon}{\kern0pt}{\isacharcolon}{\kern0pt}\ {\isachardoublequoteopen}nat\ {\isasymRightarrow}\ {\isacharprime}{\kern0pt}a\ set{\isachardoublequoteclose}\ \isakeyword{and}\isanewline
\isamarkupcmt{Each dimension has a domain%
}\isanewline
\isanewline
dims\ {\isacharcolon}{\kern0pt}{\isacharcolon}{\kern0pt}\ nat\isanewline
\isamarkupcmt{for each \isa{i\ {\isacharless}{\kern0pt}\ num{\isacharunderscore}{\kern0pt}c}, \isa{c\ i\ v} assigns to each vector \isa{v\ {\isasymin}\ Dom\ num{\isacharunderscore}{\kern0pt}c} a real value%
}\ \isakeyword{and}\isanewline
\isanewline
prefix\ {\isacharcolon}{\kern0pt}{\isacharcolon}{\kern0pt}\ {\isacartoucheopen}{\isacharprime}{\kern0pt}x{\isacartoucheclose}\ \isakeyword{and}\isanewline
\isamarkupcmt{prefixes are added to private LP variables to make them distinct between branches%
}\isanewline
\isanewline
order\ {\isacharcolon}{\kern0pt}{\isacharcolon}{\kern0pt}\ {\isachardoublequoteopen}nat\ {\isasymRightarrow}\ nat{\isachardoublequoteclose}\isanewline
\isakeyword{begin}%
\isanewline
\begin{isamarkuptext}%
Enumerate all functions in c + all states with matching scopes.%
\end{isamarkuptext}\isamarkuptrue%
\isacommand{definition}\isamarkupfalse%
\ {\isacartoucheopen}vars{\isacharunderscore}{\kern0pt}c\ {\isacharequal}{\kern0pt}\ {\isacharbraceleft}{\kern0pt}var{\isacharunderscore}{\kern0pt}f{\isacharprime}{\kern0pt}\ {\isacharparenleft}{\kern0pt}f{\isacharunderscore}{\kern0pt}c\ i{\isacharparenright}{\kern0pt}\ z\ {\isacharbar}{\kern0pt}\ z\ i{\isachardot}{\kern0pt}\ z\ {\isasymin}\ vecs{\isacharunderscore}{\kern0pt}on\ {\isacharparenleft}{\kern0pt}scope{\isacharunderscore}{\kern0pt}c\ i{\isacharparenright}{\kern0pt}\ {\isasymand}\ i\ {\isacharless}{\kern0pt}\ num{\isacharunderscore}{\kern0pt}c{\isacharbraceright}{\kern0pt}{\isacartoucheclose}%
\isanewline
\isacommand{definition}\isamarkupfalse%
\ {\isacartoucheopen}constr{\isacharunderscore}{\kern0pt}c\ z\ i\ {\isacharequal}{\kern0pt}\ Eq\ {\isacharparenleft}{\kern0pt}{\isasymlambda}v{\isachardot}{\kern0pt}\isanewline
\ \ \ \ if\ v\ {\isacharequal}{\kern0pt}\ var{\isacharunderscore}{\kern0pt}f{\isacharprime}{\kern0pt}\ {\isacharparenleft}{\kern0pt}f{\isacharunderscore}{\kern0pt}c\ i{\isacharparenright}{\kern0pt}\ z\ then\ {\isacharminus}{\kern0pt}{\isadigit{1}}\isanewline
\ \ \ \ else\ if\ v\ {\isacharequal}{\kern0pt}\ var{\isacharunderscore}{\kern0pt}w\ i\ then\ c{\isacharunderscore}{\kern0pt}r\ i\ z\isanewline
\ \ \ \ else\ {\isadigit{0}}{\isacharparenright}{\kern0pt}\ {\isadigit{0}}{\isacartoucheclose}\isanewline
\isanewline
\isacommand{definition}\isamarkupfalse%
\ {\isachardoublequoteopen}constrs{\isacharunderscore}{\kern0pt}c\ {\isacharequal}{\kern0pt}\ \isanewline
\ \ {\isacharbraceleft}{\kern0pt}constr{\isacharunderscore}{\kern0pt}c\ z\ i\ {\isacharbar}{\kern0pt}\ z\ i{\isachardot}{\kern0pt}\ z\ {\isasymin}\ vecs{\isacharunderscore}{\kern0pt}on\ {\isacharparenleft}{\kern0pt}scope{\isacharunderscore}{\kern0pt}c\ i{\isacharparenright}{\kern0pt}\ {\isasymand}\ i\ {\isacharless}{\kern0pt}\ num{\isacharunderscore}{\kern0pt}c{\isacharbraceright}{\kern0pt}{\isachardoublequoteclose}\isanewline
\isanewline
\isacommand{definition}\isamarkupfalse%
\ {\isachardoublequoteopen}vars{\isacharunderscore}{\kern0pt}b\ {\isacharequal}{\kern0pt}\ {\isacharbraceleft}{\kern0pt}\ var{\isacharunderscore}{\kern0pt}f{\isacharprime}{\kern0pt}\ {\isacharparenleft}{\kern0pt}f{\isacharunderscore}{\kern0pt}b\ j{\isacharparenright}{\kern0pt}\ z\ {\isacharbar}{\kern0pt}\ z\ j{\isachardot}{\kern0pt}\ \isanewline
\ \ z\ {\isasymin}\ vecs{\isacharunderscore}{\kern0pt}on\ {\isacharparenleft}{\kern0pt}scope{\isacharunderscore}{\kern0pt}b\ j{\isacharparenright}{\kern0pt}\ {\isasymand}\ j\ {\isacharless}{\kern0pt}\ num{\isacharunderscore}{\kern0pt}b{\isacharbraceright}{\kern0pt}{\isachardoublequoteclose}%
\begin{isamarkuptext}%
Ensures \isa{f\ {\isacharparenleft}{\kern0pt}b{\isacharcomma}{\kern0pt}\ j{\isacharcomma}{\kern0pt}\ z{\isacharparenright}{\kern0pt}\ {\isacharequal}{\kern0pt}\ b{\isacharunderscore}{\kern0pt}j\ z}%
\end{isamarkuptext}\isamarkuptrue%
\isacommand{definition}\isamarkupfalse%
\ {\isachardoublequoteopen}constr{\isacharunderscore}{\kern0pt}b\ z\ j\ {\isacharequal}{\kern0pt}\isanewline
\ \ Eq\ {\isacharparenleft}{\kern0pt}{\isasymlambda}v{\isachardot}{\kern0pt}\ if\ v\ {\isacharequal}{\kern0pt}\ var{\isacharunderscore}{\kern0pt}f{\isacharprime}{\kern0pt}\ {\isacharparenleft}{\kern0pt}f{\isacharunderscore}{\kern0pt}b\ j{\isacharparenright}{\kern0pt}\ z\ then\ {\isadigit{1}}\ else\ {\isadigit{0}}{\isacharparenright}{\kern0pt}\ {\isacharparenleft}{\kern0pt}real{\isacharunderscore}{\kern0pt}of{\isacharunderscore}{\kern0pt}ereal\ {\isacharparenleft}{\kern0pt}b{\isacharunderscore}{\kern0pt}r\ j\ z{\isacharparenright}{\kern0pt}{\isacharparenright}{\kern0pt}{\isachardoublequoteclose}\isanewline
\isanewline
\isacommand{definition}\isamarkupfalse%
\ {\isachardoublequoteopen}constrs{\isacharunderscore}{\kern0pt}b\ {\isacharequal}{\kern0pt}\ \isanewline
\ \ {\isacharbraceleft}{\kern0pt}constr{\isacharunderscore}{\kern0pt}b\ z\ j\ {\isacharbar}{\kern0pt}\ z\ j{\isachardot}{\kern0pt}\ z\ {\isasymin}\ vecs{\isacharunderscore}{\kern0pt}on\ {\isacharparenleft}{\kern0pt}scope{\isacharunderscore}{\kern0pt}b\ j{\isacharparenright}{\kern0pt}\ {\isasymand}\ j\ {\isacharless}{\kern0pt}\ num{\isacharunderscore}{\kern0pt}b\ {\isasymand}\ b{\isacharunderscore}{\kern0pt}r\ j\ z\ {\isasymnoteq}\ {\isacharminus}{\kern0pt}{\isasyminfinity}{\isacharbraceright}{\kern0pt}{\isachardoublequoteclose}\isanewline
\isanewline
\isacommand{definition}\isamarkupfalse%
\ {\isachardoublequoteopen}vars{\isacharunderscore}{\kern0pt}w\ {\isacharequal}{\kern0pt}\ {\isacharbraceleft}{\kern0pt}var{\isacharunderscore}{\kern0pt}w\ i{\isacharbar}{\kern0pt}\ i{\isachardot}{\kern0pt}\ i\ {\isacharless}{\kern0pt}\ num{\isacharunderscore}{\kern0pt}c{\isacharbraceright}{\kern0pt}{\isachardoublequoteclose}%
\begin{isamarkuptext}%
Initial set of function to maximize constructed from B and C.%
\end{isamarkuptext}\isamarkuptrue%
\isacommand{definition}\isamarkupfalse%
\ {\isachardoublequoteopen}{\isasymF}{\isacharunderscore}{\kern0pt}init\ {\isacharequal}{\kern0pt}\ {\isacharbraceleft}{\kern0pt}{\isacharparenleft}{\kern0pt}f{\isacharunderscore}{\kern0pt}c\ i{\isacharparenright}{\kern0pt}\ {\isacharbar}{\kern0pt}\ i{\isachardot}{\kern0pt}\ i\ {\isacharless}{\kern0pt}\ num{\isacharunderscore}{\kern0pt}c{\isacharbraceright}{\kern0pt}\ {\isasymunion}\ {\isacharbraceleft}{\kern0pt}{\isacharparenleft}{\kern0pt}f{\isacharunderscore}{\kern0pt}b\ i{\isacharparenright}{\kern0pt}\ {\isacharbar}{\kern0pt}\ i{\isachardot}{\kern0pt}\ i\ {\isacharless}{\kern0pt}\ num{\isacharunderscore}{\kern0pt}b{\isacharbraceright}{\kern0pt}{\isachardoublequoteclose}\isanewline
\isanewline
\isacommand{definition}\isamarkupfalse%
\ {\isachardoublequoteopen}scopes{\isacharunderscore}{\kern0pt}init\ f\ {\isacharequal}{\kern0pt}\ {\isacharparenleft}{\kern0pt}case\ f\ of\ f{\isacharunderscore}{\kern0pt}b\ i\ {\isasymRightarrow}\ scope{\isacharunderscore}{\kern0pt}b\ i\ {\isacharbar}{\kern0pt}\ f{\isacharunderscore}{\kern0pt}c\ i\ {\isasymRightarrow}\ scope{\isacharunderscore}{\kern0pt}c\ i\ {\isacharbar}{\kern0pt}\ {\isacharunderscore}{\kern0pt}\ {\isasymRightarrow}\ {\isacharbraceleft}{\kern0pt}{\isacharbraceright}{\kern0pt}{\isacharparenright}{\kern0pt}{\isachardoublequoteclose}\isanewline
\isanewline
\isacommand{definition}\isamarkupfalse%
\ {\isachardoublequoteopen}constrs{\isacharunderscore}{\kern0pt}init\ {\isacharequal}{\kern0pt}\ constrs{\isacharunderscore}{\kern0pt}c\ {\isasymunion}\ constrs{\isacharunderscore}{\kern0pt}b{\isachardoublequoteclose}\isanewline
\isanewline
\isacommand{definition}\isamarkupfalse%
\ {\isachardoublequoteopen}vars{\isacharunderscore}{\kern0pt}init\ {\isacharequal}{\kern0pt}\ vars{\isacharunderscore}{\kern0pt}c\ {\isasymunion}\ vars{\isacharunderscore}{\kern0pt}b\ {\isasymunion}\ vars{\isacharunderscore}{\kern0pt}w{\isachardoublequoteclose}%
\begin{isamarkuptext}%
Create constraints that ensure \isa{f{\isacharunderscore}{\kern0pt}e} is at least the sum of the functions in E:
  sum f in E, \isatt{f{\kern0pt}({\kern0pt}z{\kern0pt}({\kern0pt}l{\kern0pt}\ :{\kern0pt}={\kern0pt}\ x{\kern0pt}l{\kern0pt}){\kern0pt}\ {\char`\<}{\kern0pt}={\kern0pt}\ f{\kern0pt}({\kern0pt}e{\kern0pt}{\char`\,}{\kern0pt}\ l{\kern0pt}){\kern0pt}}.%
\end{isamarkuptext}\isamarkuptrue%
\isacommand{definition}\isamarkupfalse%
\ {\isachardoublequoteopen}constr{\isacharunderscore}{\kern0pt}max\ E\ l\ scopes\ z\ xl\ {\isacharequal}{\kern0pt}\isanewline
\ \ Le\ {\isacharparenleft}{\kern0pt}{\isasymlambda}v{\isachardot}{\kern0pt}\isanewline
\ \ \ \ case\ v\ of\isanewline
\ \ \ \ \ \ var{\isacharunderscore}{\kern0pt}f\ p\ f\ z{\isacharprime}{\kern0pt}\ {\isasymRightarrow}\isanewline
\ \ \ \ if\ p\ {\isacharequal}{\kern0pt}\ prefix\ {\isasymand}\ f\ {\isacharequal}{\kern0pt}\ f{\isacharunderscore}{\kern0pt}e\ l\ {\isasymand}\ z{\isacharprime}{\kern0pt}\ {\isacharequal}{\kern0pt}\ z\ then\ {\isacharminus}{\kern0pt}{\isadigit{1}}\isanewline
\ \ \ \ else\ if\ p\ {\isacharequal}{\kern0pt}\ prefix\ {\isasymand}\ f\ {\isasymin}\ E\ {\isasymand}\ z{\isacharprime}{\kern0pt}\ {\isacharequal}{\kern0pt}\ {\isacharparenleft}{\kern0pt}z{\isacharparenleft}{\kern0pt}l\ {\isasymmapsto}\ xl{\isacharparenright}{\kern0pt}{\isacharparenright}{\kern0pt}\ {\isacharbar}{\kern0pt}{\isacharbackquote}{\kern0pt}\ {\isacharparenleft}{\kern0pt}scopes\ f{\isacharparenright}{\kern0pt}\ then\ {\isadigit{1}}\isanewline
\ \ \ \ else\ {\isadigit{0}}\isanewline
\ \ \ \ {\isacharbar}{\kern0pt}\ {\isacharunderscore}{\kern0pt}\ {\isasymRightarrow}\ {\isadigit{0}}{\isacharparenright}{\kern0pt}\ {\isadigit{0}}{\isachardoublequoteclose}\isanewline
\isanewline
\isacommand{definition}\isamarkupfalse%
\ {\isachardoublequoteopen}constrs{\isacharunderscore}{\kern0pt}max\ E\ l\ scopes\ scope{\isacharunderscore}{\kern0pt}e\ {\isacharequal}{\kern0pt}\ \isanewline
\ \ {\isacharbraceleft}{\kern0pt}constr{\isacharunderscore}{\kern0pt}max\ E\ l\ scopes\ z\ xl\ {\isacharbar}{\kern0pt}\ xl\ z{\isachardot}{\kern0pt}\ z\ {\isasymin}\ vecs{\isacharunderscore}{\kern0pt}on\ scope{\isacharunderscore}{\kern0pt}e\ {\isasymand}\ xl\ {\isasymin}\ doms\ l{\isacharbraceright}{\kern0pt}{\isachardoublequoteclose}%
\begin{isamarkuptext}%
Single iteration step: 
1. select variable to eliminate, 
2. partition functions based on that,
3. create new function (variable) with constraints to ensure maximization.%
\end{isamarkuptext}\isamarkuptrue%
\isacommand{definition}\isamarkupfalse%
\ {\isachardoublequoteopen}elim{\isacharunderscore}{\kern0pt}step\ {\isasymOmega}\ {\isasymF}\ scopes\ i\ {\isacharequal}{\kern0pt}\ {\isacharparenleft}{\kern0pt}let\isanewline
\ \ l\ {\isacharequal}{\kern0pt}\ order\ i{\isacharsemicolon}{\kern0pt}\isanewline
\ \ E\ {\isacharequal}{\kern0pt}\ {\isacharbraceleft}{\kern0pt}e\ {\isacharbar}{\kern0pt}\ e{\isachardot}{\kern0pt}\ e\ {\isasymin}\ {\isasymF}\ {\isasymand}\ l\ {\isasymin}\ scopes\ e{\isacharbraceright}{\kern0pt}{\isacharsemicolon}{\kern0pt}\isanewline
\ \ scope{\isacharunderscore}{\kern0pt}e\ {\isacharequal}{\kern0pt}\ {\isacharparenleft}{\kern0pt}{\isasymUnion}e\ {\isasymin}\ E{\isachardot}{\kern0pt}\ scopes\ e{\isacharparenright}{\kern0pt}\ {\isacharminus}{\kern0pt}\ {\isacharbraceleft}{\kern0pt}l{\isacharbraceright}{\kern0pt}{\isacharsemicolon}{\kern0pt}\isanewline
\ \ {\isasymOmega}{\isacharprime}{\kern0pt}\ {\isacharequal}{\kern0pt}\ {\isasymOmega}\ {\isasymunion}\ constrs{\isacharunderscore}{\kern0pt}max\ E\ l\ scopes\ scope{\isacharunderscore}{\kern0pt}e\ in\isanewline
\ \ {\isacharparenleft}{\kern0pt}{\isasymOmega}{\isacharprime}{\kern0pt}{\isacharcomma}{\kern0pt}\ {\isasymF}\ {\isacharminus}{\kern0pt}\ E\ {\isasymunion}\ {\isacharbraceleft}{\kern0pt}f{\isacharunderscore}{\kern0pt}e\ l{\isacharbraceright}{\kern0pt}{\isacharcomma}{\kern0pt}\ scopes{\isacharparenleft}{\kern0pt}f{\isacharunderscore}{\kern0pt}e\ l\ {\isacharcolon}{\kern0pt}{\isacharequal}{\kern0pt}\ scope{\isacharunderscore}{\kern0pt}e{\isacharparenright}{\kern0pt}{\isacharcomma}{\kern0pt}\ i{\isacharplus}{\kern0pt}{\isadigit{1}}{\isacharparenright}{\kern0pt}\isanewline
{\isacharparenright}{\kern0pt}{\isachardoublequoteclose}\isanewline
\isanewline
\isacommand{definition}\isamarkupfalse%
\ {\isachardoublequoteopen}elim{\isacharunderscore}{\kern0pt}vars{\isacharunderscore}{\kern0pt}aux\ {\isacharequal}{\kern0pt}\ \isanewline
\ \ {\isacharparenleft}{\kern0pt}{\isacharparenleft}{\kern0pt}{\isasymlambda}{\isacharparenleft}{\kern0pt}{\isasymOmega}{\isacharcomma}{\kern0pt}\ {\isasymF}{\isacharcomma}{\kern0pt}\ scopes{\isacharcomma}{\kern0pt}\ i{\isacharparenright}{\kern0pt}{\isachardot}{\kern0pt}\ elim{\isacharunderscore}{\kern0pt}step\ {\isasymOmega}\ {\isasymF}\ scopes\ i{\isacharparenright}{\kern0pt}\ {\isacharcircum}{\kern0pt}{\isacharcircum}{\kern0pt}\ dims{\isacharparenright}{\kern0pt}\ {\isacharparenleft}{\kern0pt}constrs{\isacharunderscore}{\kern0pt}init{\isacharcomma}{\kern0pt}\ {\isasymF}{\isacharunderscore}{\kern0pt}init{\isacharcomma}{\kern0pt}\ scopes{\isacharunderscore}{\kern0pt}init{\isacharcomma}{\kern0pt}\ {\isadigit{0}}{\isacharparenright}{\kern0pt}{\isachardoublequoteclose}\isanewline
\isanewline
\isacommand{definition}\isamarkupfalse%
\ {\isachardoublequoteopen}gen{\isacharunderscore}{\kern0pt}constrs\ arg\ {\isacharequal}{\kern0pt}\ {\isacharparenleft}{\kern0pt}\isanewline
\ \ case\ arg\ of\ {\isacharparenleft}{\kern0pt}{\isasymOmega}{\isacharcomma}{\kern0pt}\ {\isasymF}{\isacharcomma}{\kern0pt}\ scopes{\isacharcomma}{\kern0pt}\ i{\isacharparenright}{\kern0pt}\ {\isasymRightarrow}\ {\isacharparenleft}{\kern0pt}{\isasymOmega}\ {\isasymunion}\ {\isacharbraceleft}{\kern0pt}Le\ {\isacharparenleft}{\kern0pt}{\isasymlambda}v{\isachardot}{\kern0pt}\isanewline
\ \ \ \ case\ v\ of\isanewline
\ \ \ \ \ \ var{\isacharunderscore}{\kern0pt}f\ p\ f\ z\ {\isasymRightarrow}\ if\ p\ {\isacharequal}{\kern0pt}\ prefix\ {\isasymand}\ f\ {\isasymin}\ {\isasymF}\ {\isasymand}\ z\ {\isacharequal}{\kern0pt}\ Map{\isachardot}{\kern0pt}empty\ then\ {\isadigit{1}}\ else\ {\isadigit{0}}\isanewline
\ \ \ \ {\isacharbar}{\kern0pt}\ var{\isacharunderscore}{\kern0pt}phi\ {\isasymRightarrow}\ {\isacharminus}{\kern0pt}{\isadigit{1}}\isanewline
\ \ \ \ {\isacharbar}{\kern0pt}\ {\isacharunderscore}{\kern0pt}\ {\isasymRightarrow}\ {\isadigit{0}}{\isacharparenright}{\kern0pt}\ {\isadigit{0}}{\isacharbraceright}{\kern0pt}{\isacharparenright}{\kern0pt}{\isacharparenright}{\kern0pt}{\isachardoublequoteclose}\isanewline
\isanewline
\isacommand{definition}\isamarkupfalse%
\ {\isachardoublequoteopen}elim{\isacharunderscore}{\kern0pt}vars\ {\isacharequal}{\kern0pt}\ gen{\isacharunderscore}{\kern0pt}constrs\ elim{\isacharunderscore}{\kern0pt}vars{\isacharunderscore}{\kern0pt}aux{\isachardoublequoteclose}\isanewline
\isanewline
\isacommand{end}\isamarkupfalse%
\end{isasnipenv}

\begin{isasnipenv}[snip:flpprop]{Factored LP Correctness}
\isacommand{lemma}\isamarkupfalse%
\ constr{\isacharunderscore}{\kern0pt}set{\isacharunderscore}{\kern0pt}factored{\isacharunderscore}{\kern0pt}eq{\isacharprime}{\kern0pt}{\isacharcolon}{\kern0pt}\isanewline
\ {\isacartoucheopen}{\isacharbraceleft}{\kern0pt}{\isacharparenleft}{\kern0pt}x\ var{\isacharunderscore}{\kern0pt}phi{\isacharcomma}{\kern0pt}\ {\isasymlambda}i{\isachardot}{\kern0pt}\ x\ {\isacharparenleft}{\kern0pt}var{\isacharunderscore}{\kern0pt}w\ i{\isacharparenright}{\kern0pt}{\isacharparenright}{\kern0pt}\ {\isacharbar}{\kern0pt}x{\isachardot}{\kern0pt}\ x\ {\isasymin}\ {\isacharparenleft}{\kern0pt}{\isacharbraceleft}{\kern0pt}v\ {\isacharbar}{\kern0pt}v{\isachardot}{\kern0pt}\ Ball\ elim{\isacharunderscore}{\kern0pt}vars\ {\isacharparenleft}{\kern0pt}sat{\isacharunderscore}{\kern0pt}constr\ v{\isacharparenright}{\kern0pt}{\isacharbraceright}{\kern0pt}{\isacharparenright}{\kern0pt}{\isacharbraceright}{\kern0pt}\ {\isacharequal}{\kern0pt}\isanewline
\ \ \ \ {\isacharbraceleft}{\kern0pt}{\isacharparenleft}{\kern0pt}{\isasymphi}{\isacharcomma}{\kern0pt}\ w{\isacharparenright}{\kern0pt}{\isachardot}{\kern0pt}\ {\isasymforall}x{\isasymin}full{\isacharunderscore}{\kern0pt}vecs{\isachardot}{\kern0pt}\isanewline
\ \ \ \ \ \   ereal\ {\isacharparenleft}{\kern0pt}{\isasymSum}i{\isacharless}{\kern0pt}length\ C{\isachardot}{\kern0pt}\ w\ i\ {\isacharasterisk}{\kern0pt}\ fst\ {\isacharparenleft}{\kern0pt}C\ {\isacharbang}{\kern0pt}\ i{\isacharparenright}{\kern0pt}\ x{\isacharparenright}{\kern0pt}\ {\isacharplus}{\kern0pt}\isanewline
\ \ \ \ \ \ \ \  {\isacharparenleft}{\kern0pt}{\isasymSum}i{\isacharless}{\kern0pt}length\ B{\isachardot}{\kern0pt}\ fst\ {\isacharparenleft}{\kern0pt}B\ {\isacharbang}{\kern0pt}\ i{\isacharparenright}{\kern0pt}\ x{\isacharparenright}{\kern0pt}\ {\isasymle}\ ereal\ {\isasymphi}{\isacharbraceright}{\kern0pt}{\isacartoucheclose}
\end{isasnipenv}

\end{document}